\documentclass[11pt]{article}

\usepackage{acl}

\usepackage{times}
\usepackage{latexsym}

\usepackage[T1]{fontenc}

\usepackage[utf8]{inputenc}

\usepackage{microtype}

\usepackage{inconsolata}

\usepackage{graphicx}

\usepackage{multirow}
\usepackage{graphicx}
\usepackage{booktabs}
\usepackage{amsmath}
\usepackage{amssymb}
\usepackage{siunitx}
\usepackage{enumitem}

\title{Dissecting Training-Free Uncertainty Estimation in Multimodal LLMs}

\author{Soroush Seifi\textsuperscript{1,\dag} \quad
  Vaggelis Dorovatas\textsuperscript{2} \quad
  Lin Li\textsuperscript{3} \quad
  Yarin Gal\textsuperscript{3} \quad
  Rahaf Aljundi\textsuperscript{1} \\[6pt]
  \textsuperscript{1}Toyota Motor Europe\\
  \textsuperscript{2}New York University \\
  \textsuperscript{3}University of Oxford \\[4pt]}

\begin{document}
\maketitle
\begingroup
\renewcommand\thefootnote{\dag}
\footnotetext{providing contracted services at Toyota Motor Europe.}
\endgroup

\begin{abstract}
Multimodal Large Language Models (MLLMs) are increasingly deployed in safety-critical settings, yet estimating when their answers can be trusted remains underexplored. We systematically compare training-free verbalized, token-level and semantic uncertainty estimation methods on twelve MLLMs and five VQA benchmarks. We find that answer length decides which family wins: token-level entropy on short answers, verbalized methods on sentence-length responses, and semantic methods on long-form generation. Token-level entropy becomes far more informative when estimated at a higher temperature than the low ones used to generate answers, and asking a model to judge its own answer becomes more reliable with model size. Building on this, a length-aware recipe that additionally flags responses on which the model abstains outperforms every individual method on average at about a quarter of the cost of semantic methods, although no method attains the coverage required for safe deployment. These results provide practitioners with principled guidance for selecting estimators by answer length and model scale.
\end{abstract}

\section{Introduction}
\label{sec:intro}
Large language models (LLMs)~\cite{brown2020language, zhao2023survey} and their multimodal counterparts (MLLMs)~\cite{li2025survey, ghosh2024exploring} are increasingly relied upon~\cite{eloundou2024gpts} for high-stakes applications, including agentic systems~\cite{wang2024survey, yehudai2025survey} that autonomously execute multi-step workflows. Yet no current model is reliable across all tasks, and hallucinated outputs~\cite{huang2025survey, liu2024survey, li2023evaluating} constrain deployment in safety-critical domains such as medical diagnosis~\cite{kim2025medical, hakim2024need} or autonomous driving~\cite{dona2025bettercheck}. Safe deployment therefore requires reliable uncertainty estimates, enabling a system to abstain, defer to a human, trigger verification, or route the query to a stronger model.

\begin{figure}[!h]
    \centering
    \includegraphics[width=\linewidth]{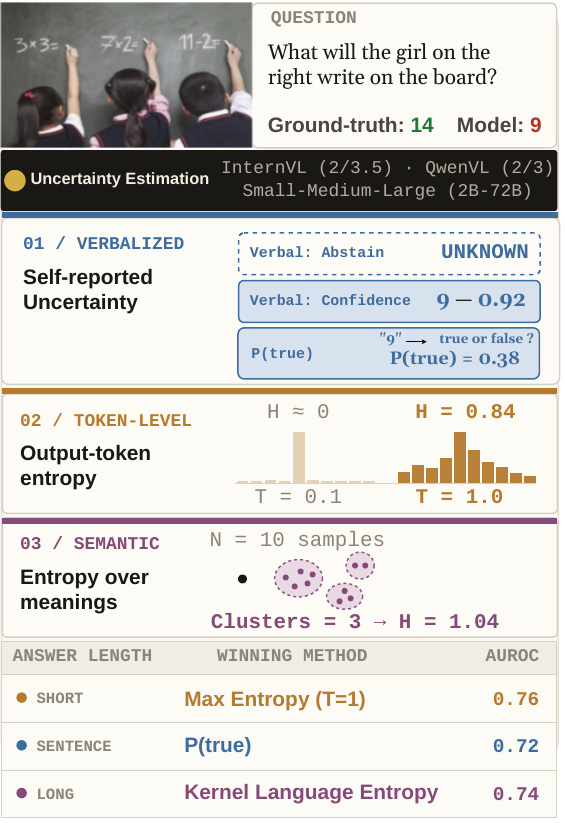}
    \caption{We benchmark three families of uncertainty estimation methods — verbalized, token-level, and semantic — across 12 multimodal LLMs and 5 benchmarks, finding that the optimal strategy depends strongly on response length.}
    \label{fig:teaser}
\end{figure}

While Uncertainty Estimation (UE) has been extensively studied for LLMs~\cite{xia-etal-2025-survey, shorinwa2025survey}, MLLMs introduce additional challenges that remain underexplored: visual inputs add their own sources of ambiguity, and VQA answers range from a single token to multi-sentence descriptions. Most UE methods, however, are validated on short-answer QA~\cite{zhang2024vl, kadavath2022language,lin2022teaching} and a narrow set of models~\cite{zhang2024vl, chen-etal-2025-unveiling}, leaving practitioners without guidance on how UE behavior varies with model family, generation, scale, and response length.

To address this gap, we present an empirical analysis of training-free UE in MLLMs, targeting methods that are practical to deploy without additional training or paired data. We organize existing approaches into three families --- verbalized~\cite{lin2022teaching, kadavath2022language, xiong2024can}, token-level~\cite{malinin2020uncertainty, guerreiro2023looking, li-etal-2024-reference}, and semantic~\cite{kuhn2023semantic, farquhar2024detecting, nikitin2024kernel} --- and analyze how their reliability varies with model family, generation, size and response-length regime. Concretely, we compare representative methods of each family on twelve open-weight MLLMs (InternVL and QwenVL, 2B to 72B) and five VQA benchmarks, judging every estimator by how well it predicts whether the model's answer is correct. 
Our analysis yields four main takeaways:
\begin{itemize}[leftmargin=*,itemsep=1pt,topsep=2pt]
\item \textbf{Answer length decides which UE family wins.} Token-level entropy nearly matches the best method on short answers at a fraction of its cost, verbalized methods lead on sentence-length answers, semantic methods on long-form answers.
\item \textbf{Token-level entropy needs a higher temperature than answer generation.} At the low generation temperature, entropy carries little signal; an extra higher-temperature pass recovers it.
\item \textbf{Self-evaluation scales with model size.} A model's judgement of its own answer is unreliable for small MLLMs but improves with size; for large models it beats sampling-based methods on long-form answers at a fraction of the cost.
\item \textbf{No method is ready for safe deployment.} Even at a loose risk budget, the best estimators answer far fewer questions than perfect ranking allows.
\end{itemize}

Building on these findings, we turn the regime dependence into a deployment recipe: a length-aware router chooses the estimator from the length of the model's answer and additionally flags answers on which the model abstains. Averaged over all benchmarks, it outperforms every single method at about a quarter of the cost of semantic methods.

\section{Related Work}
\paragraph{Uncertainty estimation in LLMs.}
Uncertainty estimation has been extensively studied for unimodal LLMs, spanning verbalized confidence~\cite{lin2022teaching, kadavath2022language, xiong2024can}, token-level signals~\cite{malinin2020uncertainty, guerreiro2023looking, li-etal-2024-reference}, semantic-level methods~\cite{kuhn2023semantic, farquhar2024detecting} and their soft-similarity refinement~\cite{nikitin2024kernel}, consistency-based detectors that localize unsupported claims in long-form text~\cite{manakul-etal-2023-selfcheckgpt, zhang-etal-2024-luq}, and long-form calibration~\cite{zhang-etal-2025-atomic}; see~\citet{xia-etal-2025-survey, shorinwa2025survey, geng-etal-2024-survey} for overviews. LM-Polygraph~\cite{fadeeva-etal-2023-lm, vashurin-etal-2025-benchmarking} benchmarks them for unimodal LLMs under one protocol; training-based probes~\cite{kossen2024semantic} are cheaper alternatives. How these methods transfer to multimodal settings---where visual ambiguity and a wider range of output formats come into play---remains an open question.

\paragraph{Uncertainty and hallucination in MLLMs.}
Recent surveys characterize hallucination in MLLMs as a misalignment between visual inputs and textual outputs, organizing errors into object-, attribute-, and relation-level inconsistencies and reviewing language-prior bias, weak visual grounding, and decoding-level mitigations~\cite{bai2024hallucination, liu2024survey}; \citet{chen2026survey} add confidence- and uncertainty-based detection signals, and \citet{sahoo-etal-2024-comprehensive} adopt a broader foundation-model view. Estimators for vision--language models have been proposed one at a time: \citet{khan2024consistency} use answer consistency across rephrasings for black-box selective VQA, VL-Uncertainty~\cite{zhang2024vl} clusters responses sampled under visual perturbations, and VAUQ~\cite{park-etal-2026-vauq} scores how much the image reduces the model's own predictive uncertainty. All are evaluated on short-answer VQA with models of up to 13B parameters.

\paragraph{Our position.}
Our taxonomy follows the three families established in the LLM literature. The study covers twelve open-weight models from 2B to 72B parameters, beyond the scales of prior comparisons, and response-length regimes from multiple choice to multi-sentence generation, with controlled length and format ablations. We treat uncertainty estimation as a general reliability framework for MLLMs, of which hallucination detection is one application, and judge every estimator on the same task as confidence estimation~\cite{kadavath2022language, tian-etal-2023-just, xiong2024can} and selective prediction~\cite{geifman2017selective, kamath-etal-2020-selective, whitehead2022reliablevqa, dancette2023improving}: how well its score predicts whether the model's answer is correct, the decision a deployed system must make before acting on that answer. From this comparison we derive actionable guidance for practitioners, including a length-aware deployment recipe.

\section{Uncertainty Estimation Families}
\label{sec:method}
We analyze three families of training-free UE, distinguished by where uncertainty is read off: the model's own statements (verbalized), its output distributions (token-level), or the variability of sampled responses at the meaning level (semantic).

\subsection{Verbalized Uncertainty}
\label{sec:verbalized}

Prior studies find MLLMs overconfident and poorly calibrated when verbalizing numerical confidence~\cite{groot-valdenegro-toro-2024-overconfidence, borszukovszki-etal-2025-know, xuan-etal-2025-seeing, zhao2025object}, while abstention prompting~\cite{wang2025vision} and prompt optimization~\cite{kriz2025prompt4trust} improve robustness. We study three variants: (i) \emph{abstention}, where the model outputs ``\texttt{UNKNOWN}'' when uncertain, (ii) \emph{numerical confidence}, where the model outputs a confidence score alongside each answer, and (iii) \emph{P(true)}~\cite{kadavath2022language}, a self-evaluation where the model is shown its own answer and asked whether it is (A)~True or (B)~False; we take the probability of \emph{True}, normalized over the two options, $1 - P(\text{true})$ is then used as the uncertainty estimate (prompt and option parsing in App.~\ref{app:impl}).
The first two verbalized methods apply to proprietary and open-source models at no extra inference cost; P(true) reads the output distribution and therefore requires logit access and one additional forward pass. All three rely on the model's self-calibration.

\subsection{Token-level Uncertainty}
Prior work leverages model-internal signals to detect hallucinations in MLLMs, either by suppressing uncertain visual tokens~\cite{fang2024enhancing, seo2025epistemic} or by quantifying uncertainty in output token distributions during autoregressive decoding~\cite{li-etal-2024-reference}. Following the latter, we adopt token-level entropy and negative log-likelihood (NLL). Given an autoregressive MLLM generating $\mathbf{y} = (y_1, \dots, y_L)$ conditioned on input $\mathbf{x}$, with predictive distribution $p_\theta(y_l \mid \mathbf{x}, y_{<l})$ at position $l \in \{1,\dots,L\}$ over vocabulary $\mathcal{V}$ at sampling temperature $T$:
\begin{equation}
H_l = - \!\!\sum_{v \in \mathcal{V}}\! p_\theta(v \mid \mathbf{x}, y_{<l}) \log p_\theta(v \mid \mathbf{x}, y_{<l}),
\end{equation}
\begin{equation}
\mathrm{NLL}_l = - \log p_\theta(y_l \mid \mathbf{x}, y_{<l}).
\end{equation}
Entropy captures uncertainty over all continuations; NLL reflects confidence in the realized trajectory. We aggregate each over the response by \emph{mean} (length-normalized global uncertainty) and \emph{max} (localized spikes corresponding to weakly grounded tokens). These methods require logit access and do not apply to closed-source models; they cost nothing extra when read from the answer's own decoding pass. Estimating at a different temperature needs a second, estimation-only generation: a different token sampled at any step changes every later distribution, so $T{=}1$ statistics cannot be obtained by rescaling the stored $T{=}0.1$ logits. The deployed answer is always the $T{=}0.1$ one and all metrics score its correctness; divergence of the $T{=}1$ pass from it is the signal, not a confound.

\subsection{Semantic Uncertainty}
Semantic Entropy (SE)~\cite{farquhar2024detecting} computes uncertainty over \emph{semantic meanings} rather than surface forms, accounting for paraphrastic and lexical invariances. Given $N$ samples from $p(s \mid x)$, responses are grouped into $K$ semantic classes $\{C_k\}_{k=1}^{K}$ via bi-directional entailment, and entropy is computed over the induced clusters:
\begin{equation*}
\mathrm{SE}(x) \approx -\sum_{k=1}^{K} \hat{p}(C_k \mid x)\log \hat{p}(C_k \mid x).
\end{equation*}
In our experiments, $\hat{p}(C_k \mid x)$ is estimated from length-normalized sequence log-likelihoods following~\citet{farquhar2024detecting}; for closed-source models where token-level likelihoods are unavailable, cluster frequencies $|C_k|/N$ can be used instead. \textit{Kernel Language Entropy} (KLE)~\cite{nikitin2024kernel} keeps the same $N$ samples and entailment judge but replaces hard clusters with a weighted graph: $W_{ij}$ counts the entailment directions that hold between samples $s_i$ and $s_j$, and uncertainty is the von Neumann entropy of a kernel on this graph. KLE coincides with SE when edges are restricted to cluster membership and needs no likelihoods, so it is black-box by construction; we compute it from the same $N{=}10$ samples and entailment judge as SE, at the same cost (App.~\ref{app:impl}).
\textit{VL-Uncertainty}~\cite{zhang2024vl} replaces stochastic sampling with progressive perturbations of image and text and clusters as SE does; VASE~\cite{liao2025vision} contrasts clusters from original and noisy images to separate linguistic priors from image-conditioned reasoning. All semantic methods apply to closed-source models but need multiple inference runs and an entailment model for clustering or graph construction over the sampled responses.

\section{Experimental Analysis}
\label{sec:experiments}
\subsection{Models and Benchmarks}
\label{sec:models}
We evaluate twelve open-weight MLLMs from two widely adopted model series: InternVL~\cite{chen2024far, wang2025internvl3_5} (generations 2 and 3.5) and QwenVL~\cite{bai2023qwen, wang2024qwen2, bai2025qwen3} (generations 2 and 3), each at three scales (2B; 7--8B; 26B--72B). This setup enables comparison along three axes: model family, generation, and size. We group benchmarks by response length. \textbf{Short.} \emph{ScienceQA}~\cite{lu2022learn} (2,017 validation items, 2--4 options) and \emph{MMMU}~\cite{yue2023mmmu} (825 validation items, 2--9 options) are multiple-choice; \emph{MM-Vet}~\cite{yu2023mm} (218 items) asks for concise free-form answers. \textbf{Sentence-level.} \emph{FSVQA}~\cite{shin2016color} requires complete, grammatical sentences; we use a 200-item subset (FSVQA200) and release the sampled IDs. \textbf{Long.} \emph{LLaVA-Bench}~\cite{liu2023visual} (60 items) pairs images with open-ended instructions; \emph{MM-Vet Long} is MM-Vet with the instruction to give a complete response; we treat it as a separate evaluation set, so all averages run over five benchmarks and six sets.

\subsection{Metrics}
\label{sec:metrics}
We adopt three complementary metrics: AUROC for ranking quality, Coverage at Risk for selective prediction, and CHDA for fixed-threshold deployment; model accuracy is reported for reference.

\textbf{Terminology:} Following~\citet{farquhar2024detecting} and~\citet{zhang2024vl}, the target of all metrics is answer correctness: the positive class is a response judged incorrect against the reference. For multiple-choice and short answers this coincides with hallucination in the usual sense; for sentence- and long-form answers, incorrect responses are heterogeneous (fabricated content, omissions, over-cautious non-answers), so we speak of error detection and reserve hallucination for ungrounded content. We keep the name CHDA for continuity with HDA~\cite{zhang2024vl}.

Let $\mathcal{M}$ denote a multimodal LLM and $\mathcal{U}$ an uncertainty estimator assigning a scalar score $u(x) \in \mathbb{R}$ to each input $x$, where higher values indicate more uncertain (and more likely incorrect) responses.

\textbf{AUROC:} Area under the ROC curve with incorrect responses as the positive class; a threshold-independent measure of how well $u(x)$ ranks correct above incorrect responses.

\textbf{Coverage at Risk (Cov@R):} Following~\citet{whitehead2022reliablevqa}, we recast error detection as selective prediction: the model abstains on samples whose uncertainty exceeds a threshold $\tau$. Coverage $\mathrm{Cov}(\tau)$ is the fraction of answered questions and risk $\mathrm{Risk}(\tau)$ is the error rate on those answers. Cov@$r$ denotes the maximum coverage achievable while keeping risk below budget $r$:
\begin{equation}
    \mathrm{Cov@}r = \max_{\tau \in \mathbb{R}} \mathrm{Cov}(\tau) \quad \text{s.t.} \quad \mathrm{Risk}(\tau) \leq r.
\end{equation}
We report $\mathrm{Cov@}20$ ($r{=}20\%$); higher values indicate more questions answered within the error tolerance, which matters for safety-critical deployment.

\textbf{Calibrated Hallucination Detection Accuracy (CHDA):}
Hallucination detection accuracy (HDA)~\citep{zhang2024vl} scores the decision to flag a response as uncertain when $u(x)>\tau$: a true positive is a flagged incorrect response and a true negative an unflagged correct one,

\begin{equation}
    \mathrm{HDA}(\tau; \mathcal{D}) =
    \frac{\mathrm{TP}(\tau) + \mathrm{TN}(\tau)}{|\mathcal{D}|}.
\end{equation}

HDA assumes a known threshold; CHDA, our deployment-time protocol, selects an optimal threshold in two steps. \textbf{(i) Calibration:} we first calibrate a detection threshold for each model and method on a small subset of up to 60 samples drawn uniformly at random from each of the considered benchmarks with a fixed seed, shared across all models and methods, yielding a pooled, mixed-format calibration set $\mathcal{D}_{\mathrm{cal}}$, and we set $\tau^{\star} = \operatorname*{arg\,max}_{\tau \in \mathbb{R}} \mathrm{HDA}(\tau; \mathcal{D}_{\mathrm{cal}})$. \textbf{(ii) Evaluation:} we report $\mathrm{CHDA}(\mathcal{M}, \mathcal{U}) = \mathrm{HDA}(\tau^{\star}; \mathcal{D}_{\mathrm{eval}})$ on the full evaluation set, applying the same $\tau^{\star}$ unchanged to every benchmark. $\tau^{\star}$ is fit separately for each of the three seed runs and CHDA is averaged over them.

\paragraph{Calibration metrics.} The three previously mentioned metrics depend only on the ranking of the scores. To assess them as probabilities, we map each score to an error probability on $\mathcal{D}_{\text{cal}}$ (quantile binning with isotonic regression; one map per model and method, also used to put the router's branches on a common scale) and report ECE~\cite{guo2017calibration} and the Brier score~\cite{brier1950verification} next to a constant-prediction baseline (App.~\ref{app:ece}).

\subsection{Results}
\label{sec:results}
We structure our analysis around four research questions, addressed in turn below. Across tables, \textbf{Bold} and \underline{underlined} values mark the best and second-best per column. Rows marked with (T=1) use an extra uncertainty estimation-only pass at temperature T = 1.0 on top of the base answer generation at T = 0.1.
\subsubsection{How does response length affect uncertainty estimation?}
\label{length_claims}

\begin{table*}[h!]
\centering
\resizebox{\textwidth}{!}{
\begin{tabular}{lc|ccc|ccc|ccc|ccc}
\toprule
Method & Cost & \multicolumn{3}{c|}{Short} & \multicolumn{3}{c|}{Sentence} & \multicolumn{3}{c|}{Long} & \multicolumn{3}{c}{All benchmarks} \\
 & & AUROC & Cov@20 & CHDA & AUROC & Cov@20 & CHDA & AUROC & Cov@20 & CHDA & AUROC & Cov@20 & CHDA \\
\midrule
\texttt{Model Acc} & -- & \multicolumn{3}{c|}{62.0} & \multicolumn{3}{c|}{68.7} & \multicolumn{3}{c|}{55.6} & \multicolumn{3}{c}{61.0} \\
\midrule
\multicolumn{14}{l}{\textit{Verbalized}} \\
Verbal Abstain & $1\times$ & 0.61 & 0.22 & 66.0 & 0.67 & 0.34 & \textbf{76.0} & 0.54 & 0.00 & 61.9 & 0.60 & 0.17 & 66.3 \\
Verbal Confidence & $1\times$ & 0.58 & 0.33 & 65.5 & 0.62 & 0.39 & 70.8 & 0.61 & 0.14 & 64.7 & 0.60 & 0.28 & 66.1 \\
P(true) & $2\times$ & 0.69 & 0.45 & 64.3 & \textbf{0.72} & \textbf{0.65} & \underline{71.0} & 0.72 & \textbf{0.36} & 68.0 & \underline{0.71} & \underline{0.45} & 66.6 \\
\midrule
\multicolumn{14}{l}{\textit{Token-level}} \\
Avg Entropy & $1\times$ & 0.61 & 0.32 & 68.6 & 0.60 & 0.32 & 62.9 & 0.55 & 0.07 & 59.4 & 0.59 & 0.24 & 64.6 \\
Max Entropy & $1\times$ & 0.62 & 0.32 & 68.2 & 0.60 & 0.19 & 61.1 & 0.56 & 0.08 & 60.3 & 0.60 & 0.22 & 64.4 \\
Avg Entropy ($T{=}1$) & $2\times$ & 0.73 & 0.47 & 68.5 & 0.65 & 0.50 & 66.3 & 0.55 & 0.08 & 60.1 & 0.65 & 0.35 & 65.3 \\
Max Entropy ($T{=}1$) & $2\times$ & \underline{0.75} & \underline{0.49} & \underline{70.0} & 0.64 & 0.47 & 62.3 & 0.57 & 0.09 & 58.7 & 0.67 & 0.36 & 64.9 \\
\midrule
\multicolumn{14}{l}{\textit{Semantic}} \\
SE ($N{=}10$) & $11\times$ & 0.73 & 0.47 & 67.8 & 0.64 & 0.44 & 65.0 & 0.72 & 0.30 & 66.7 & \underline{0.71} & 0.41 & 67.0 \\
SE ($N{=}5$) & $6\times$ & 0.69 & 0.44 & 67.9 & 0.62 & 0.42 & 60.6 & 0.69 & 0.24 & 66.7 & 0.68 & 0.37 & 66.3 \\
VASE ($N{=}5$) & $6\times$ & 0.73 & 0.47 & 67.7 & 0.65 & 0.49 & 65.5 & 0.72 & 0.30 & 68.6 & \underline{0.71} & 0.42 & \underline{67.6} \\
VL-Unc.\ ($N{=}10$) & $11\times$ & 0.72 & 0.48 & 67.1 & 0.61 & 0.40 & 63.1 & \underline{0.73} & 0.31 & \textbf{68.9} & \underline{0.71} & 0.41 & 67.0 \\
KLE ($N{=}10$) & $11\times$ & \textbf{0.76} & \textbf{0.51} & \textbf{73.4} & \underline{0.68} & \underline{0.56} & 63.7 & \textbf{0.74} & \underline{0.33} & \underline{68.8} & \textbf{0.74} & \textbf{0.46} & \textbf{70.2} \\
\midrule[\heavyrulewidth]
\multicolumn{14}{l}{\textit{Length-aware router (Sec.~\ref{sec:router}); excluded from bold / underline}} \\
Entropy ($T{=}1$) / P(true) & $2\times$ & 0.76 & 0.49 & 73.9 & 0.72 & 0.66 & 72.2 & 0.73 & 0.36 & 69.3 & 0.74 & 0.48 & 72.1 \\
\quad + abstain override & $3\times$ & 0.81 & 0.49 & 77.2 & 0.79 & 0.68 & 75.4 & 0.74 & 0.36 & 70.2 & 0.79 & 0.48 & 74.6 \\
\bottomrule
\end{tabular}
}
\vspace{-0.1cm}
    \caption{UE performance by expected answer length and over all benchmarks, averaged over 12 MLLMs and three seeds. Cost: forward passes per sample. Seed std is at most 0.01 AUROC, 0.03 Cov@20 and 1.4 CHDA per cell. \textbf{Bold} / \underline{underlined}: best / second best per column among single methods; router rows unmarked. Token-level entropy at $T{=}1$ nearly matches the best single method on short answers at a fraction of the cost, P(true) leads on sentence-length answers, and KLE gives the best long-form AUROC with P(true) leading coverage. The length-aware router selects the estimator by answer length; with the abstain override it matches or exceeds the best single method's AUROC in every regime and, averaged over all benchmarks, beats every single method on all metrics at $3\times$ cost.
}
\label{tab:answer_length}
\end{table*}

\begin{figure*}[!h]
    \centering
    \includegraphics[width=0.9\textwidth]{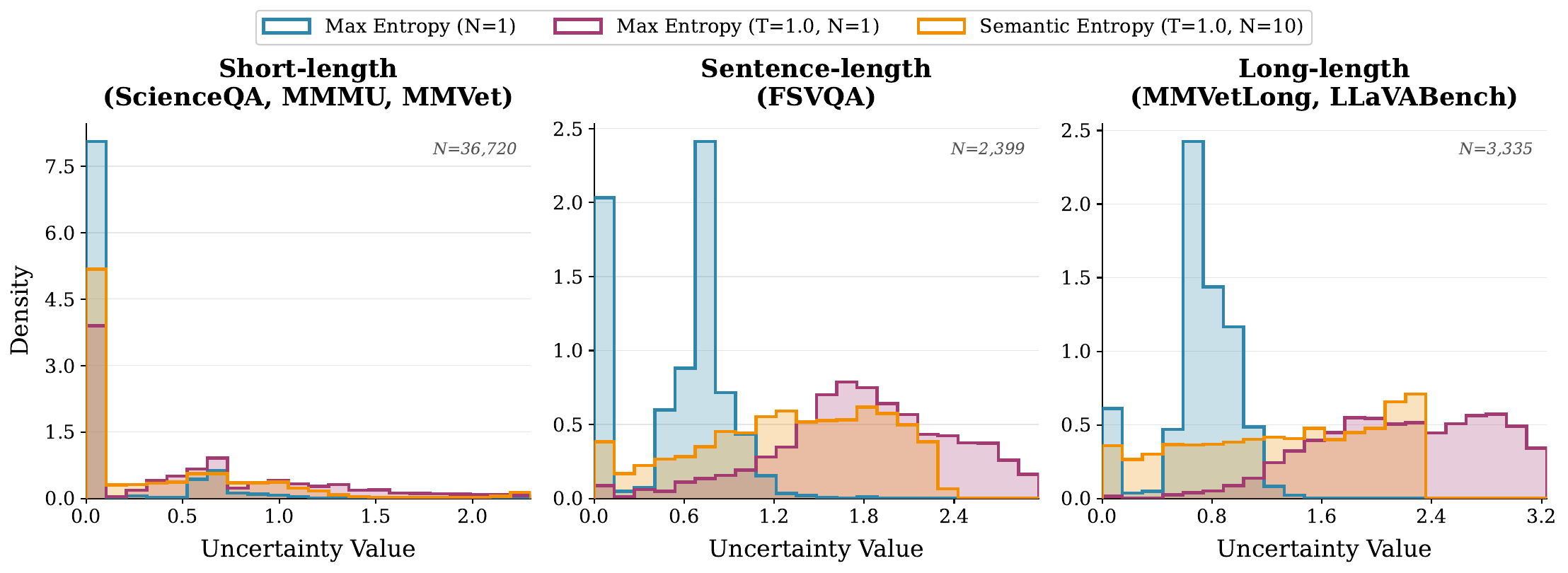}
    \vspace{-0.3cm}
    \caption{Uncertainty distributions across answer-length regimes, aggregated over all MLLMs and seeds (densities normalized per method). Max Entropy at $T{=}0.1$ collapses near zero as low-temperature decoding sharpens the logit distribution; $T{=}1.0$ recovers a broad, informative signal. Semantic Entropy is uninformative on short answers---identical generations fall into a single cluster (Table~\ref{tab:abstain_cluster_merged})---but yields well-spread distributions once lexical variation grows in sentence- and long-length regimes.}
    \label{fig:uncert_distros_grouped}
    \vspace{-0.3cm}
\end{figure*}

Tab.~\ref{tab:answer_length} reports the main results grouped by expected response length (per-benchmark tables in App.~\ref{app:per_benchmark}). No single method dominates; the best family changes with answer length:

\textbf{Short-length regime.} 
Token-level entropy estimated at temperature $T{=}1.0$ nearly matches the best method at a fraction of the cost: Max Entropy ($T{=}1$) reaches 0.75 AUROC, 0.49 Cov@20 and 70.0 CHDA with one extra forward pass ($2\times$), against 0.76 / 0.51 / 73.4 for KLE at $11\times$ (Tab.~\ref{tab:answer_length}). Using the common low inference temperature of 0.1 instead substantially degrades token-level estimates (0.62 AUROC for the same statistic). At low temperatures, the logit distribution becomes overly sharp, collapsing most token probabilities toward zero and reducing the usefulness of entropy as an uncertainty signal. As shown in Fig.~\ref{fig:uncert_distros_grouped}, entropy computed at $ T=1.0$  produces a much broader and more informative uncertainty distribution.

The same figure exposes a key limitation of semantic methods (with SE, N=10, as a representative). Even at temperature 1.0, models often generate identical short responses, collapsing generations into a single cluster and driving inter-cluster entropy toward $0.0$. Tab.~\ref{tab:abstain_cluster_merged} confirms this: multiple-choice answers fall into 1.4--2.3 clusters on average, free-form answers into 5.2--5.6.

\textbf{Sentence-length regime.} On FSVQA200 the verbalized family leads: P(true) has the best AUROC and coverage (0.72 / 0.65) and Verbal Abstain the best CHDA (76.0), five points above P(true), at $1\times$--$2\times$ cost. We hypothesize that self-evaluation works well on FSVQA200 because each question has a single underlying fact to be uncertain about: a single \texttt{UNKNOWN} token, or a single true/false judgment of the answer, can express the model's uncertainty about that one fact. By contrast, multiple-choice questions bias the model to commit to one of the given options, and long-form questions spread uncertainty across many sub-claims that a single verdict cannot localize. Abstention rates support this (Tab.~\ref{tab:abstain_cluster_merged}): they are highest on MM-Vet and FSVQA200 and sharply lower on multiple-choice and long-form questions. Semantic methods follow (KLE 0.68 AUROC, 0.56 Cov@20) where richer lexical variation results in more clusters and broader uncertainty distributions (Fig.~\ref{fig:uncert_distros_grouped}), but at $6$--$11\times$ cost, and token-level methods lose the advantage they hold on short answers (Max Entropy $T{=}1$: 0.64).

\textbf{Long-length regime.} On long-form generation, semantic methods have the best AUROC (KLE 0.74) while P(true) leads coverage (0.36 vs.\ 0.33 for KLE; Tab.~\ref{tab:answer_length}). The diverse responses elicited at $T{=}1.0$ spread over many semantic clusters (Tab.~\ref{tab:abstain_cluster_merged}), giving semantic methods a stable and informative signal, whereas token-level methods lose ground because the correctness of a long answer depends far less on the entropy of any single token. The semantic advantage is confined to small and medium models: at 26--72B, P(true) overtakes every sampling-based method (AUROC 0.76 vs.\ 0.72 for KLE; App.~\ref{app:size_longform}).

App.~\ref{sec:qualitative} shows examples and failure cases per regime (Figs.~\ref{fig:qualitative_short}--\ref{fig:qualitative_long_fail}).

\begin{table}[t]
\centering
\scriptsize
\begin{tabular}{llcc}
\toprule
Answer Type & Benchmark & Mean Clusters & Abst.\ (\%) \\
\midrule
Short & ScienceQA & 1.42 $\pm$0.37 & 4.83 $\pm$3.63 \\
 & MMMU & 2.28 $\pm$0.58 & 6.20 $\pm$5.99 \\
 & MM-Vet & 5.62 $\pm$1.44 & 16.79 $\pm$8.05 \\
 & \emph{All Short} & 3.11 $\pm$2.03 & 9.27 $\pm$8.16 \\
\midrule
Sentence & FSVQA200 & 5.24 $\pm$1.07 & 13.32 $\pm$7.93 \\
 & \emph{All Sentence} & 5.24 $\pm$1.07 & 13.32 $\pm$7.93 \\
\midrule
Long & MM-Vet Long & 5.48 $\pm$1.37 & 3.31 $\pm$3.82 \\
 & LLaVA-Bench & 5.35 $\pm$1.16 & 5.60 $\pm$6.04 \\
 & \emph{All Long} & 5.41 $\pm$1.27 & 4.46 $\pm$5.18 \\
\bottomrule
\end{tabular}
\vspace{-0.1cm}
\caption{Average semantic cluster counts of SE ($T{=}1.0$, $N{=}10$) and mean verbal abstention rates across 12 MLLMs, three seeds and benchmarks. Lower cluster diversity on multiple-choice benchmarks explains SE's weaker performance there; single-word and sentence-level tasks show the highest abstention rates.}
\vspace{-0.5cm}
\label{tab:abstain_cluster_merged}
\end{table}


\subsubsection{How do model generation and scale affect uncertainty estimation? }

Tab.~\ref{tab:vlm_method_avgs} reports the effect of MLLM family, scale, and generation on uncertainty estimation for the overall-best method of each family (P(true), Max Entropy ($T{=}1$) and KLE), averaged across all benchmarks; Tab.~\ref{tab:avg_vlms_benchmarks} in App.~\ref{app:extra_tables} averages the three methods per model.
 Both scaling parameters and moving to newer generations consistently improve base task accuracy (e.g., Qw2-VL 2B$\to$7B: 48.0 $\to$ 61.2; IVL2$\to$IVL3.5 at 8B: 57.3 $\to$ 67.4). These gains, however, translate unevenly into uncertainty estimation quality:

\textbf{Method choice drives AUROC.} AUROC varies between 0.59 and 0.74 across uncertainty-estimation methods (Tab.~\ref{tab:answer_length}), but only between 0.67 and 0.75 across the twelve models (Tab.~\ref{tab:avg_vlms_benchmarks}). The estimation method is therefore the primary determinant of ranking quality, while model family, scale, and generation have limited effect on discriminative performance.

\textbf{Model choice dominates coverage.} Coverage, in contrast, is driven primarily by the base model. Across base models, Cov@20 spans 0.41 points (0.20 to 0.61; Tab.~\ref{tab:avg_vlms_benchmarks}), well above the 0.29-point spread across estimation methods (0.17 to 0.46; Tab.~\ref{tab:answer_length}). This pattern largely reflects differences in base accuracy (48.0\% to 70.7\%): stronger models achieve higher selective coverage under the same confidence-thresholding regime. Within a family, scale (e.g., Qw3-VL 2B$\to$8B: 0.28 $\to$ 0.44; Qw2-VL 7B$\to$72B: 0.44 $\to$ 0.58) and newer generations (e.g., IVL2$\to$IVL3.5 at 8B: 0.36 $\to$ 0.52) yield gains in both accuracy and coverage, with minimal impact on AUROC.

\textbf{CHDA improves with model quality, but modestly.} A single threshold calibrated once transfers better for stronger models: CHDA ranges from 62.0 to 72.2, tracking scale and generation (e.g., IVL3.5 2B$\to$8B: 65.3$\to$70.8; IVL2$\to$IVL3.5 at 8B: 66.7$\to$70.8), and the four model generations converge to 68.5--72.2 at their largest size.

\textbf{Self-evaluation scales; sampling does not.} The one exception to the flat AUROC is P(true): averaged over families its AUROC rises from 0.66 at 2B to 0.71 at 7--8B and 0.75 at 26--72B, whereas KLE stays at 0.73--0.75 and Max Entropy at 0.67--0.68 (Tab.~\ref{tab:vlm_method_avgs}). KLE therefore has the best AUROC for every model up to 8B, ties P(true) on Qwen3-VL-8B, InternVL3.5-38B and Qwen2-VL-72B, and is overtaken on Qwen3-VL-32B (0.68 vs 0.75). Larger models know better when they are wrong, and read this off their own answer more cheaply than through resampling.

Together, these observations suggest that practitioners should prioritize \textit{method selection} for ranking-quality use cases, \textit{model selection} for selective prediction, and \textit{both jointly} for fixed-threshold deployment; at 26B and above, P(true) becomes a cheap substitute for sampling-based methods.

\subsubsection{Are UE methods compatible with safety requirements?}
\label{sec:safety}
Even at a loose 20\% risk budget no method reaches deployable coverage: the best single-method Cov@20 is 0.51 on short answers (KLE), 0.65 on sentence-length answers and 0.36 on long-form answers (both P(true)), and the router does not change this (0.49 / 0.68 / 0.36; Tab.~\ref{tab:answer_length}).

To isolate the UE contribution from base-model accuracy, observe that a hypothetical perfect UE method --- one that ranks every correct answer above every wrong one --- achieves $\text{Cov@}r = a/(1-r)$, where $a$ is base accuracy: it answers all correct samples plus as many wrong samples as the risk budget allows. With $r{=}20\%$ and observed accuracies of 0.56--0.69 across regimes (Tab.~\ref{tab:answer_length}), the perfect-UE ceilings are 0.70 (long), 0.78 (short) and 0.86 (sentence). Current methods deliver 0.36, 0.51 and 0.65 respectively, that is 52--76\% of the achievable coverage, and the largest models reach 0.53--0.65 with their best estimator (Tab.~\ref{tab:vlm_method_avgs}).
The gap comes from confidently wrong answers that no estimator ranks high (AUROC 0.72--0.76 for the best methods), and recalibration does not close it: after the map of Sec.~\ref{sec:metrics} every method has an ECE of 0.11--0.13, the same as a constant prediction of the error rate, while the Brier score ranks the methods like AUROC (Tab.~\ref{tab:appendix_ece}, App.~\ref{app:ece}). Better ranking of correct over incorrect answers, not calibration or base accuracy alone, is therefore the lever for safe deployment.

\subsubsection{How do individual UE methods compare within each family?}
\label{sec:model_by_model}
\noindent\textbf{Verbal methods.} The effectiveness of \textbf{Verbal Abstain} varies considerably across models. Newer and larger MLLMs abstain more reliably: on Qwen2-VL-72B and Qwen3-VL-8B it attains the highest all-benchmark CHDA of any method (75.1 and 73.2, compared with 71.9 and 72.1 for KLE). Smaller, older models rarely abstain; on ScienceQA, for instance, InternVL2-2B attains only 15.9 CHDA, compared with 91.4 for Qwen2-VL-72B (Tab.~\ref{tab:scienceqa}). \textbf{Verbal Confidence} is among the weakest estimators: 80.7\% of parseable predictions carry the maximum confidence regardless of correctness and 9.6\% of responses fail to parse (2.7--25.9\% across models), and scale resolves neither (Tab.~\ref{tab:verbal-conf-stats}, App.~\ref{app:extra_tables}). \textbf{P(true)} reads the model's self-assessment from its output distribution instead of generated text, avoiding the parse failures and overconfidence of Verbal Confidence (0.71 vs.\ 0.60 average AUROC). It is the strongest method on sentence-length answers and, for larger models, on long-form answers (Tabs.~\ref{tab:answer_length}, \ref{tab:size_longform}). Its threshold, however, transfers less effectively across benchmarks (short-form CHDA 64.3), and like Verbal Confidence it requires recalibration to yield usable probabilities (raw ECE 0.25 and 0.29; App.~\ref{app:ece}).


\noindent\textbf{Token-level methods} are computationally inexpensive and, when estimated at $T{=}1.0$, competitive with costlier alternatives (Tabs.~\ref{tab:answer_length}, \ref{tab:vlm_method_avgs}), rendering them well suited to resource-constrained settings. \textbf{Max Entropy} outperforms \textbf{Avg Entropy}, indicating that the signal is concentrated in the most uncertain token. The NLL variants closely mirror their entropy counterparts (App.~\ref{app:per_benchmark}).

\begin{table}[t]
\centering
\setlength{\tabcolsep}{3pt}
\scriptsize
\begin{tabular}{ll|ccc|ccc|ccc}
\toprule
Family & Size & \multicolumn{3}{c|}{AUROC} & \multicolumn{3}{c|}{CHDA} & \multicolumn{3}{c}{Cov@20} \\
 & & Ent & P(t) & KLE & Ent & P(t) & KLE & Ent & P(t) & KLE \\
\midrule
IVL2 & 2B & 0.68 & 0.61 & 0.76 & 64.8 & 61.9 & 70.5 & 0.29 & 0.28 & 0.37 \\
 & 8B & 0.69 & 0.71 & 0.77 & 63.6 & 65.6 & 71.1 & 0.34 & 0.37 & 0.37 \\
 & 26B & 0.70 & 0.74 & 0.75 & 64.8 & 69.9 & 70.6 & 0.37 & 0.50 & 0.53 \\
\midrule
IVL3.5 & 2B & 0.68 & 0.71 & 0.72 & 62.8 & 66.7 & 66.4 & 0.33 & 0.37 & 0.37 \\
 & 8B & 0.69 & 0.71 & 0.76 & 70.0 & 67.9 & 74.6 & 0.43 & 0.54 & 0.60 \\
 & 38B & 0.66 & 0.74 & 0.74 & 71.0 & 72.1 & 73.7 & 0.55 & 0.64 & 0.64 \\
\midrule
Qw2-VL & 2B & 0.65 & 0.65 & 0.71 & 60.8 & 62.1 & 65.5 & 0.16 & 0.19 & 0.25 \\
 & 7B & 0.69 & 0.70 & 0.75 & 58.6 & 62.7 & 68.9 & 0.37 & 0.46 & 0.48 \\
 & 72B & 0.71 & 0.77 & 0.77 & 68.9 & 69.6 & 71.9 & 0.46 & 0.64 & 0.65 \\
\midrule
Qw3-VL & 2B & 0.66 & 0.66 & 0.73 & 58.4 & 62.3 & 65.3 & 0.22 & 0.33 & 0.29 \\
 & 8B & 0.63 & 0.72 & 0.72 & 66.1 & 67.1 & 72.1 & 0.34 & 0.49 & 0.48 \\
 & 32B & 0.63 & 0.75 & 0.68 & 69.4 & 71.9 & 72.3 & 0.40 & 0.61 & 0.45 \\
\midrule
\multicolumn{2}{l|}{\emph{Average}} & 0.67 & 0.71 & 0.74 & 64.9 & 66.6 & 70.2 & 0.36 & 0.45 & 0.46 \\
\bottomrule
\end{tabular}
\vspace{-0.2cm}
\caption{Cross-benchmark averages (three seeds) per model for the best method of each UE family: Max Entropy ($T{=}1$; Ent), P(true) (P(t)) and KLE ($N{=}10$). Seed standard deviation is at most 0.01 AUROC, 1.7 CHDA and 0.05 Cov@20 per cell. KLE has the best AUROC and CHDA for 11 of 12 models; P(true) is second and closes the gap at the largest scale, while Max Entropy stays 0.04--0.09 AUROC below KLE.}
\vspace{-0.3cm}
\label{tab:vlm_method_avgs}
\end{table}

\noindent\textbf{Semantic methods} are the most consistent top performers across regimes (Tab.~\ref{tab:answer_length}), and \textbf{KLE} is the strongest of them (Tabs.~\ref{tab:answer_length}, \ref{tab:vlm_method_avgs}). Its gain over SE comes from items on which hard clustering saturates: free-form answers often spread the $N{=}10$ samples over five to six clusters on average (Tab.~\ref{tab:abstain_cluster_merged}), where cluster-frequency entropy stops ranking items but graded entailment weights still do; on multiple choice the samples collapse into one or two clusters and the two coincide. Neither perturbations nor larger sample budgets improve consistently over plain temperature sampling (Tabs.~\ref{tab:SE_perturbation_ablation}, \ref{tab:SE_N_ablation}, App.~\ref{app:extra_tables}).

\subsection{A length-aware recipe}
\label{sec:router}
Tab.~\ref{tab:answer_length} suggests a simple recipe: choose the estimator from the length of the answer the model has already produced. We split answers into three length bins that mirror the regimes of Tab.~\ref{tab:answer_length} ($\leq$5, 6--60 and $>$60 words), use Max Entropy ($T{=}1$) for the short-length bin and P(true) for the other two, and fit one threshold and one probability map per bin on the same $\mathcal{D}_{\text{cal}}$ as for single methods. Using SE or KLE for long answers changes AUROC by at most 0.01 while having a higher cost (Tab.~\ref{tab:router_variants}, App.~\ref{app:router}).

Verbal abstention provides a signal that is complementary to score-based estimators. As a binary indicator it is a weak ranker in isolation (0.60 AUROC averaged over all benchmarks), yet the responses on which a model abstains include errors that score-based estimators fail to rank highly. We therefore introduce an optional \emph{abstain override}: an additional pass under the abstention prompt, after which any response on which the model abstains is flagged as uncertain irrespective of the routed score. This improves the router in every regime, most markedly on sentence-length answers, where abstention is frequent (Tab.~\ref{tab:answer_length}).

The performance of the proposed router is reported in the last two rows of Tab.~\ref{tab:answer_length}. Routing alone (2$\times$: decoding plus either the $T{=}1$ or the P(true) pass) is within 0.01 AUROC of the best single method in every regime. With the override (3$\times$), it matches or exceeds the best single method's AUROC in every regime, with the largest gains on short (0.81 vs.\ 0.76) and sentence-length answers (0.79 vs.\ 0.72). Averaged over all benchmarks, it outperforms every single method on all three metrics (0.79 / 0.48 / 74.6 AUROC / Cov@20 / CHDA vs.\ 0.74 / 0.46 / 70.2 for KLE) at about a quarter of KLE's 11$\times$ cost.

Each part of the recipe fixes a known weakness. Routing avoids semantic methods on short answers, where all samples agree, and token entropy on long answers, where single tokens say little about correctness. The override avoids a score based decision on (sentence-length) samples that the model flags for abstaining. Using SE or KLE instead of P(true) for long answers costs 5--6$\times$ and does not improve the performance (Tab.~\ref{tab:router_variants}, App.~\ref{app:router}).

\section{Conclusion and  Main Takeaways}
We organized training-free UE methods for MLLMs into three families and compared them across model series, scales, benchmarks and answer-length regimes. (1) Answer length decides the winner: token-level entropy on short answers at a fraction of the cost, verbalized methods on sentence-length responses, semantic methods on long-form generation. (2) Token-level uncertainty should be estimated at a higher temperature than the one used at decoding: a second pass at $T{=}1.0$ recovers the signal that low-temperature inference collapses. (3) Self-evaluation scales with model size: P(true) is the only estimator whose ranking improves from 2B to 72B, overtaking sampling-based methods on long-form answers at 26B and above, and abstention improves with scale, while numeric verbal confidence stays unreliable. (4) Routing each answer by length to token entropy or P(true), with an abstention override, beats every single method averaged over all benchmarks and models at 3$\times$ cost.
(5) No method reaches safe-deployment coverage: Cov@20 improves with scale but stays well below the ceiling set by base accuracy even at 72B  model parameters.

\section*{Limitations}

\paragraph{Models and access.} We evaluate open-weight models only: four model generations from two series (InternVL and Qwen-VL), spanning 2B to 72B parameters. Token-level methods and P(true) read the model's output distribution, so the three families can only be compared where logits are available. Verbalized abstention and confidence need only the generated text, and the semantic methods can run black-box with cluster frequencies in place of likelihoods, but the length-aware router does not transfer as is, since its short-answer branch relies on token entropy. Black-box substitutes on proprietary models and further model families are left to future work.

\paragraph{Scope.} We study MLLMs in isolation, motivated by the lack of in-depth uncertainty analyses in this setting; a matched comparison with unimodal LLMs is left for future work. We restrict ourselves to training-free estimators for their ease of deployment; training-based approaches may be more reliable where the methods studied here fall short. We use general-purpose benchmarks and standard methods; domain-specific evaluations may lead to different conclusions.

\paragraph{Evaluation design.} All sampling-based semantic methods (SE, VASE, VL-Uncertainty, KLE) share one entailment model (Qwen2.5-7B-Instruct), so their errors are correlated and we do not ablate the judge. KLE is computed offline from the stored SE samples, which makes the KLE--SE comparison paired but means KLE did not receive independent generations. Benchmark averages weight every benchmark equally, so the 60-item LLaVA-Bench weighs as much as the 218-item MM-Vet Long in the long-form numbers and as much as the 2,017-item ScienceQA in all-benchmark averages; it is also the only benchmark whose seed standard deviation of accuracy exceeds one point.

\paragraph{Correctness labels.} Labels are binary, response-level and produced by an LLM judge. For long-form answers they do not separate fabricated content from omissions, over-cautious non-answers or partially correct responses, and a single wrong claim may mark the whole response as incorrect. A claim-level error taxonomy, and estimators that localize which part of a response is unreliable, are left to future work.

\section*{Ethical Considerations}
All models and benchmarks are publicly released research artifacts used under their licenses (App.~\ref{sec:license}); the images shown in Figs.~3--8 are taken from these benchmarks and depict no private individuals beyond what the benchmarks themselves publish. Correctness labels are produced by a commercial LLM judge, whose errors are shared across all compared estimators but may encode its own biases. Uncertainty estimates studied here are not a safety guarantee: as \S4.3.3 shows, no method reaches deployable coverage, and they should complement, not replace, human oversight in high-stakes settings.

\bibliography{custom}

@inproceedings{borszukovszki-etal-2025-know,
    title = "Know What You do Not Know: Verbalized Uncertainty Estimation Robustness on Corrupted Images in Vision-Language Models",
    author = "Borszukovszki, Mirko  and
      De Jong, Ivo Pascal  and
      Valdenegro-Toro, Matias",
    editor = "Cao, Trista  and
      Das, Anubrata  and
      Kumarage, Tharindu  and
      Wan, Yixin  and
      Krishna, Satyapriya  and
      Mehrabi, Ninareh  and
      Dhamala, Jwala  and
      Ramakrishna, Anil  and
      Galystan, Aram  and
      Kumar, Anoop  and
      Gupta, Rahul  and
      Chang, Kai-Wei",
    booktitle = "Proceedings of the 5th Workshop on Trustworthy NLP (TrustNLP 2025)",
    month = may,
    year = "2025",
    address = "Albuquerque, New Mexico",
    publisher = "Association for Computational Linguistics",
    url = "https://aclanthology.org/2025.trustnlp-main.16/",
    doi = "10.18653/v1/2025.trustnlp-main.16",
    pages = "247--265",
    ISBN = "979-8-89176-233-6"
}

@inproceedings{xuan-etal-2025-seeing,
    title = "Seeing is Believing, but How Much? A Comprehensive Analysis of Verbalized Calibration in Vision-Language Models",
    author = "Xuan, Weihao  and
      Zeng, Qingcheng  and
      Qi, Heli  and
      Wang, Junjue  and
      Yokoya, Naoto",
    editor = "Christodoulopoulos, Christos  and
      Chakraborty, Tanmoy  and
      Rose, Carolyn  and
      Peng, Violet",
    booktitle = "Proceedings of the 2025 Conference on Empirical Methods in Natural Language Processing",
    month = nov,
    year = "2025",
    address = "Suzhou, China",
    publisher = "Association for Computational Linguistics",
    url = "https://aclanthology.org/2025.emnlp-main.74/",
    doi = "10.18653/v1/2025.emnlp-main.74",
    pages = "1408--1450",
    ISBN = "979-8-89176-332-6"
}

@article{wang2025vision,
  title={Are Vision Language Models Robust to Uncertain Inputs?},
  author={Wang, Xi and Nalisnick, Eric},
  journal={arXiv preprint arXiv:2505.11804},
  year={2025}
}

@inproceedings{groot-valdenegro-toro-2024-overconfidence,
    title = "Overconfidence is Key: Verbalized Uncertainty Evaluation in Large Language and Vision-Language Models",
    author = "Groot, Tobias  and
      Valdenegro-Toro, Matias",
    editor = "Ovalle, Anaelia  and
      Chang, Kai-Wei  and
      Cao, Yang Trista  and
      Mehrabi, Ninareh  and
      Zhao, Jieyu  and
      Galstyan, Aram  and
      Dhamala, Jwala  and
      Kumar, Anoop  and
      Gupta, Rahul",
    booktitle = "Proceedings of the 4th Workshop on Trustworthy Natural Language Processing (TrustNLP 2024)",
    month = jun,
    year = "2024",
    address = "Mexico City, Mexico",
    publisher = "Association for Computational Linguistics",
    url = "https://aclanthology.org/2024.trustnlp-1.13/",
    doi = "10.18653/v1/2024.trustnlp-1.13",
    pages = "145--171"
}

@article{zhao2025object,
  title={Object-Level Verbalized Confidence Calibration in Vision-Language Models via Semantic Perturbation},
  author={Zhao, Yunpu and Zhang, Rui and Xiao, Junbin and Hou, Ruibo and Guo, Jiaming and Zhang, Zihao and Hao, Yifan and Chen, Yunji},
  journal={arXiv preprint arXiv:2504.14848},
  year={2025}
}

@inproceedings{kriz2025prompt4trust,
  title={{Prompt4Trust}: A Reinforcement Learning Prompt Augmentation Framework for Clinically-Aligned Confidence Calibration in Multimodal Large Language Models},
  author={Kriz, Anita and Janes, Elizabeth Laura and Shen, Xing and Arbel, Tal},
  booktitle={Proceedings of the IEEE/CVF International Conference on Computer Vision (ICCV)},
  pages={1320--1329},
  year={2025}
}

@article{zhang2024vl,
  title={{VL-Uncertainty}: Detecting Hallucination in Large Vision-Language Model via Uncertainty Estimation},
  author={Zhang, Ruiyang and Zhang, Hu and Zheng, Zhedong},
  journal={arXiv preprint arXiv:2411.11919},
  year={2024}
}

@article{farquhar2024detecting,
  title={Detecting Hallucinations in Large Language Models Using Semantic Entropy},
  author={Farquhar, Sebastian and Kossen, Jannik and Kuhn, Lorenz and Gal, Yarin},
  journal={Nature},
  volume={630},
  number={8017},
  pages={625--630},
  year={2024}
}

@inproceedings{liao2025vision,
  title={Vision-Amplified Semantic Entropy for Hallucination Detection in Medical Visual Question Answering},
  author={Liao, Zehui and Hu, Shishuai and Zou, Ke and Fu, Huazhu and Zhen, Liangli and Xia, Yong},
  booktitle={International Conference on Medical Image Computing and Computer-Assisted Intervention ({MICCAI})},
  pages={669--679},
  year={2025},
  publisher={Springer}
}

@inproceedings{li-etal-2024-reference,
    title = "Reference-free Hallucination Detection for Large Vision-Language Models",
    author = "Li, Qing  and
      Geng, Jiahui  and
      Lyu, Chenyang  and
      Zhu, Derui  and
      Panov, Maxim  and
      Karray, Fakhri",
    editor = "Al-Onaizan, Yaser  and
      Bansal, Mohit  and
      Chen, Yun-Nung",
    booktitle = "Findings of the Association for Computational Linguistics: EMNLP 2024",
    month = nov,
    year = "2024",
    address = "Miami, Florida, USA",
    publisher = "Association for Computational Linguistics",
    url = "https://aclanthology.org/2024.findings-emnlp.262/",
    doi = "10.18653/v1/2024.findings-emnlp.262",
    pages = "4542--4551"
}

@article{fang2024enhancing,
  title={Enhancing Vision-Language Model Reliability with Uncertainty-Guided Dropout Decoding},
  author={Fang, Yixiong and Yang, Ziran and Chen, Zhaorun and Zhao, Zhuokai and Zhou, Jiawei},
  journal={arXiv preprint arXiv:2412.06474},
  year={2024}
}

@article{seo2025epistemic,
  title={On Epistemic Uncertainty of Visual Tokens for Object Hallucinations in Large Vision-Language Models},
  author={Seo, Hoigi and Kang, Dong Un and Cho, Hyunjin and Lee, Joohoon and Chun, Se Young},
  journal={arXiv preprint arXiv:2510.09008},
  year={2025}
}

@inproceedings{whitehead2022reliablevqa,
  title={Reliable Visual Question Answering: Abstain Rather Than Answer Incorrectly},
  author={Whitehead, Spencer and Petryk, Suzanne and Shakib, Vedaad and Gonzalez, Joseph and Darrell, Trevor and Rohrbach, Anna and Rohrbach, Marcus},
  booktitle={Proceedings of the European Conference on Computer Vision (ECCV)},
  pages={148--166},
  year={2022}
}

@inproceedings{lu2022learn,
  title={Learn to Explain: Multimodal Reasoning via Thought Chains for Science Question Answering},
  author={Lu, Pan and Mishra, Swaroop and Xia, Tony and Qiu, Liang and Chang, Kai-Wei and Zhu, Song-Chun and Tafjord, Oyvind and Clark, Peter and Kalyan, Ashwin},
  booktitle={Advances in Neural Information Processing Systems},
  volume={35},
  year={2022}
}

@article{yu2023mm,
  title={{MM-Vet}: Evaluating Large Multimodal Models for Integrated Capabilities},
  author={Yu, Weihao and Yang, Zhengyuan and Li, Linjie and Wang, Jianfeng and Lin, Kevin and Liu, Zicheng and Wang, Xinchao and Wang, Lijuan},
  journal={arXiv preprint arXiv:2308.02490},
  year={2023}
}

@inproceedings{liu2023visual,
  title={Visual Instruction Tuning},
  author={Liu, Haotian and Li, Chunyuan and Wu, Qingyang and Lee, Yong Jae},
  booktitle={Advances in Neural Information Processing Systems},
  volume={36},
  pages={34892--34916},
  year={2023}
}

@article{bai2024hallucination,
  title={Hallucination of Multimodal Large Language Models: A Survey},
  author={Bai, Zechen and Wang, Pichao and Xiao, Tianjun and He, Tong and Han, Zongbo and Zhang, Zheng and Shou, Mike Zheng},
  journal={arXiv preprint arXiv:2404.18930},
  year={2024}
}

@article{liu2024survey,
  title={A Survey on Hallucination in Large Vision-Language Models},
  author={Liu, Hanchao and Xue, Wenyuan and Chen, Yifei and Chen, Dapeng and Zhao, Xiutian and Wang, Ke and Hou, Liping and Li, Rongjun and Peng, Wei},
  journal={arXiv preprint arXiv:2402.00253},
  year={2024}
}

@inproceedings{sahoo-etal-2024-comprehensive,
    title = "A Comprehensive Survey of Hallucination in Large Language, Image, Video and Audio Foundation Models",
    author = "Sahoo, Pranab  and
      Meharia, Prabhash  and
      Ghosh, Akash  and
      Saha, Sriparna  and
      Jain, Vinija  and
      Chadha, Aman",
    editor = "Al-Onaizan, Yaser  and
      Bansal, Mohit  and
      Chen, Yun-Nung",
    booktitle = "Findings of the Association for Computational Linguistics: EMNLP 2024",
    month = nov,
    year = "2024",
    address = "Miami, Florida, USA",
    publisher = "Association for Computational Linguistics",
    url = "https://aclanthology.org/2024.findings-emnlp.685/",
    doi = "10.18653/v1/2024.findings-emnlp.685",
    pages = "11709--11724"
}

@article{chen2026survey,
  title={A Survey of Multimodal Hallucination Evaluation and Detection},
  author={Chen, Zhiyuan and Min, Yuecong and Zhang, Jie and Yan, Bei and Wang, Jiahao and Wang, Xiaozhen and Shan, Shiguang},
  journal={International Journal of Computer Vision},
  volume={134},
  number={3},
  pages={131},
  year={2026}
}

@inproceedings{yue2023mmmu,
  title={{MMMU}: A Massive Multi-discipline Multimodal Understanding and Reasoning Benchmark for Expert {AGI}},
  author={Yue, Xiang and Ni, Yuansheng and Zhang, Kai and Zheng, Tianyu and Liu, Ruoqi and Zhang, Ge and Stevens, Samuel and Jiang, Dongfu and Ren, Weiming and Sun, Yuxuan and Wei, Cong and Yu, Botao and Yuan, Ruibin and Sun, Renliang and Yin, Ming and Zheng, Boyuan and Yang, Zhenzhu and Liu, Yibo and Huang, Wenhao and Sun, Huan and Su, Yu and Chen, Wenhu},
  booktitle={Proceedings of the IEEE/CVF Conference on Computer Vision and Pattern Recognition (CVPR)},
  pages={9556--9567},
  year={2024}
}

@article{wang2025internvl3_5,
  title={{InternVL3.5}: Advancing Open-Source Multimodal Models in Versatility, Reasoning, and Efficiency},
  author={Wang, Weiyun and Gao, Zhangwei and Gu, Lixin and Pu, Hengjun and Cui, Long and Wei, Xingguang and Liu, Zhaoyang and Jing, Linglin and Ye, Shenglong and Shao, Jie and others},
  journal={arXiv preprint arXiv:2508.18265},
  year={2025}
}

@article{chen2024far,
  title={How Far Are We to {GPT-4V}? Closing the Gap to Commercial Multimodal Models with Open-Source Suites},
  author={Chen, Zhe and Wang, Weiyun and Tian, Hao and Ye, Shenglong and Gao, Zhangwei and Cui, Erfei and Tong, Wenwen and Hu, Kongzhi and Luo, Jiapeng and Ma, Zheng and others},
  journal={Science China Information Sciences},
  volume={67},
  number={12},
  pages={220101},
  year={2024}
}

@article{shin2016color,
  title={The Color of the Cat is Gray: 1 Million Full-Sentences Visual Question Answering ({FSVQA})},
  author={Shin, Andrew and Ushiku, Yoshitaka and Harada, Tatsuya},
  journal={arXiv preprint arXiv:1609.06657},
  year={2016}
}

@inproceedings{brown2020language,
  title={Language Models are Few-Shot Learners},
  author={Brown, Tom and Mann, Benjamin and Ryder, Nick and Subbiah, Melanie and Kaplan, Jared D. and Dhariwal, Prafulla and Neelakantan, Arvind and Shyam, Pranav and Sastry, Girish and Askell, Amanda and others},
  booktitle={Advances in Neural Information Processing Systems},
  volume={33},
  pages={1877--1901},
  year={2020}
}

@article{zhao2023survey,
  title={A Survey of Large Language Models},
  author={Zhao, Wayne Xin and Zhou, Kun and Li, Junyi and Tang, Tianyi and Wang, Xiaolei and Hou, Yupeng and Min, Yingqian and Zhang, Beichen and Zhang, Junjie and Dong, Zican and others},
  journal={arXiv preprint arXiv:2303.18223},
  year={2023}
}

@inproceedings{li2025survey,
  title={A Survey of State of the Art Large Vision Language Models: Benchmark Evaluations and Challenges},
  author={Li, Zongxia and Wu, Xiyang and Du, Hongyang and Liu, Fuxiao and Nghiem, Huy and Shi, Guangyao},
  booktitle={Proceedings of the IEEE/CVF Conference on Computer Vision and Pattern Recognition (CVPR) Workshops},
  pages={1587--1606},
  year={2025}
}

@article{ghosh2024exploring,
  title={Exploring the Frontier of Vision-Language Models: A Survey of Current Methodologies and Future Directions},
  author={Ghosh, Akash and Acharya, Arkadeep and Saha, Sriparna and Jain, Vinija and Chadha, Aman},
  journal={arXiv preprint arXiv:2404.07214},
  year={2024}
}

@article{eloundou2024gpts,
  title={{GPTs} are {GPTs}: Labor Market Impact Potential of {LLMs}},
  author={Eloundou, Tyna and Manning, Sam and Mishkin, Pamela and Rock, Daniel},
  journal={Science},
  volume={384},
  number={6702},
  pages={1306--1308},
  year={2024}
}

@article{wang2024survey,
  title={A Survey on Large Language Model Based Autonomous Agents},
  author={Wang, Lei and Ma, Chen and Feng, Xueyang and Zhang, Zeyu and Yang, Hao and Zhang, Jingsen and Chen, Zhiyuan and Tang, Jiakai and Chen, Xu and Lin, Yankai and others},
  journal={Frontiers of Computer Science},
  volume={18},
  number={6},
  pages={186345},
  year={2024}
}

@article{yehudai2025survey,
  title={Survey on Evaluation of {LLM}-based Agents},
  author={Yehudai, Asaf and Eden, Lilach and Li, Alan and Uziel, Guy and Zhao, Yilun and Bar-Haim, Roy and Cohan, Arman and Shmueli-Scheuer, Michal},
  journal={arXiv preprint arXiv:2503.16416},
  year={2025}
}

@article{huang2025survey,
  title={A Survey on Hallucination in Large Language Models: Principles, Taxonomy, Challenges, and Open Questions},
  author={Huang, Lei and Yu, Weijiang and Ma, Weitao and Zhong, Weihong and Feng, Zhangyin and Wang, Haotian and Chen, Qianglong and Peng, Weihua and Feng, Xiaocheng and Qin, Bing and others},
  journal={ACM Transactions on Information Systems},
  volume={43},
  number={2},
  pages={1--55},
  year={2025}
}

@inproceedings{li2023evaluating,
    title = "Evaluating Object Hallucination in Large Vision-Language Models",
    author = "Li, Yifan  and
      Du, Yifan  and
      Zhou, Kun  and
      Wang, Jinpeng  and
      Zhao, Wayne Xin  and
      Wen, Ji-Rong",
    editor = "Bouamor, Houda  and
      Pino, Juan  and
      Bali, Kalika",
    booktitle = "Proceedings of the 2023 Conference on Empirical Methods in Natural Language Processing",
    month = dec,
    year = "2023",
    address = "Singapore",
    publisher = "Association for Computational Linguistics",
    url = "https://aclanthology.org/2023.emnlp-main.20/",
    doi = "10.18653/v1/2023.emnlp-main.20",
    pages = "292--305"
}

@article{kim2025medical,
  title={Medical Hallucinations in Foundation Models and Their Impact on Healthcare},
  author={Kim, Yubin and Jeong, Hyewon and Chen, Shan and Li, Shuyue Stella and Park, Chanwoo and Lu, Mingyu and Alhamoud, Kumail and Mun, Jimin and Grau, Cristina and Jung, Minseok and others},
  journal={arXiv preprint arXiv:2503.05777},
  year={2025}
}

@article{hakim2024need,
  title={The Need for Guardrails with Large Language Models in Medical Safety-Critical Settings: An Artificial Intelligence Application in the Pharmacovigilance Ecosystem},
  author={Hakim, Joe B. and Painter, Jeffery L. and Ramcharran, Darmendra and Kara, Vijay and Powell, Greg and Sobczak, Paulina and Sato, Chiho and Bate, Andrew and Beam, Andrew},
  journal={arXiv preprint arXiv:2407.18322},
  year={2024}
}

@article{dona2025bettercheck,
  title={{BetterCheck}: Towards Safeguarding {VLMs} for Automotive Perception Systems},
  author={Dona, Malsha Ashani Mahawatta and Cabrero-Daniel, Beatriz and Yu, Yinan and Berger, Christian},
  journal={arXiv preprint arXiv:2507.17722},
  year={2025}
}

@inproceedings{malinin2020uncertainty,
  title={Uncertainty Estimation in Autoregressive Structured Prediction},
  author={Malinin, Andrey and Gales, Mark},
  booktitle={International Conference on Learning Representations},
  year={2021}
}

@inproceedings{guerreiro2023looking,
    title = "Looking for a Needle in a Haystack: A Comprehensive Study of Hallucinations in Neural Machine Translation",
    author = "Guerreiro, Nuno M.  and
      Voita, Elena  and
      Martins, Andr{\'e}",
    editor = "Vlachos, Andreas  and
      Augenstein, Isabelle",
    booktitle = "Proceedings of the 17th Conference of the European Chapter of the Association for Computational Linguistics",
    month = may,
    year = "2023",
    address = "Dubrovnik, Croatia",
    publisher = "Association for Computational Linguistics",
    url = "https://aclanthology.org/2023.eacl-main.75/",
    doi = "10.18653/v1/2023.eacl-main.75",
    pages = "1059--1075"
}

@article{lin2022teaching,
  title={Teaching Models to Express Their Uncertainty in Words},
  author={Lin, Stephanie and Hilton, Jacob and Evans, Owain},
  journal={Transactions on Machine Learning Research},
  year={2022}
}

@article{kadavath2022language,
  title={Language Models (Mostly) Know What They Know},
  author={Kadavath, Saurav and Conerly, Tom and Askell, Amanda and Henighan, Tom and Drain, Dawn and Perez, Ethan and Schiefer, Nicholas and Hatfield-Dodds, Zac and DasSarma, Nova and Tran-Johnson, Eli and others},
  journal={arXiv preprint arXiv:2207.05221},
  year={2022}
}

@inproceedings{xiong2024can,
  title={Can {LLMs} Express Their Uncertainty? An Empirical Evaluation of Confidence Elicitation in {LLMs}},
  author={Xiong, Miao and Hu, Zhiyuan and Lu, Xinyang and Li, Yifei and Fu, Jie and He, Junxian and Hooi, Bryan},
  booktitle={International Conference on Learning Representations},
  year={2024}
}

@inproceedings{kuhn2023semantic,
  title={Semantic Uncertainty: Linguistic Invariances for Uncertainty Estimation in Natural Language Generation},
  author={Kuhn, Lorenz and Gal, Yarin and Farquhar, Sebastian},
  booktitle={International Conference on Learning Representations},
  year={2023}
}

@inproceedings{nikitin2024kernel,
  title={Kernel Language Entropy: Fine-grained Uncertainty Quantification for {LLMs} from Semantic Similarities},
  author={Nikitin, Alexander and Kossen, Jannik and Gal, Yarin and Marttinen, Pekka},
  booktitle={Advances in Neural Information Processing Systems},
  volume={37},
  pages={8901--8929},
  year={2024}
}

@article{bai2023qwen,
  title={{Qwen-VL}: A Versatile Vision-Language Model for Understanding, Localization, Text Reading, and Beyond},
  author={Bai, Jinze and Bai, Shuai and Yang, Shusheng and Wang, Shijie and Tan, Sinan and Wang, Peng and Lin, Junyang and Zhou, Chang and Zhou, Jingren},
  journal={arXiv preprint arXiv:2308.12966},
  year={2023}
}

@article{shorinwa2025survey,
  title={A Survey on Uncertainty Quantification of Large Language Models: Taxonomy, Open Research Challenges, and Future Directions},
  author={Shorinwa, Ola and Mei, Zhiting and Lidard, Justin and Ren, Allen Z. and Majumdar, Anirudha},
  journal={ACM Computing Surveys},
  volume={58},
  number={3},
  pages={1--38},
  year={2025}
}

@inproceedings{xia-etal-2025-survey,
    title = "A Survey of Uncertainty Estimation Methods on Large Language Models",
    author = "Xia, Zhiqiu  and
      Xu, Jinxuan  and
      Zhang, Yuqian  and
      Liu, Hang",
    editor = "Che, Wanxiang  and
      Nabende, Joyce  and
      Shutova, Ekaterina  and
      Pilehvar, Mohammad Taher",
    booktitle = "Findings of the Association for Computational Linguistics: ACL 2025",
    month = jul,
    year = "2025",
    address = "Vienna, Austria",
    publisher = "Association for Computational Linguistics",
    url = "https://aclanthology.org/2025.findings-acl.1101/",
    doi = "10.18653/v1/2025.findings-acl.1101",
    pages = "21381--21396",
    ISBN = "979-8-89176-256-5"
}

@inproceedings{zhang-etal-2025-atomic,
    title = "Atomic Calibration of {LLM}s in Long-Form Generations",
    author = "Zhang, Caiqi  and
      Yang, Ruihan  and
      Zhang, Zhisong  and
      Huang, Xinting  and
      Yang, Sen  and
      Yu, Dong  and
      Collier, Nigel",
    editor = "Inui, Kentaro  and
      Sakti, Sakriani  and
      Wang, Haofen  and
      Wong, Derek F.  and
      Bhattacharyya, Pushpak  and
      Banerjee, Biplab  and
      Ekbal, Asif  and
      Chakraborty, Tanmoy  and
      Singh, Dhirendra Pratap",
    booktitle = "Proceedings of the 14th International Joint Conference on Natural Language Processing and the 4th Conference of the Asia-Pacific Chapter of the Association for Computational Linguistics",
    month = dec,
    year = "2025",
    address = "Mumbai, India",
    publisher = "The Asian Federation of Natural Language Processing and The Association for Computational Linguistics",
    url = "https://aclanthology.org/2025.findings-ijcnlp.9/",
    doi = "10.18653/v1/2025.findings-ijcnlp.9",
    pages = "148--169",
    ISBN = "979-8-89176-303-6"
}

@inproceedings{chen-etal-2025-unveiling,
    title = "Unveiling Uncertainty: A Deep Dive into Calibration and Performance of Multimodal Large Language Models",
    author = "Chen, Zijun  and
      Hu, Wenbo  and
      He, Guande  and
      Deng, Zhijie  and
      Zhang, Zheng  and
      Hong, Richang",
    editor = "Rambow, Owen  and
      Wanner, Leo  and
      Apidianaki, Marianna  and
      Al-Khalifa, Hend  and
      Eugenio, Barbara Di  and
      Schockaert, Steven",
    booktitle = "Proceedings of the 31st International Conference on Computational Linguistics",
    month = jan,
    year = "2025",
    address = "Abu Dhabi, UAE",
    publisher = "Association for Computational Linguistics",
    url = "https://aclanthology.org/2025.coling-main.208/",
    pages = "3095--3109"
}

@article{wang2024qwen2,
  title={{Qwen2-VL}: Enhancing Vision-Language Model's Perception of the World at Any Resolution},
  author={Wang, Peng and Bai, Shuai and Tan, Sinan and Wang, Shijie and Fan, Zhihao and Bai, Jinze and Chen, Keqin and Liu, Xuejing and Wang, Jialin and Ge, Wenbin and others},
  journal={arXiv preprint arXiv:2409.12191},
  year={2024}
}

@article{bai2025qwen3,
  title={{Qwen3-VL} Technical Report},
  author={Bai, Shuai and Cai, Yuxuan and Chen, Ruizhe and Chen, Keqin and Chen, Xionghui and Cheng, Zesen and Deng, Lianghao and Ding, Wei and Gao, Chang and Ge, Chunjiang and others},
  journal={arXiv preprint arXiv:2511.21631},
  year={2025}
}

@article{brier1950verification,
  title={Verification of Forecasts Expressed in Terms of Probability},
  author={Brier, Glenn W.},
  journal={Monthly Weather Review},
  volume={78},
  number={1},
  pages={1--3},
  year={1950}
}

@inproceedings{guo2017calibration,
  title={On Calibration of Modern Neural Networks},
  author={Guo, Chuan and Pleiss, Geoff and Sun, Yu and Weinberger, Kilian Q.},
  booktitle={Proceedings of the 34th International Conference on Machine Learning},
  series={Proceedings of Machine Learning Research},
  volume={70},
  pages={1321--1330},
  year={2017}
}

@inproceedings{manakul-etal-2023-selfcheckgpt,
    title = "{S}elf{C}heck{GPT}: Zero-Resource Black-Box Hallucination Detection for Generative Large Language Models",
    author = "Manakul, Potsawee  and
      Liusie, Adian  and
      Gales, Mark",
    editor = "Bouamor, Houda  and
      Pino, Juan  and
      Bali, Kalika",
    booktitle = "Proceedings of the 2023 Conference on Empirical Methods in Natural Language Processing",
    month = dec,
    year = "2023",
    address = "Singapore",
    publisher = "Association for Computational Linguistics",
    url = "https://aclanthology.org/2023.emnlp-main.557/",
    doi = "10.18653/v1/2023.emnlp-main.557",
    pages = "9004--9017"
}

@inproceedings{zhang-etal-2024-luq,
    title = "{LUQ}: Long-text Uncertainty Quantification for {LLM}s",
    author = "Zhang, Caiqi  and
      Liu, Fangyu  and
      Basaldella, Marco  and
      Collier, Nigel",
    editor = "Al-Onaizan, Yaser  and
      Bansal, Mohit  and
      Chen, Yun-Nung",
    booktitle = "Proceedings of the 2024 Conference on Empirical Methods in Natural Language Processing",
    month = nov,
    year = "2024",
    address = "Miami, Florida, USA",
    publisher = "Association for Computational Linguistics",
    url = "https://aclanthology.org/2024.emnlp-main.299/",
    doi = "10.18653/v1/2024.emnlp-main.299",
    pages = "5244--5262"
}

@inproceedings{park-etal-2026-vauq,
    title = "{VAUQ}: Vision-Aware Uncertainty Quantification for {LVLM} Self-Evaluation",
    author = "Park, Seongheon  and
      Oh, Changdae  and
      Choi, Hyeong Kyu  and
      Du, Sean  and
      Li, Sharon",
    editor = "Liakata, Maria  and
      Moreira, Viviane P.  and
      Zhang, Jiajun  and
      Jurgens, David",
    booktitle = "Findings of the {A}ssociation for {C}omputational {L}inguistics: {ACL} 2026",
    month = jul,
    year = "2026",
    address = "San Diego, California, United States",
    publisher = "Association for Computational Linguistics",
    url = "https://aclanthology.org/2026.findings-acl.1321/",
    doi = "10.18653/v1/2026.findings-acl.1321",
    pages = "26534--26550",
    ISBN = "979-8-89176-395-1"
}

@inproceedings{geifman2017selective,
  title={Selective Classification for Deep Neural Networks},
  author={Geifman, Yonatan and El-Yaniv, Ran},
  booktitle={Advances in Neural Information Processing Systems},
  volume={30},
  year={2017}
}

@inproceedings{fadeeva-etal-2023-lm,
    title = "{LM}-Polygraph: Uncertainty Estimation for Language Models",
    author = "Fadeeva, Ekaterina  and
      Vashurin, Roman  and
      Tsvigun, Akim  and
      Vazhentsev, Artem  and
      Petrakov, Sergey  and
      Fedyanin, Kirill  and
      Vasilev, Daniil  and
      Goncharova, Elizaveta  and
      Panchenko, Alexander  and
      Panov, Maxim  and
      Baldwin, Timothy  and
      Shelmanov, Artem",
    booktitle = "Proceedings of the 2023 Conference on Empirical Methods in Natural Language Processing: System Demonstrations",
    month = dec,
    year = "2023",
    address = "Singapore",
    publisher = "Association for Computational Linguistics",
    url = "https://aclanthology.org/2023.emnlp-demo.41/",
    doi = "10.18653/v1/2023.emnlp-demo.41",
    pages = "446--461"
}

@article{vashurin-etal-2025-benchmarking,
    title = "Benchmarking Uncertainty Quantification Methods for Large Language Models with {LM}-Polygraph",
    author = "Vashurin, Roman  and
      Fadeeva, Ekaterina  and
      Vazhentsev, Artem  and
      Rvanova, Lyudmila  and
      Vasilev, Daniil  and
      Tsvigun, Akim  and
      Petrakov, Sergey  and
      Xing, Rui  and
      Sadallah, Abdelrahman  and
      Grishchenkov, Kirill  and
      Panchenko, Alexander  and
      Baldwin, Timothy  and
      Nakov, Preslav  and
      Panov, Maxim  and
      Shelmanov, Artem",
    journal = "Transactions of the Association for Computational Linguistics",
    volume = "13",
    year = "2025",
    address = "Cambridge, MA",
    publisher = "MIT Press",
    url = "https://aclanthology.org/2025.tacl-1.11/",
    doi = "10.1162/tacl_a_00737",
    pages = "220--248"
}

@inproceedings{geng-etal-2024-survey,
    title = "A Survey of Confidence Estimation and Calibration in Large Language Models",
    author = "Geng, Jiahui  and
      Cai, Fengyu  and
      Wang, Yuxia  and
      Koeppl, Heinz  and
      Nakov, Preslav  and
      Gurevych, Iryna",
    booktitle = "Proceedings of the 2024 Conference of the North American Chapter of the Association for Computational Linguistics: Human Language Technologies (Volume 1: Long Papers)",
    month = jun,
    year = "2024",
    address = "Mexico City, Mexico",
    publisher = "Association for Computational Linguistics",
    url = "https://aclanthology.org/2024.naacl-long.366/",
    doi = "10.18653/v1/2024.naacl-long.366",
    pages = "6577--6595"
}

@inproceedings{tian-etal-2023-just,
    title = "Just Ask for Calibration: Strategies for Eliciting Calibrated Confidence Scores from Language Models Fine-Tuned with Human Feedback",
    author = "Tian, Katherine  and
      Mitchell, Eric  and
      Zhou, Allan  and
      Sharma, Archit  and
      Rafailov, Rafael  and
      Yao, Huaxiu  and
      Finn, Chelsea  and
      Manning, Christopher",
    booktitle = "Proceedings of the 2023 Conference on Empirical Methods in Natural Language Processing",
    month = dec,
    year = "2023",
    address = "Singapore",
    publisher = "Association for Computational Linguistics",
    url = "https://aclanthology.org/2023.emnlp-main.330/",
    doi = "10.18653/v1/2023.emnlp-main.330",
    pages = "5433--5442"
}

@inproceedings{kamath-etal-2020-selective,
    title = "Selective Question Answering under Domain Shift",
    author = "Kamath, Amita  and
      Jia, Robin  and
      Liang, Percy",
    booktitle = "Proceedings of the 58th Annual Meeting of the Association for Computational Linguistics",
    month = jul,
    year = "2020",
    address = "Online",
    publisher = "Association for Computational Linguistics",
    url = "https://aclanthology.org/2020.acl-main.503/",
    doi = "10.18653/v1/2020.acl-main.503",
    pages = "5684--5696"
}

@inproceedings{dancette2023improving,
  title={Improving Selective Visual Question Answering by Learning from Your Peers},
  author={Dancette, Corentin and Whitehead, Spencer and Maheshwary, Rishabh and Vedantam, Ramakrishna and Scherer, Stefan and Chen, Xinlei and Cord, Matthieu and Rohrbach, Marcus},
  booktitle={Proceedings of the IEEE/CVF Conference on Computer Vision and Pattern Recognition (CVPR)},
  pages={24049--24059},
  year={2023}
}

@inproceedings{khan2024consistency,
  title={Consistency and Uncertainty: Identifying Unreliable Responses From Black-Box Vision-Language Models for Selective Visual Question Answering},
  author={Khan, Zaid and Fu, Yun},
  booktitle={Proceedings of the IEEE/CVF Conference on Computer Vision and Pattern Recognition (CVPR)},
  pages={10854--10863},
  year={2024}
}

@article{kossen2024semantic,
  title={Semantic Entropy Probes: Robust and Cheap Hallucination Detection in {LLMs}},
  author={Kossen, Jannik and Han, Jiatong and Razzak, Muhammed and Schut, Lisa and Malik, Shreshth and Gal, Yarin},
  journal={arXiv preprint arXiv:2406.15927},
  year={2024}
}

\clearpage
\appendix
\section{Appendix}
\label{sec:appendix}
\subsection{Implementation Details}
\label{app:impl}

\paragraph{Models and decoding.}
All MLLMs are loaded from Hugging Face in bfloat16 with FlashAttention-2 (InternVL: up to four image tiles). Every answer and sample is generated with at most 170 new tokens, top-$k{=}50$, top-$p{=}0.9$ and repetition penalty~1.05, at $T{=}0.1$ for the deployed answer and $T{=}1.0$ for the estimation pass and all semantic-method samples.

\paragraph{Prompts.}
Multiple choice: question, one option per line, then ``This is a single choice question, answer only with choice number in \{\dots\}''. MM-Vet appends ``NOTE: Provide only the final answer. Do not provide unrelated details.''; MM-Vet Long replaces it with ``NOTE: Please provide a detailed and thorough answer.''; FSVQA200 and LLaVA-Bench use the raw question. Verbal Abstain appends ``Notice that if you cannot answer confidently, return `unknown' as the answer.''; a response containing \emph{unknown} counts as abstention. Verbal Confidence appends ``Moreover, please rate your confidence in your answer between 0 and 100\%. Answer only with the following format: \texttt{\{"answer": "<your answer>",}\allowbreak\texttt{ "confidence": <number>\}}. Do not output anything else.''; unparseable outputs receive uncertainty~1 and their answer text is still judged.

\paragraph{P(true).}
Zero-shot prompt of \citet{kadavath2022language}: ``\{question\}\textbackslash n Possible Answer: \{answer\}\textbackslash n Is the possible answer: (A) True (B) False\textbackslash n The possible answer is:''. The self-evaluation pass is decoded greedily on the raw logits (no top-$k$, top-$p$ or repetition penalty, which would otherwise pin the option probabilities to 0.5 or 1.0). At the first decoding step (within the first eight) that commits to an option, the masses of all surface forms of each option (\texttt{A}, \texttt{(A}, \texttt{A)}, \texttt{True}, \dots; likewise for B/False) are pooled and normalized over the two options; the uncertainty is $1-P(\text{true})$.

\paragraph{Correctness labels.}
Every deployed answer is labelled by GPT-4o (Azure OpenAI, temperature~0), which receives the question, the reference answer and the model answer and returns \emph{Yes}/\emph{No} on whether they match meaningfully, accepting synonyms and paraphrases. For multiple choice the reference is the option index, so the judgement reduces to option matching while tolerating formats such as ``(1)'' or ``1: \dots''; for the open-ended variants of App.~\ref{app:length_format} the reference is the option text. The same judge and prompt are used for all models, methods, benchmarks and seeds; items whose judge call failed are excluded.

\paragraph{Token-level.}
Entropy and NLL are computed from the per-step distribution the model samples from, i.e.\ after temperature, top-$k$, top-$p$ and repetition penalty, and aggregated over the generated tokens by mean or max.

\paragraph{Semantic.}
$N\in\{5,10\}$ samples at $T{=}1$. Clusters are formed by bidirectional entailment with Qwen2.5-7B-Instruct (``Does `$a$' entail `$b$'? Respond with either `Yes' or `No' only.'', $T{=}0.1$); byte-identical samples are merged without a judge call, and for multiple choice the cluster is the selected option. SE uses length-normalized sequence log-likelihoods \citep{farquhar2024detecting}. VL-Uncertainty applies progressive Gaussian blur (radii 0.6--1.4) and LLM rephrasing of the question; VASE uses $\alpha{=}1$. KLE is computed offline from the stored $N{=}10$ SE samples and their entailment judgements, with binary weights $W_{ij}\in\{0,1,2\}$ (number of entailment directions that hold), heat kernel $t{=}0.3$ and unit-trace normalization as in \citet{nikitin2024kernel}; its cost therefore equals SE's. The cost column of Tab.~\ref{tab:answer_length} counts MLLM forward passes only; entailment clustering adds up to $N(N{-}1)$ calls to the 7B judge per item (90 for $N{=}10$, fewer after exact-match merging), and KLE reuses the same judgements.

\paragraph{Calibration and router.}
$\mathcal{D}_{\text{cal}}$ is a fixed draw (seed 0) of 60 indices per benchmark from the full benchmark index range; indices outside the evaluated subsets of ScienceQA and MMMU are skipped. A single threshold and one 8-bin isotonic probability map per model and method are fit on the pooled set and applied unchanged to every evaluation set, so they are not tuned to any one of them. The router counts whitespace-separated words of the deployed answer, with bins ${\le}5$, 6--60 and ${>}60$, and fits threshold and map per bin; overridden items receive error probability~1, so they rank above every scored item.

\paragraph{Compute and seeds.}
Experiments ran with 8$\times$A100-40GB and 8$\times$ Blackwell Server Edition RTX PRO~6000 GPUs. All 12 models, methods and evaluation sets use seeds 0--2; the format ablation uses seed~0 for the four largest models. Code, per-item results and the FSVQA200 IDs are provided as supplementary material and will be released. Software: Python 3.11, PyTorch 2.5.1 (CUDA 12.1; 2.8.0 with CUDA 12.8 on the RTX PRO 6000 nodes), Transformers 4.57.6 (4.37.2 for InternVL2), FlashAttention 2.8.3, Accelerate 1.12.0, Datasets 4.5.0, NumPy 2.3.5 and SciPy 1.17.0; the GPT-4o judge is queried through the Azure OpenAI API.

\subsection{Individual Benchmark Results}
\label{app:per_benchmark}
In this section we provide the results for all model, method and metric combinations for the individual benchmarks studied in the main paper. We organize the discussion by benchmark, ordered along the answer-length axis---from multiple-choice (ScienceQA, MMMU) through short open-ended (MM-Vet), sentence-length (FSVQA200), and finally long-form multi-sentence generation (MM-Vet Long, LLaVA-Bench)---which we find to be the principal axis along which the optimal uncertainty estimator varies. For each benchmark we report AUROC, selective-prediction coverage at a 20\% risk budget, and CHDA, the deployment-time metric we focus on in the main paper; all values are means over three seeds (the seed standard deviation of model accuracy is below one point except on the 60-item LLaVA-Bench).

\paragraph{ScienceQA~\cite{lu2022learn}} Table~\ref{tab:scienceqa} reports our full results on the multiple-choice ScienceQA benchmark (2--4 options; mean accuracy 84.9\%). Consistent with the main paper, token-level methods with $T{=}1$ resampling offer the best cost--quality trade-off: at $2\times$ inference cost, Max Entropy ($T{=}1$) reaches 0.79 average AUROC, the same as Semantic Entropy at $11\times$, and leads CHDA (85.7 against 85.2 for SE and KLE). Semantic methods yield tangible gains only on the strongest backbones (Semantic Entropy reaches 0.89 AUROC on Qwen2-VL-72B and 0.88 on InternVL2-8B), while the text-based verbalized methods remain unreliable (0.63 AUROC for Verbal Abstain, whose CHDA ranges from 15.9 on InternVL2-2B to 94.9 on InternVL3.5-38B), since the multiple-choice format biases models toward committing to a single option. P(true) ranks well (0.77 AUROC) but its threshold transfers poorly (64.3 CHDA). Cov@20 saturates between 0.90 and 0.95 for every family except Verbal Abstain (0.67), reflecting the high base accuracy: in this regime even minimal uncertainty signals are sufficient to drive deployment-time selection.

\paragraph{MMMU~\cite{yue2023mmmu}} Table~\ref{tab:MMMU} reports our full results on this multiple-choice benchmark, which extends ScienceQA's setup to a larger answer space (2--9 options) and consequently a substantially lower average accuracy (45.8\% vs.\ 84.9\% on ScienceQA). The harder regime sharpens the ranking result of the main paper: Max Entropy ($T{=}1$) leads AUROC (0.72), while P(true), SE and KLE tie at 0.68 and Verbal Abstain stays at chance (0.55). Coverage is low for every method (Cov@20 $\leq 0.20$) because few answers are correct enough to fill a 20\% risk budget. The CHDA picture diverges from AUROC in a deployment-relevant way: P(true) transfers its threshold best (66.6), KLE follows (58.5), token-level $T{=}1$ methods reach 56.1, and the remaining semantic methods fall to 47.5--50.7, so ranking-only metrics overstate the deployment value of expensive sampling-based estimators in this regime. Overall, MMMU reinforces that token-level $T{=}1$ resampling is the most robust ranking signal for multiple-choice questions, and that self-evaluation gives the most transferable threshold.

\paragraph{MM-Vet~\cite{yu2023mm}} Table~\ref{tab:mmvet} reports results on MM-Vet, where models produce short open-ended free-form answers (mean accuracy 55.4\%). This is the benchmark where KLE separates most clearly from the rest of its family: it leads every metric (0.81 AUROC, 76.3 CHDA, 0.49 Cov@20) against 0.71 / 68.7 / 0.37 for Semantic Entropy, because short free-form answers spread the ten samples over many clusters (Tab.~\ref{tab:abstain_cluster_merged}) and graded entailment weights keep ranking items that hard clustering no longer separates. Max Entropy ($T{=}1$) is the best cheap option (0.74 / 68.2 / 0.38 at $2\times$). The open short-form format admits a clean ``unknown'' surface form for Verbal Abstain, which abstains on 16.8\% of samples---the highest rate across our benchmarks (Tab.~\ref{tab:abstain_cluster_merged})---and reaches a moderate 0.66 AUROC and 67.5 CHDA; on specific backbones (Qwen3-VL-8B 77.8, Qwen2-VL-72B 76.9 CHDA) it is the preferred choice, hinting that the optimal estimator is not entirely backbone-agnostic. P(true) is weak here (0.63 AUROC), since a single-word answer gives the model little to evaluate.

\paragraph{FSVQA200~\cite{shin2016color}} Table~\ref{tab:fsvqa200} reports our full results on FSVQA200, where models produce sentence-length free-form answers (mean accuracy 68.7\%). The longer format inverts the trends of the short-form benchmarks: the verbalized family leads, with P(true) first on AUROC (0.72) and Cov@20 (0.65) and Verbal Abstain first on CHDA (76.0, five points above P(true)) at single-pass $1\times$ cost. Token-level methods degrade markedly (Max Entropy $T{=}1$: 0.64 AUROC, 62.3 CHDA), as uncertainty is no longer localized in a few decisive tokens but diffused across the generated meaning. Semantic methods stay competitive on the strongest backbones (Semantic Entropy reaches 0.73 AUROC on InternVL3.5-8B, KLE 0.68 on average) but their $6$--$11\times$ cost is hard to justify against the verbalized methods. The harder Qwen2-VL-2B setting (accuracy 47.3\%) remains the principal discriminator, where Verbal Abstain and P(true) both reach 0.83 AUROC, far ahead of the sampling-based methods. From a deployment perspective, CHDA and AUROC agree on the family but not on the member: P(true) ranks answers best, while the binary abstention signal gives the threshold that transfers best.

\paragraph{MM-Vet Long} Table~\ref{tab:mmvetlong} reports results on MM-Vet Long, a variant of MM-Vet~\cite{yu2023mm} in which the instruction to produce a short final answer is replaced with one requesting a complete, multi-sentence response (mean accuracy 54.6\%, comparable to MM-Vet's 55.4\%). Because the question set is identical, the comparison with Table~\ref{tab:mmvet} isolates the effect of answer length; Appendix~\ref{app:length_format} discusses it together with the converse format ablation. In brief, the ranking inverts: KLE still leads (0.75 AUROC, 69.4 CHDA), P(true) follows closely and has the best coverage (0.73, 68.3, 0.34 Cov@20 against 0.30 for KLE), the token-level $T{=}1$ methods that were competitive on MM-Vet collapse (0.57 AUROC, 59.0 CHDA, 0.06 Cov@20), and Verbal Abstain drops to chance (0.53 AUROC vs.\ 0.65 on MM-Vet) as the long-answer instruction suppresses abstention (3.3\%, Tab.~\ref{tab:abstain_cluster_merged}).

\paragraph{LLaVA-Bench~\cite{liu2023visual}} Table~\ref{tab:llava_bench} reports results on LLaVA-Bench, where models produce long-form, multi-sentence open-ended responses (60 items; mean accuracy 56.6\%). The results reinforce the trend observed on MM-Vet Long: sampling-based semantic methods dominate on AUROC and CHDA, with VL-Uncertainty best (0.75, 70.6), VASE (0.72, 69.5) and KLE (0.72, 68.3) close behind, while token-level methods---both $T{=}0.1$ and $T{=}1$ variants---collapse to 0.56 AUROC, the largest gap of any benchmark in our suite. P(true) is competitive (0.71 AUROC, 67.6 CHDA, 0.37 Cov@20) at a fifth of the cost. Verbal Abstain remains near chance on AUROC (0.56) but is a notable outlier on CHDA (65.3), driven by strong per-backbone performance on Qwen3-VL-8B (77.8) and Qwen2-VL-7B (75.0), so for cost-sensitive deployments on specific backbones the $1\times$ verbal route remains a defensible fallback even in long-form regimes. Overall, LLaVA-Bench cements meaning-level consistency as the robust choice for genuinely long-form generation, with P(true) as the cheap alternative once models are large enough to judge their own answers.

\begin{table*}
\centering
\scriptsize
{\footnotesize\setlength{\tabcolsep}{2.6pt}
\begin{tabular}{lc|ccc|ccc|ccc|ccc|c}
\toprule
Method & Cost & \multicolumn{3}{c|}{InternVL2} & \multicolumn{3}{c|}{InternVL3.5} & \multicolumn{3}{c|}{Qwen2-VL} & \multicolumn{3}{c|}{Qwen3-VL} & Avg \\
 & & 2B & 8B & 26B & 2B & 8B & 38B & 2B & 7B & 72B & 2B & 8B & 32B & \\
\midrule
\texttt{Model Acc} & -- & 84.7 & 90.1 & 91.9 & 81.4 & 89.8 & 92.3 & 64.9 & 79.7 & 88.9 & 74.8 & 89.9 & 90.9 & 84.9 \\
\midrule
\multicolumn{15}{l}{\textbf{AUROC $\uparrow$}} \\
\midrule
Avg Entropy & $1\times$ & 0.57 & 0.55 & 0.58 & 0.65 & 0.53 & 0.61 & 0.61 & 0.60 & 0.57 & 0.70 & 0.58 & 0.52 & 0.59 \\
Max Entropy & $1\times$ & 0.56 & 0.55 & 0.58 & 0.64 & 0.53 & 0.62 & 0.60 & 0.60 & 0.57 & 0.71 & 0.58 & 0.51 & 0.59 \\
Avg NLL & $1\times$ & 0.57 & 0.55 & 0.58 & 0.64 & 0.53 & 0.62 & 0.61 & 0.59 & 0.57 & 0.70 & 0.58 & 0.51 & 0.59 \\
Max NLL & $1\times$ & 0.57 & 0.56 & 0.57 & 0.65 & 0.53 & 0.62 & 0.61 & 0.59 & 0.57 & 0.72 & 0.58 & 0.51 & 0.59 \\
Avg Entropy (T=1) & $2\times$ & 0.74 & 0.82 & 0.84 & 0.77 & 0.78 & 0.79 & 0.72 & 0.83 & 0.88 & 0.78 & 0.75 & 0.68 & 0.78 \\
Max Entropy (T=1) & $2\times$ & \textbf{0.80} & 0.83 & \textbf{0.85} & \underline{0.79} & \underline{0.78} & 0.80 & 0.72 & 0.82 & 0.87 & \textbf{0.84} & 0.75 & 0.68 & \textbf{0.79} \\
Avg NLL (T=1) & $2\times$ & 0.75 & 0.75 & 0.82 & 0.77 & 0.77 & 0.80 & 0.67 & 0.78 & 0.83 & 0.79 & 0.75 & 0.67 & 0.76 \\
Max NLL (T=1) & $2\times$ & 0.77 & 0.71 & 0.81 & \textbf{0.79} & 0.77 & 0.80 & 0.66 & 0.76 & 0.80 & \underline{0.83} & 0.76 & 0.67 & 0.76 \\
\midrule
Verbal Abstain & $1\times$ & 0.50 & 0.51 & 0.59 & 0.69 & 0.57 & \underline{0.81} & 0.59 & 0.56 & 0.74 & 0.62 & 0.68 & 0.66 & 0.63 \\
P(true) & $2\times$ & 0.65 & 0.77 & 0.84 & 0.75 & \textbf{0.79} & \textbf{0.87} & 0.56 & 0.77 & \underline{0.88} & 0.63 & \textbf{0.81} & \textbf{0.88} & 0.77 \\
Verbal Confidence & $1\times$ & 0.55 & 0.60 & 0.56 & 0.77 & 0.57 & 0.62 & 0.48 & 0.57 & 0.50 & 0.49 & 0.48 & \underline{0.73} & 0.58 \\
\midrule
Semantic Entropy (N=10) & $11\times$ & \underline{0.79} & \textbf{0.88} & 0.85 & 0.74 & 0.75 & 0.78 & \underline{0.74} & \underline{0.84} & \textbf{0.89} & 0.80 & 0.74 & 0.67 & \underline{0.79} \\
Semantic Entropy (N=5) & $6\times$ & 0.74 & 0.81 & 0.77 & 0.66 & 0.71 & 0.70 & 0.70 & 0.79 & 0.85 & 0.72 & 0.70 & 0.63 & 0.73 \\
VASE (N=5) & $6\times$ & 0.78 & 0.85 & 0.85 & 0.74 & 0.77 & 0.78 & 0.69 & 0.81 & 0.87 & 0.79 & \underline{0.76} & 0.69 & 0.78 \\
VL-Uncertainty (N=10) & $11\times$ & 0.78 & 0.84 & \underline{0.85} & 0.76 & 0.74 & 0.77 & \textbf{0.75} & \textbf{0.84} & 0.87 & 0.77 & 0.71 & 0.63 & 0.78 \\
KLE (N=10) & $11\times$ & 0.79 & \underline{0.87} & 0.83 & 0.73 & 0.75 & 0.77 & 0.72 & 0.83 & 0.88 & 0.79 & 0.74 & 0.67 & 0.78 \\
\midrule
\multicolumn{15}{l}{\textbf{Cov@20 $\uparrow$}} \\
\midrule
Avg Entropy & $1\times$ & \textbf{1.00} & \textbf{1.00} & \textbf{1.00} & \textbf{1.00} & \textbf{1.00} & \textbf{1.00} & 0.00 & 0.98 & \textbf{1.00} & 0.88 & \textbf{1.00} & \textbf{1.00} & 0.91 \\
Max Entropy & $1\times$ & \textbf{1.00} & \textbf{1.00} & \textbf{1.00} & \textbf{1.00} & \textbf{1.00} & \textbf{1.00} & 0.00 & 0.98 & \textbf{1.00} & 0.87 & \textbf{1.00} & \textbf{1.00} & 0.90 \\
Avg NLL & $1\times$ & \textbf{1.00} & \textbf{1.00} & \textbf{1.00} & \textbf{1.00} & \textbf{1.00} & \textbf{1.00} & 0.00 & \underline{1.00} & \textbf{1.00} & 0.89 & \textbf{1.00} & \textbf{1.00} & 0.91 \\
Max NLL & $1\times$ & \textbf{1.00} & \textbf{1.00} & \textbf{1.00} & \textbf{1.00} & \textbf{1.00} & \textbf{1.00} & 0.00 & \textbf{1.00} & \textbf{1.00} & 0.92 & \textbf{1.00} & \textbf{1.00} & 0.91 \\
Avg Entropy (T=1) & $2\times$ & \textbf{1.00} & \textbf{1.00} & \textbf{1.00} & \textbf{1.00} & \textbf{1.00} & \textbf{1.00} & 0.48 & 0.99 & \textbf{1.00} & 0.85 & \textbf{1.00} & \textbf{1.00} & 0.94 \\
Max Entropy (T=1) & $2\times$ & \textbf{1.00} & \textbf{1.00} & \textbf{1.00} & \textbf{1.00} & \textbf{1.00} & \textbf{1.00} & 0.47 & 0.99 & \textbf{1.00} & \underline{0.92} & \textbf{1.00} & \textbf{1.00} & 0.95 \\
Avg NLL (T=1) & $2\times$ & \textbf{1.00} & \textbf{1.00} & \textbf{1.00} & \textbf{1.00} & \textbf{1.00} & \textbf{1.00} & 0.42 & 0.98 & \textbf{1.00} & 0.83 & \textbf{1.00} & \textbf{1.00} & 0.94 \\
Max NLL (T=1) & $2\times$ & \textbf{1.00} & \textbf{1.00} & \textbf{1.00} & \textbf{1.00} & \textbf{1.00} & \textbf{1.00} & 0.41 & 0.99 & \textbf{1.00} & 0.92 & \textbf{1.00} & \textbf{1.00} & 0.94 \\
\midrule
Verbal Abstain & $1\times$ & \textbf{1.00} & \textbf{1.00} & \textbf{1.00} & 0.00 & \textbf{1.00} & \textbf{1.00} & 0.00 & 0.00 & \textbf{1.00} & 0.00 & \textbf{1.00} & \textbf{1.00} & 0.67 \\
P(true) & $2\times$ & \textbf{1.00} & \textbf{1.00} & \textbf{1.00} & \textbf{1.00} & \textbf{1.00} & \textbf{1.00} & 0.16 & 0.98 & \textbf{1.00} & 0.63 & \textbf{1.00} & \textbf{1.00} & 0.90 \\
Verbal Confidence & $1\times$ & \textbf{1.00} & \textbf{1.00} & \textbf{1.00} & \underline{0.82} & \textbf{1.00} & \textbf{1.00} & 0.05 & 0.95 & \textbf{1.00} & \textbf{1.00} & \textbf{1.00} & \textbf{1.00} & 0.90 \\
\midrule
Semantic Entropy (N=10) & $11\times$ & \textbf{1.00} & \textbf{1.00} & \textbf{1.00} & \textbf{1.00} & \textbf{1.00} & \textbf{1.00} & \underline{0.52} & 1.00 & \textbf{1.00} & 0.88 & \textbf{1.00} & \textbf{1.00} & \textbf{0.95} \\
Semantic Entropy (N=5) & $6\times$ & \textbf{1.00} & \textbf{1.00} & \textbf{1.00} & \textbf{1.00} & \textbf{1.00} & \textbf{1.00} & 0.44 & 0.99 & \textbf{1.00} & 0.88 & \textbf{1.00} & \textbf{1.00} & 0.94 \\
VASE (N=5) & $6\times$ & \textbf{1.00} & \textbf{1.00} & \textbf{1.00} & \textbf{1.00} & \textbf{1.00} & \textbf{1.00} & 0.36 & 0.99 & \textbf{1.00} & 0.88 & \textbf{1.00} & \textbf{1.00} & 0.94 \\
VL-Uncertainty (N=10) & $11\times$ & \textbf{1.00} & \textbf{1.00} & \textbf{1.00} & \textbf{1.00} & \textbf{1.00} & \textbf{1.00} & \textbf{0.53} & 0.99 & \textbf{1.00} & 0.86 & \textbf{1.00} & \textbf{1.00} & \underline{0.95} \\
KLE (N=10) & $11\times$ & \textbf{1.00} & \textbf{1.00} & \textbf{1.00} & \textbf{1.00} & \textbf{1.00} & \textbf{1.00} & 0.45 & 0.99 & \textbf{1.00} & 0.87 & \textbf{1.00} & \textbf{1.00} & 0.94 \\
\midrule
\multicolumn{15}{l}{\textbf{CHDA $\uparrow$}} \\
\midrule
Avg Entropy & $1\times$ & 84.6 & 84.4 & 89.6 & \textbf{83.2} & 90.0 & 92.1 & 67.3 & 80.3 & 88.0 & 76.0 & 90.0 & 90.9 & 84.7 \\
Max Entropy & $1\times$ & 84.3 & 84.4 & 91.0 & 82.9 & 89.8 & 92.3 & 66.8 & 79.7 & 88.9 & \textbf{82.4} & 90.0 & 90.9 & 85.3 \\
Avg NLL & $1\times$ & 84.5 & 84.3 & 89.7 & 82.9 & 89.9 & 92.0 & \underline{67.6} & 80.0 & 88.0 & 73.6 & 90.0 & 90.8 & 84.4 \\
Max NLL & $1\times$ & 84.7 & 84.6 & 91.2 & \underline{83.1} & 89.9 & 92.0 & 67.0 & 80.0 & \underline{89.0} & \underline{81.3} & 90.0 & 90.9 & \underline{85.3} \\
Avg Entropy (T=1) & $2\times$ & 83.8 & \underline{90.3} & 91.9 & 80.6 & 90.2 & 92.3 & 61.9 & 80.6 & 89.0 & 75.0 & 90.2 & 91.2 & 84.7 \\
Max Entropy (T=1) & $2\times$ & \textbf{85.0} & 90.2 & 91.9 & 81.9 & 89.8 & 92.2 & 66.9 & \textbf{80.9} & 88.9 & 79.8 & 89.8 & 90.9 & \textbf{85.7} \\
Avg NLL (T=1) & $2\times$ & 81.0 & 82.7 & 91.1 & 77.4 & 89.8 & 91.9 & 58.0 & 77.5 & 85.7 & 74.8 & 89.9 & 90.9 & 82.5 \\
Max NLL (T=1) & $2\times$ & 78.3 & 87.3 & 91.7 & 79.8 & 89.9 & 92.1 & 60.7 & 79.9 & 89.0 & 77.0 & 89.8 & 90.9 & 83.9 \\
\midrule
Verbal Abstain & $1\times$ & 15.9 & 89.7 & 91.9 & 80.0 & 90.2 & \textbf{94.9} & 64.7 & 79.8 & \textbf{91.4} & 77.7 & \textbf{91.5} & \textbf{92.3} & 80.0 \\
P(true) & $2\times$ & 47.5 & 61.9 & 69.9 & 67.4 & 80.3 & 82.7 & 37.2 & 54.1 & 71.4 & 50.6 & 68.3 & 79.9 & 64.3 \\
Verbal Confidence & $1\times$ & 53.2 & 87.5 & 91.5 & 81.9 & \textbf{90.3} & \underline{93.8} & 42.1 & 78.5 & 88.2 & 80.5 & 90.2 & 90.7 & 80.7 \\
\midrule
Semantic Entropy (N=10) & $11\times$ & 84.6 & 90.0 & 91.8 & 81.3 & 89.9 & 92.3 & \textbf{68.2} & 80.2 & 88.8 & 74.7 & 90.2 & 90.8 & 85.2 \\
Semantic Entropy (N=5) & $6\times$ & 84.9 & \textbf{90.4} & 91.8 & 81.3 & 89.9 & 92.3 & 66.5 & 79.9 & 88.4 & 74.6 & 90.2 & 91.0 & 85.1 \\
VASE (N=5) & $6\times$ & 84.5 & 90.2 & \textbf{92.0} & 80.9 & 90.0 & 92.2 & 66.5 & 79.6 & 88.5 & 74.5 & 89.8 & 90.9 & 85.0 \\
VL-Uncertainty (N=10) & $11\times$ & 84.4 & 90.0 & \underline{91.9} & 81.3 & 90.0 & 92.3 & 66.8 & \underline{80.7} & 88.7 & 74.6 & 90.0 & 90.9 & 85.1 \\
KLE (N=10) & $11\times$ & \underline{85.0} & 90.2 & 91.8 & 80.5 & \underline{90.2} & 91.7 & 66.7 & 80.0 & 87.5 & 76.7 & \underline{90.5} & \underline{91.5} & 85.2 \\
\bottomrule
\end{tabular}}

\caption{Per-model results on multiple-choice benchmark \textbf{ScienceQA} across AUROC, selective-prediction coverage at 20\% risk, and CHDA (12 MLLMs, mean over three seeds). Token-level methods with $T{=}1$ resampling match the costlier semantic methods on AUROC and lead CHDA; text-based verbalized methods are unreliable, and P(true) ranks well but its threshold transfers poorly.}
\label{tab:scienceqa}
\end{table*}

\begin{table*}
\centering
\scriptsize
{\footnotesize\setlength{\tabcolsep}{2.6pt}
\begin{tabular}{lc|ccc|ccc|ccc|ccc|c}
\toprule
Method & Cost & \multicolumn{3}{c|}{InternVL2} & \multicolumn{3}{c|}{InternVL3.5} & \multicolumn{3}{c|}{Qwen2-VL} & \multicolumn{3}{c|}{Qwen3-VL} & Avg \\
 & & 2B & 8B & 26B & 2B & 8B & 38B & 2B & 7B & 72B & 2B & 8B & 32B & \\
\midrule
\texttt{Model Acc} & -- & 33.3 & 42.6 & 41.9 & 34.8 & 55.4 & 56.1 & 34.1 & 48.9 & 56.8 & 30.1 & 53.5 & 61.8 & 45.8 \\
\midrule
\multicolumn{15}{l}{\textbf{AUROC $\uparrow$}} \\
\midrule
Avg Entropy & $1\times$ & 0.54 & 0.55 & 0.61 & 0.70 & 0.61 & 0.68 & 0.55 & 0.55 & 0.60 & 0.71 & 0.56 & 0.53 & 0.60 \\
Max Entropy & $1\times$ & 0.54 & 0.54 & 0.67 & 0.74 & 0.60 & 0.67 & 0.59 & 0.54 & 0.61 & 0.74 & 0.56 & 0.54 & 0.61 \\
Avg NLL & $1\times$ & 0.54 & 0.54 & 0.63 & 0.72 & 0.61 & 0.68 & 0.55 & 0.56 & 0.61 & 0.72 & 0.56 & 0.53 & 0.60 \\
Max NLL & $1\times$ & 0.55 & 0.53 & 0.69 & 0.75 & 0.61 & 0.68 & 0.60 & 0.55 & 0.62 & 0.74 & 0.56 & 0.54 & 0.62 \\
Avg Entropy (T=1) & $2\times$ & 0.57 & 0.65 & 0.66 & 0.62 & 0.74 & 0.71 & 0.60 & \underline{0.70} & 0.72 & 0.59 & 0.69 & 0.70 & 0.66 \\
Max Entropy (T=1) & $2\times$ & \textbf{0.59} & 0.68 & \textbf{0.77} & \textbf{0.78} & \textbf{0.77} & \underline{0.78} & \textbf{0.66} & \textbf{0.70} & 0.77 & \textbf{0.76} & \textbf{0.71} & \underline{0.70} & \textbf{0.72} \\
Avg NLL (T=1) & $2\times$ & 0.57 & 0.64 & 0.67 & 0.67 & 0.74 & 0.73 & 0.57 & 0.65 & 0.70 & 0.63 & 0.70 & 0.70 & 0.66 \\
Max NLL (T=1) & $2\times$ & 0.57 & 0.61 & \underline{0.75} & \underline{0.78} & \underline{0.76} & 0.78 & 0.62 & 0.65 & 0.70 & \underline{0.76} & \underline{0.70} & 0.70 & \underline{0.70} \\
\midrule
Verbal Abstain & $1\times$ & 0.50 & 0.50 & 0.51 & 0.53 & 0.52 & 0.53 & 0.58 & 0.66 & 0.56 & 0.61 & 0.55 & 0.55 & 0.55 \\
P(true) & $2\times$ & 0.53 & 0.64 & 0.72 & 0.77 & 0.73 & \textbf{0.78} & 0.55 & 0.67 & 0.74 & 0.62 & 0.69 & \textbf{0.73} & 0.68 \\
Verbal Confidence & $1\times$ & 0.51 & 0.49 & 0.47 & 0.68 & 0.60 & 0.60 & 0.52 & 0.52 & 0.55 & 0.53 & 0.52 & 0.57 & 0.55 \\
\midrule
Semantic Entropy (N=10) & $11\times$ & \underline{0.58} & \textbf{0.68} & 0.72 & 0.65 & 0.74 & 0.71 & 0.62 & 0.70 & \textbf{0.77} & 0.66 & 0.68 & 0.68 & 0.68 \\
Semantic Entropy (N=5) & $6\times$ & 0.57 & 0.65 & 0.68 & 0.59 & 0.70 & 0.65 & 0.59 & 0.67 & 0.75 & 0.61 & 0.63 & 0.64 & 0.64 \\
VASE (N=5) & $6\times$ & 0.57 & 0.68 & 0.70 & 0.63 & 0.73 & 0.71 & \underline{0.62} & 0.68 & 0.75 & 0.64 & 0.67 & 0.69 & 0.67 \\
VL-Uncertainty (N=10) & $11\times$ & 0.58 & \underline{0.68} & 0.70 & 0.63 & 0.73 & 0.73 & 0.62 & 0.69 & 0.77 & 0.62 & 0.66 & 0.68 & 0.67 \\
KLE (N=10) & $11\times$ & 0.58 & 0.67 & 0.72 & 0.64 & 0.74 & 0.71 & 0.61 & 0.69 & \underline{0.77} & 0.66 & 0.68 & 0.68 & 0.68 \\
\midrule
\multicolumn{15}{l}{\textbf{Cov@20 $\uparrow$}} \\
\midrule
Avg Entropy & $1\times$ & \underline{0.00} & 0.00 & 0.00 & \underline{0.00} & 0.00 & 0.00 & 0.00 & 0.00 & 0.00 & 0.00 & 0.00 & 0.00 & 0.00 \\
Max Entropy & $1\times$ & \underline{0.00} & 0.00 & 0.00 & \underline{0.00} & 0.00 & 0.00 & 0.00 & 0.00 & 0.00 & 0.00 & 0.00 & 0.00 & 0.00 \\
Avg NLL & $1\times$ & \underline{0.00} & 0.00 & 0.00 & \underline{0.00} & 0.00 & 0.00 & 0.00 & 0.00 & 0.00 & 0.00 & 0.00 & 0.00 & 0.00 \\
Max NLL & $1\times$ & \underline{0.00} & 0.00 & 0.00 & \underline{0.00} & 0.00 & 0.00 & 0.00 & 0.00 & 0.00 & 0.00 & 0.00 & 0.00 & 0.00 \\
Avg Entropy (T=1) & $2\times$ & \underline{0.00} & 0.06 & 0.08 & \underline{0.00} & 0.39 & 0.42 & 0.01 & 0.19 & 0.28 & 0.00 & 0.00 & 0.00 & 0.12 \\
Max Entropy (T=1) & $2\times$ & \underline{0.00} & 0.09 & 0.13 & \underline{0.00} & \textbf{0.43} & 0.44 & \textbf{0.04} & 0.20 & 0.43 & 0.00 & 0.00 & 0.00 & \underline{0.15} \\
Avg NLL (T=1) & $2\times$ & \underline{0.00} & 0.08 & 0.12 & \underline{0.00} & 0.42 & 0.44 & 0.03 & 0.18 & 0.33 & 0.00 & 0.00 & 0.00 & 0.13 \\
Max NLL (T=1) & $2\times$ & \underline{0.00} & 0.08 & 0.09 & \underline{0.00} & \underline{0.42} & \textbf{0.47} & 0.04 & 0.18 & 0.37 & 0.00 & 0.00 & 0.00 & 0.14 \\
\midrule
Verbal Abstain & $1\times$ & \underline{0.00} & 0.00 & 0.00 & \underline{0.00} & 0.00 & 0.00 & 0.00 & 0.00 & 0.00 & 0.00 & 0.00 & 0.00 & 0.00 \\
P(true) & $2\times$ & \textbf{0.00} & 0.10 & 0.16 & \underline{0.00} & 0.36 & \underline{0.46} & 0.00 & 0.15 & 0.40 & 0.00 & \textbf{0.27} & \textbf{0.48} & \textbf{0.20} \\
Verbal Confidence & $1\times$ & \underline{0.00} & 0.00 & 0.00 & \underline{0.00} & 0.00 & 0.00 & 0.00 & 0.00 & 0.00 & 0.00 & 0.00 & \underline{0.08} & 0.01 \\
\midrule
Semantic Entropy (N=10) & $11\times$ & \underline{0.00} & \underline{0.13} & \textbf{0.20} & \textbf{0.00} & 0.14 & 0.00 & \underline{0.04} & \textbf{0.21} & 0.44 & \textbf{0.00} & 0.00 & 0.00 & 0.10 \\
Semantic Entropy (N=5) & $6\times$ & \underline{0.00} & 0.00 & 0.00 & \underline{0.00} & 0.00 & 0.00 & 0.00 & 0.00 & \textbf{0.45} & \underline{0.00} & 0.00 & 0.00 & 0.04 \\
VASE (N=5) & $6\times$ & \underline{0.00} & \textbf{0.15} & 0.17 & \underline{0.00} & 0.14 & 0.00 & 0.00 & 0.17 & 0.43 & 0.00 & 0.00 & 0.00 & 0.09 \\
VL-Uncertainty (N=10) & $11\times$ & \underline{0.00} & 0.02 & 0.16 & \underline{0.00} & 0.13 & 0.38 & 0.03 & \underline{0.20} & \underline{0.44} & 0.00 & \underline{0.00} & 0.00 & 0.11 \\
KLE (N=10) & $11\times$ & \underline{0.00} & 0.00 & \underline{0.19} & \underline{0.00} & 0.14 & 0.00 & 0.03 & 0.17 & 0.42 & 0.00 & 0.00 & 0.00 & 0.08 \\
\midrule
\multicolumn{15}{l}{\textbf{CHDA $\uparrow$}} \\
\midrule
Avg Entropy & $1\times$ & 47.0 & 49.7 & 49.7 & 71.2 & 58.7 & 57.6 & 57.2 & 53.6 & 60.8 & 45.9 & 55.3 & 62.7 & 55.8 \\
Max Entropy & $1\times$ & 46.5 & 49.4 & 52.4 & 71.0 & 56.6 & 56.1 & 53.0 & 48.9 & 56.8 & \underline{69.0} & 53.7 & 61.8 & 56.3 \\
Avg NLL & $1\times$ & 46.2 & 49.3 & \underline{57.1} & 71.7 & 57.5 & 58.0 & 56.6 & 53.9 & 60.8 & 33.7 & 55.4 & 62.8 & 55.3 \\
Max NLL & $1\times$ & 47.1 & 48.6 & 54.1 & 72.0 & 56.9 & 56.8 & 54.0 & 48.9 & 56.6 & 62.0 & 53.9 & 62.0 & 56.1 \\
Avg Entropy (T=1) & $2\times$ & 38.7 & \underline{60.6} & 49.4 & 38.3 & 59.5 & 57.3 & 60.2 & 56.1 & \underline{67.3} & 32.7 & 57.4 & 65.9 & 53.6 \\
Max Entropy (T=1) & $2\times$ & 52.0 & 52.4 & 46.9 & 55.4 & 56.0 & 56.5 & 61.1 & \underline{62.1} & 56.8 & 58.7 & 53.6 & 61.9 & 56.1 \\
Avg NLL (T=1) & $2\times$ & 41.3 & 58.5 & 49.9 & 49.9 & 57.0 & 55.9 & \textbf{66.5} & 53.6 & 64.8 & 33.5 & 57.5 & 64.7 & 54.4 \\
Max NLL (T=1) & $2\times$ & 57.0 & 48.8 & 55.8 & \textbf{73.9} & 55.5 & 56.1 & 63.3 & 48.5 & 56.6 & 41.9 & 53.7 & 61.8 & 56.1 \\
\midrule
Verbal Abstain & $1\times$ & \textbf{68.8} & 41.3 & 41.4 & 37.2 & 49.5 & 52.3 & 40.5 & 59.9 & 55.8 & 41.9 & 55.8 & 61.3 & 50.5 \\
P(true) & $2\times$ & \underline{67.7} & \textbf{64.9} & \textbf{68.9} & \underline{73.5} & \textbf{66.8} & \textbf{70.8} & \underline{65.4} & 56.5 & 65.1 & \textbf{69.3} & \underline{64.6} & \underline{65.9} & \textbf{66.6} \\
Verbal Confidence & $1\times$ & 45.7 & 49.3 & 49.4 & 65.8 & 58.2 & 63.2 & 59.0 & 50.6 & 59.4 & 47.2 & 54.5 & 62.9 & 55.4 \\
\midrule
Semantic Entropy (N=10) & $11\times$ & 37.1 & 43.0 & 42.0 & 34.1 & 55.3 & 55.7 & 63.6 & 52.5 & 63.7 & 30.3 & 54.6 & 62.3 & 49.5 \\
Semantic Entropy (N=5) & $6\times$ & 41.6 & 49.6 & 44.5 & 34.5 & 56.0 & 56.7 & 56.6 & 53.9 & 65.2 & 31.6 & 54.1 & 64.6 & 50.7 \\
VASE (N=5) & $6\times$ & 33.0 & 43.1 & 41.6 & 34.5 & 54.8 & 56.3 & 60.0 & 48.1 & 57.1 & 30.0 & 53.5 & 61.7 & 47.8 \\
VL-Uncertainty (N=10) & $11\times$ & 33.0 & 42.7 & 42.6 & 34.1 & 54.5 & 55.8 & 49.3 & 56.0 & 56.7 & 30.3 & 53.8 & 61.7 & 47.5 \\
KLE (N=10) & $11\times$ & 48.8 & 53.0 & 55.2 & 46.9 & \underline{62.3} & \underline{63.6} & 62.4 & \textbf{62.5} & \textbf{70.3} & 44.0 & \textbf{64.8} & \textbf{68.4} & \underline{58.5} \\
\bottomrule
\end{tabular}}

\caption{Per-model results on multiple-choice benchmark \textbf{MMMU} (12 MLLMs, mean over three seeds). Token-level $T{=}1$ methods lead AUROC; P(true) gives the best CHDA and coverage; sampling-based semantic methods trail on CHDA at $6$--$11\times$ cost, and coverage stays low for every method because of the low base accuracy.}
\label{tab:MMMU}
\end{table*}

\begin{table*}
\centering
\scriptsize
{\footnotesize\setlength{\tabcolsep}{2.6pt}
\begin{tabular}{lc|ccc|ccc|ccc|ccc|c}
\toprule
Method & Cost & \multicolumn{3}{c|}{InternVL2} & \multicolumn{3}{c|}{InternVL3.5} & \multicolumn{3}{c|}{Qwen2-VL} & \multicolumn{3}{c|}{Qwen3-VL} & Avg \\
 & & 2B & 8B & 26B & 2B & 8B & 38B & 2B & 7B & 72B & 2B & 8B & 32B & \\
\midrule
\texttt{Model Acc} & -- & 34.6 & 48.2 & 52.1 & 55.4 & 66.1 & 73.2 & 43.1 & 46.8 & 64.1 & 52.8 & 62.5 & 65.9 & 55.4 \\
\midrule
\multicolumn{15}{l}{\textbf{AUROC $\uparrow$}} \\
\midrule
Avg Entropy & $1\times$ & 0.66 & 0.64 & 0.64 & 0.66 & 0.68 & 0.67 & 0.65 & 0.63 & 0.70 & 0.67 & 0.63 & 0.65 & 0.66 \\
Max Entropy & $1\times$ & 0.68 & 0.65 & 0.61 & 0.66 & 0.66 & 0.67 & 0.65 & 0.63 & 0.69 & 0.68 & 0.63 & 0.63 & 0.65 \\
Avg NLL & $1\times$ & 0.64 & 0.64 & 0.62 & 0.67 & 0.67 & 0.66 & 0.67 & 0.63 & 0.69 & 0.66 & 0.63 & 0.63 & 0.65 \\
Max NLL & $1\times$ & 0.67 & 0.67 & 0.64 & 0.66 & 0.67 & 0.66 & 0.63 & 0.62 & 0.67 & 0.65 & 0.63 & 0.61 & 0.65 \\
Avg Entropy (T=1) & $2\times$ & 0.85 & 0.77 & 0.78 & \underline{0.73} & 0.73 & 0.74 & 0.69 & 0.77 & 0.72 & 0.69 & 0.68 & 0.73 & 0.74 \\
Max Entropy (T=1) & $2\times$ & \underline{0.86} & \underline{0.77} & 0.77 & 0.72 & \underline{0.74} & 0.75 & 0.72 & \underline{0.77} & 0.75 & 0.67 & 0.72 & 0.70 & \underline{0.74} \\
Avg NLL (T=1) & $2\times$ & 0.80 & 0.74 & 0.73 & 0.72 & 0.72 & 0.74 & 0.66 & 0.73 & 0.69 & 0.66 & 0.67 & 0.71 & 0.71 \\
Max NLL (T=1) & $2\times$ & 0.81 & 0.75 & 0.74 & 0.72 & 0.73 & 0.72 & \underline{0.72} & 0.74 & 0.72 & 0.66 & 0.71 & 0.68 & 0.73 \\
\midrule
Verbal Abstain & $1\times$ & 0.55 & 0.57 & 0.62 & 0.66 & 0.72 & 0.62 & 0.69 & 0.72 & 0.72 & \underline{0.74} & \underline{0.75} & 0.60 & 0.66 \\
P(true) & $2\times$ & 0.51 & 0.73 & 0.61 & 0.70 & 0.61 & 0.65 & 0.57 & 0.55 & 0.65 & 0.63 & 0.66 & 0.71 & 0.63 \\
Verbal Confidence & $1\times$ & 0.62 & 0.63 & 0.55 & 0.62 & 0.62 & 0.66 & 0.55 & 0.59 & 0.61 & 0.56 & 0.59 & 0.70 & 0.61 \\
\midrule
Semantic Entropy (N=10) & $11\times$ & 0.74 & 0.73 & 0.76 & 0.70 & 0.61 & 0.74 & 0.66 & 0.73 & 0.77 & 0.71 & 0.62 & 0.73 & 0.71 \\
Semantic Entropy (N=5) & $6\times$ & 0.73 & 0.72 & 0.74 & 0.70 & 0.60 & 0.71 & 0.62 & 0.70 & 0.73 & 0.69 & 0.60 & 0.71 & 0.69 \\
VASE (N=5) & $6\times$ & 0.76 & 0.73 & \underline{0.81} & 0.70 & 0.61 & \textbf{0.79} & 0.66 & 0.72 & 0.83 & 0.69 & 0.61 & \textbf{0.81} & 0.73 \\
VL-Uncertainty (N=10) & $11\times$ & 0.76 & 0.72 & 0.79 & 0.65 & 0.56 & 0.76 & 0.70 & 0.72 & \underline{0.83} & 0.70 & 0.66 & 0.78 & 0.72 \\
KLE (N=10) & $11\times$ & \textbf{0.87} & \textbf{0.84} & \textbf{0.82} & \textbf{0.80} & \textbf{0.79} & \underline{0.78} & \textbf{0.82} & \textbf{0.83} & \textbf{0.84} & \textbf{0.76} & \textbf{0.78} & \underline{0.80} & \textbf{0.81} \\
\midrule
\multicolumn{15}{l}{\textbf{Cov@20 $\uparrow$}} \\
\midrule
Avg Entropy & $1\times$ & 0.00 & 0.00 & 0.00 & 0.00 & 0.00 & 0.82 & 0.00 & 0.00 & 0.00 & 0.00 & 0.00 & 0.00 & 0.07 \\
Max Entropy & $1\times$ & 0.00 & 0.00 & 0.00 & 0.00 & 0.00 & 0.80 & 0.00 & 0.00 & 0.00 & 0.00 & 0.00 & 0.00 & 0.07 \\
Avg NLL & $1\times$ & 0.00 & 0.00 & 0.00 & 0.00 & 0.00 & 0.84 & 0.00 & 0.00 & 0.00 & 0.00 & 0.00 & 0.00 & 0.07 \\
Max NLL & $1\times$ & 0.00 & 0.00 & 0.00 & 0.00 & 0.00 & 0.79 & 0.00 & 0.00 & 0.00 & 0.00 & 0.00 & 0.00 & 0.07 \\
Avg Entropy (T=1) & $2\times$ & 0.21 & 0.31 & 0.36 & \underline{0.38} & \underline{0.59} & 0.79 & 0.07 & 0.21 & 0.50 & 0.00 & 0.39 & 0.51 & 0.36 \\
Max Entropy (T=1) & $2\times$ & \underline{0.26} & \underline{0.33} & 0.37 & 0.34 & 0.54 & 0.80 & 0.18 & 0.21 & 0.54 & 0.00 & 0.42 & 0.53 & 0.38 \\
Avg NLL (T=1) & $2\times$ & 0.21 & 0.26 & 0.34 & 0.36 & 0.57 & 0.76 & 0.08 & 0.17 & 0.45 & 0.00 & 0.41 & 0.51 & 0.34 \\
Max NLL (T=1) & $2\times$ & 0.24 & 0.29 & 0.30 & 0.37 & 0.53 & 0.76 & \underline{0.21} & 0.21 & 0.47 & 0.00 & 0.38 & 0.49 & 0.35 \\
\midrule
Verbal Abstain & $1\times$ & 0.00 & 0.00 & 0.00 & 0.00 & 0.00 & 0.00 & 0.00 & 0.00 & 0.00 & 0.00 & 0.00 & 0.00 & 0.00 \\
P(true) & $2\times$ & 0.03 & 0.16 & 0.19 & 0.30 & 0.31 & 0.67 & 0.06 & 0.05 & 0.33 & 0.10 & 0.26 & 0.49 & 0.25 \\
Verbal Confidence & $1\times$ & 0.00 & 0.00 & 0.00 & 0.00 & 0.00 & 0.59 & 0.00 & 0.00 & 0.00 & 0.00 & 0.00 & 0.25 & 0.07 \\
\midrule
Semantic Entropy (N=10) & $11\times$ & 0.23 & 0.22 & 0.37 & 0.31 & 0.44 & 0.82 & 0.17 & \underline{0.30} & 0.63 & \underline{0.16} & 0.22 & 0.61 & 0.37 \\
Semantic Entropy (N=5) & $6\times$ & 0.19 & 0.26 & 0.35 & 0.35 & 0.39 & 0.82 & 0.12 & 0.27 & 0.58 & 0.11 & 0.00 & 0.61 & 0.34 \\
VASE (N=5) & $6\times$ & 0.21 & 0.21 & \underline{0.40} & 0.36 & 0.44 & \underline{0.87} & 0.15 & 0.29 & \underline{0.69} & 0.07 & 0.30 & \underline{0.73} & \underline{0.39} \\
VL-Uncertainty (N=10) & $11\times$ & 0.17 & 0.22 & \textbf{0.41} & 0.25 & 0.30 & 0.83 & 0.15 & 0.27 & 0.67 & \textbf{0.21} & \underline{0.46} & 0.69 & 0.39 \\
KLE (N=10) & $11\times$ & \textbf{0.26} & \textbf{0.39} & 0.38 & \textbf{0.47} & \textbf{0.71} & \textbf{0.90} & \textbf{0.31} & \textbf{0.37} & \textbf{0.74} & 0.00 & \textbf{0.66} & \textbf{0.75} & \textbf{0.49} \\
\midrule
\multicolumn{15}{l}{\textbf{CHDA $\uparrow$}} \\
\midrule
Avg Entropy & $1\times$ & 64.2 & 63.0 & 61.2 & 64.5 & 71.3 & 73.7 & 63.8 & 56.0 & 71.1 & 62.1 & 64.4 & 67.4 & 65.2 \\
Max Entropy & $1\times$ & 63.9 & 64.3 & 56.0 & 65.6 & 67.0 & 73.2 & 60.7 & 47.4 & 64.2 & 64.9 & 63.8 & 65.7 & 63.1 \\
Avg NLL & $1\times$ & 63.0 & 64.5 & 61.6 & 65.1 & 70.0 & 73.1 & 63.8 & 55.4 & 69.0 & 55.5 & 64.2 & 68.7 & 64.5 \\
Max NLL & $1\times$ & 63.0 & 65.2 & 58.3 & 66.7 & 66.4 & 73.1 & 61.3 & 47.9 & 63.9 & 62.2 & 62.4 & 65.6 & 63.0 \\
Avg Entropy (T=1) & $2\times$ & 79.2 & 68.3 & 63.1 & 60.9 & 67.7 & 73.7 & 64.6 & 65.0 & 68.8 & 61.4 & 64.2 & 67.9 & 67.1 \\
Max Entropy (T=1) & $2\times$ & \textbf{81.8} & 70.7 & 57.2 & 64.8 & 69.3 & 73.9 & 68.3 & \underline{70.4} & 68.5 & 62.9 & 63.3 & 66.8 & 68.2 \\
Avg NLL (T=1) & $2\times$ & 74.8 & 68.3 & 61.6 & 61.8 & 67.3 & 73.4 & 63.9 & 59.9 & 65.1 & 59.3 & 63.3 & 67.1 & 65.5 \\
Max NLL (T=1) & $2\times$ & 79.4 & \underline{71.2} & 61.8 & \underline{68.3} & 67.7 & 73.2 & \underline{70.6} & 55.6 & 66.7 & 61.3 & 63.8 & 66.4 & 67.2 \\
\midrule
Verbal Abstain & $1\times$ & 60.9 & 53.4 & 61.0 & 63.0 & \underline{73.2} & 75.5 & 58.0 & 68.2 & \textbf{76.9} & \underline{70.2} & \textbf{77.8} & 72.0 & 67.5 \\
P(true) & $2\times$ & 59.2 & 65.4 & 57.6 & 62.1 & 62.8 & 70.9 & 58.9 & 53.7 & 65.7 & 57.2 & 63.9 & 66.2 & 62.0 \\
Verbal Confidence & $1\times$ & 58.9 & 61.9 & 56.0 & 56.9 & 64.7 & 70.5 & 51.5 & 59.2 & 60.9 & 53.2 & 63.6 & 68.8 & 60.5 \\
\midrule
Semantic Entropy (N=10) & $11\times$ & 74.9 & 68.5 & 65.6 & 65.0 & 61.5 & 79.2 & 67.9 & 68.3 & 70.0 & 69.4 & 63.6 & 70.8 & 68.7 \\
Semantic Entropy (N=5) & $6\times$ & 75.8 & 67.5 & 67.3 & 64.5 & 60.9 & 74.8 & 66.0 & 67.6 & 66.4 & 69.7 & 63.0 & 69.9 & 67.8 \\
VASE (N=5) & $6\times$ & 76.0 & 68.3 & \underline{70.6} & 66.5 & 61.2 & \textbf{80.7} & 67.7 & 67.3 & \underline{76.8} & 68.7 & 63.3 & \underline{75.1} & \underline{70.2} \\
VL-Uncertainty (N=10) & $11\times$ & 75.8 & 67.6 & 70.2 & 59.7 & 55.7 & 76.5 & 68.8 & 66.7 & 75.8 & 67.8 & 66.4 & 74.2 & 68.8 \\
KLE (N=10) & $11\times$ & \underline{80.1} & \textbf{77.1} & \textbf{74.8} & \textbf{73.9} & \textbf{77.4} & \underline{80.4} & \textbf{77.8} & \textbf{73.7} & 75.1 & \textbf{70.3} & \underline{77.1} & \textbf{78.6} & \textbf{76.3} \\
\bottomrule
\end{tabular}}

\caption{Per-model results on short open-ended benchmark \textbf{MM-Vet} (12 MLLMs, mean over three seeds). KLE leads every metric; Max Entropy ($T{=}1$) is the best cheap option; verbalized abstention is frequent but only weakly correlated with correctness, and P(true) is weak on single-word answers.}
\label{tab:mmvet}
\end{table*}

\begin{table*}
\centering
\tiny
{\footnotesize\setlength{\tabcolsep}{2.6pt}
\begin{tabular}{lc|ccc|ccc|ccc|ccc|c}
\toprule
Method & Cost & \multicolumn{3}{c|}{InternVL2} & \multicolumn{3}{c|}{InternVL3.5} & \multicolumn{3}{c|}{Qwen2-VL} & \multicolumn{3}{c|}{Qwen3-VL} & Avg \\
 & & 2B & 8B & 26B & 2B & 8B & 38B & 2B & 7B & 72B & 2B & 8B & 32B & \\
\midrule
\texttt{Model Acc} & -- & 70.0 & 68.5 & 71.8 & 69.5 & 70.7 & 78.2 & 47.3 & 64.5 & 64.3 & 68.8 & 75.5 & 74.8 & 68.7 \\
\midrule
\multicolumn{15}{l}{\textbf{AUROC $\uparrow$}} \\
\midrule
Avg Entropy & $1\times$ & 0.64 & 0.55 & 0.59 & 0.58 & 0.57 & 0.61 & 0.58 & 0.60 & 0.64 & 0.60 & 0.60 & 0.64 & 0.60 \\
Max Entropy & $1\times$ & 0.62 & 0.56 & 0.56 & 0.60 & 0.58 & 0.66 & 0.63 & 0.61 & 0.59 & 0.64 & 0.55 & 0.59 & 0.60 \\
Avg NLL & $1\times$ & 0.62 & 0.56 & 0.56 & 0.62 & 0.55 & 0.63 & 0.58 & 0.65 & 0.63 & 0.55 & 0.59 & 0.62 & 0.60 \\
Max NLL & $1\times$ & 0.61 & 0.55 & 0.58 & 0.60 & 0.57 & 0.60 & 0.64 & 0.60 & 0.59 & 0.60 & 0.57 & 0.59 & 0.59 \\
Avg Entropy (T=1) & $2\times$ & 0.63 & 0.66 & 0.62 & 0.63 & 0.59 & 0.57 & 0.64 & \underline{0.72} & 0.73 & 0.67 & 0.65 & 0.65 & 0.65 \\
Max Entropy (T=1) & $2\times$ & 0.61 & 0.66 & 0.65 & 0.66 & 0.60 & 0.63 & 0.61 & 0.66 & 0.66 & 0.64 & 0.61 & 0.66 & 0.64 \\
Avg NLL (T=1) & $2\times$ & 0.60 & 0.62 & 0.64 & 0.62 & 0.61 & 0.57 & 0.59 & 0.68 & 0.70 & \underline{0.71} & 0.64 & 0.65 & 0.64 \\
Max NLL (T=1) & $2\times$ & 0.58 & 0.63 & 0.61 & 0.60 & 0.59 & 0.58 & 0.56 & 0.64 & 0.63 & 0.65 & 0.59 & 0.63 & 0.61 \\
\midrule
Verbal Abstain & $1\times$ & 0.57 & 0.53 & 0.61 & 0.67 & 0.63 & 0.59 & \textbf{0.83} & 0.69 & \underline{0.85} & \textbf{0.72} & 0.70 & \underline{0.67} & 0.67 \\
P(true) & $2\times$ & \underline{0.68} & \textbf{0.76} & \textbf{0.74} & \textbf{0.69} & 0.70 & 0.66 & \underline{0.83} & \textbf{0.75} & \textbf{0.86} & 0.63 & 0.68 & 0.61 & \textbf{0.72} \\
Verbal Confidence & $1\times$ & 0.52 & 0.62 & 0.63 & 0.53 & 0.58 & \underline{0.71} & 0.58 & 0.66 & 0.69 & 0.60 & 0.60 & \textbf{0.73} & 0.62 \\
\midrule
Semantic Entropy (N=10) & $11\times$ & 0.63 & 0.63 & 0.64 & 0.61 & \underline{0.73} & 0.69 & 0.58 & 0.59 & 0.63 & 0.67 & \underline{0.71} & 0.55 & 0.64 \\
Semantic Entropy (N=5) & $6\times$ & 0.60 & 0.63 & 0.62 & 0.61 & 0.70 & 0.68 & 0.57 & 0.61 & 0.60 & 0.64 & 0.65 & 0.55 & 0.62 \\
VASE (N=5) & $6\times$ & 0.61 & 0.67 & \underline{0.69} & 0.63 & 0.73 & 0.69 & 0.63 & 0.65 & 0.62 & 0.66 & 0.66 & 0.57 & 0.65 \\
VL-Uncertainty (N=10) & $11\times$ & 0.62 & 0.62 & 0.55 & 0.61 & 0.66 & 0.67 & 0.61 & 0.60 & 0.62 & 0.63 & 0.62 & 0.56 & 0.61 \\
KLE (N=10) & $11\times$ & \textbf{0.68} & \underline{0.70} & 0.68 & \underline{0.68} & \textbf{0.76} & \textbf{0.73} & 0.62 & 0.64 & 0.67 & 0.69 & \textbf{0.72} & 0.56 & \underline{0.68} \\
\midrule
\multicolumn{15}{l}{\textbf{Cov@20 $\uparrow$}} \\
\midrule
Avg Entropy & $1\times$ & 0.47 & 0.00 & 0.44 & 0.36 & 0.16 & 0.82 & 0.00 & 0.00 & 0.00 & 0.28 & 0.76 & 0.58 & 0.32 \\
Max Entropy & $1\times$ & 0.24 & 0.00 & 0.41 & 0.36 & 0.00 & 0.81 & 0.00 & 0.00 & 0.00 & 0.13 & 0.16 & 0.23 & 0.19 \\
Avg NLL & $1\times$ & 0.25 & 0.00 & 0.14 & 0.36 & 0.27 & 0.81 & 0.00 & 0.00 & 0.00 & 0.21 & 0.66 & 0.56 & 0.27 \\
Max NLL & $1\times$ & 0.39 & 0.00 & 0.00 & 0.36 & 0.10 & 0.69 & 0.00 & 0.00 & 0.00 & 0.34 & 0.65 & 0.51 & 0.25 \\
Avg Entropy (T=1) & $2\times$ & 0.37 & 0.44 & 0.53 & 0.25 & 0.32 & 0.79 & 0.20 & 0.48 & 0.52 & 0.54 & 0.83 & 0.66 & 0.50 \\
Max Entropy (T=1) & $2\times$ & 0.36 & 0.42 & 0.51 & 0.51 & 0.38 & 0.89 & 0.14 & 0.37 & 0.33 & 0.40 & 0.64 & 0.73 & 0.47 \\
Avg NLL (T=1) & $2\times$ & 0.31 & 0.34 & 0.46 & 0.33 & 0.45 & 0.68 & 0.16 & 0.37 & 0.46 & 0.58 & 0.77 & 0.73 & 0.47 \\
Max NLL (T=1) & $2\times$ & 0.24 & 0.44 & 0.29 & 0.19 & 0.35 & 0.74 & 0.13 & 0.26 & 0.27 & 0.49 & 0.69 & 0.62 & 0.39 \\
\midrule
Verbal Abstain & $1\times$ & 0.00 & 0.00 & 0.00 & 0.00 & 0.00 & 0.00 & \textbf{0.72} & \textbf{0.91} & 0.72 & 0.00 & 0.87 & \underline{0.89} & 0.34 \\
P(true) & $2\times$ & \underline{0.48} & \textbf{0.68} & \textbf{0.77} & \textbf{0.54} & \underline{0.72} & 0.88 & \underline{0.49} & 0.62 & \underline{0.76} & 0.58 & 0.74 & 0.54 & \textbf{0.65} \\
Verbal Confidence & $1\times$ & 0.00 & 0.00 & 0.33 & 0.00 & 0.00 & 0.75 & 0.00 & \underline{0.85} & \textbf{0.92} & 0.00 & \textbf{0.96} & \textbf{0.92} & 0.39 \\
\midrule
Semantic Entropy (N=10) & $11\times$ & 0.40 & 0.32 & 0.52 & 0.25 & 0.65 & 0.91 & 0.10 & 0.15 & 0.34 & 0.53 & \underline{0.88} & 0.30 & 0.44 \\
Semantic Entropy (N=5) & $6\times$ & 0.34 & 0.39 & 0.55 & 0.34 & 0.62 & 0.83 & 0.04 & 0.20 & 0.14 & 0.49 & 0.77 & 0.29 & 0.42 \\
VASE (N=5) & $6\times$ & 0.33 & 0.53 & 0.61 & 0.23 & 0.67 & 0.88 & 0.12 & 0.31 & 0.23 & \underline{0.59} & 0.82 & 0.50 & 0.49 \\
VL-Uncertainty (N=10) & $11\times$ & 0.38 & 0.43 & 0.15 & 0.36 & 0.57 & \underline{0.91} & 0.07 & 0.23 & 0.23 & 0.42 & 0.58 & 0.42 & 0.40 \\
KLE (N=10) & $11\times$ & \textbf{0.52} & \underline{0.62} & \underline{0.70} & \underline{0.52} & \textbf{0.79} & \textbf{0.94} & 0.14 & 0.29 & 0.37 & \textbf{0.60} & 0.86 & 0.36 & \underline{0.56} \\
\midrule
\multicolumn{15}{l}{\textbf{CHDA $\uparrow$}} \\
\midrule
Avg Entropy & $1\times$ & 52.5 & 51.7 & 62.8 & 53.0 & 60.6 & \underline{77.7} & 62.7 & 67.2 & 67.3 & 57.0 & 71.8 & 70.5 & 62.9 \\
Max Entropy & $1\times$ & 51.3 & 52.0 & 62.8 & 51.8 & 70.2 & \textbf{78.2} & 60.3 & 65.1 & 64.3 & 32.7 & 69.6 & 75.3 & 61.1 \\
Avg NLL & $1\times$ & 52.2 & 53.3 & 55.0 & 52.6 & 64.5 & 74.3 & 62.1 & 68.2 & 65.3 & 65.8 & 73.4 & 67.2 & 62.8 \\
Max NLL & $1\times$ & 52.3 & 50.5 & 61.8 & 52.2 & 68.7 & 76.0 & 61.0 & 65.2 & 64.5 & 39.8 & 73.1 & 69.3 & 61.2 \\
Avg Entropy (T=1) & $2\times$ & 46.3 & 56.3 & 66.3 & 69.9 & 66.5 & 77.0 & 60.4 & 67.2 & 68.0 & 68.4 & 75.0 & 74.3 & 66.3 \\
Max Entropy (T=1) & $2\times$ & 39.9 & 50.9 & 71.6 & 66.4 & 70.4 & 76.5 & 59.3 & 53.2 & 66.5 & 44.3 & 73.6 & 74.6 & 62.3 \\
Avg NLL (T=1) & $2\times$ & 44.5 & 52.8 & 67.2 & 57.1 & 70.6 & 77.3 & 62.8 & 66.7 & 67.8 & \underline{72.7} & 75.8 & 76.5 & 66.0 \\
Max NLL (T=1) & $2\times$ & 38.5 & 47.6 & 57.3 & 40.6 & 69.2 & \underline{77.7} & 60.4 & 62.8 & 66.0 & 57.3 & 71.6 & 75.3 & 60.4 \\
\midrule
Verbal Abstain & $1\times$ & 32.4 & \underline{70.3} & \textbf{77.3} & \textbf{73.3} & \textbf{75.9} & 76.7 & \textbf{86.0} & \textbf{85.0} & \textbf{88.2} & \textbf{81.0} & \textbf{83.0} & \textbf{82.4} & \textbf{76.0} \\
P(true) & $2\times$ & \textbf{62.3} & \textbf{71.5} & \underline{76.7} & 67.0 & 66.3 & 73.5 & \underline{82.7} & 68.8 & 73.0 & 65.3 & 69.5 & 75.8 & \underline{71.0} \\
Verbal Confidence & $1\times$ & \underline{59.0} & 57.3 & 74.0 & \underline{70.3} & 70.5 & 75.7 & 59.0 & \underline{77.2} & \underline{78.0} & 70.8 & \underline{79.8} & \underline{78.3} & 70.8 \\
\midrule
Semantic Entropy (N=10) & $11\times$ & 56.8 & 66.4 & 69.1 & 66.8 & 72.6 & 73.8 & 56.0 & 58.1 & 58.0 & 65.8 & 72.6 & 63.8 & 65.0 \\
Semantic Entropy (N=5) & $6\times$ & 51.3 & 54.5 & 61.8 & 65.3 & 70.4 & 67.7 & 56.2 & 57.7 & 54.3 & 63.8 & 68.9 & 55.4 & 60.6 \\
VASE (N=5) & $6\times$ & 55.3 & 64.8 & 68.0 & 69.0 & 71.4 & 73.0 & 56.9 & 64.0 & 61.3 & 67.5 & 70.3 & 64.0 & 65.5 \\
VL-Uncertainty (N=10) & $11\times$ & 56.7 & 62.3 & 58.2 & 66.9 & 67.3 & 74.8 & 58.0 & 55.7 & 64.3 & 64.8 & 65.9 & 61.7 & 63.1 \\
KLE (N=10) & $11\times$ & 58.7 & 66.1 & 65.1 & 63.6 & \underline{73.6} & 65.2 & 60.3 & 59.4 & 58.7 & 63.2 & 71.2 & 58.9 & 63.7 \\
\bottomrule
\end{tabular}}

\caption{Per-model results on sentence-length free-form benchmark \textbf{FSVQA200} (12 MLLMs, mean over three seeds). P(true) leads AUROC and coverage and Verbal Abstain leads CHDA, both at low cost; token-level methods lose the advantage they hold on short answers, while semantic methods remain competitive at $6$--$11\times$ cost.}
\label{tab:fsvqa200}
\end{table*}

\begin{table*}
\centering
\tiny
{\footnotesize\setlength{\tabcolsep}{2.6pt}
\begin{tabular}{lc|ccc|ccc|ccc|ccc|c}
\toprule
Method & Cost & \multicolumn{3}{c|}{InternVL2} & \multicolumn{3}{c|}{InternVL3.5} & \multicolumn{3}{c|}{Qwen2-VL} & \multicolumn{3}{c|}{Qwen3-VL} & Avg \\
 & & 2B & 8B & 26B & 2B & 8B & 38B & 2B & 7B & 72B & 2B & 8B & 32B & \\
\midrule
\texttt{Model Acc} & -- & 39.3 & 48.3 & 56.0 & 55.0 & 67.3 & 69.0 & 44.3 & 57.6 & 62.8 & 45.1 & 53.1 & 57.3 & 54.6 \\
\midrule
\multicolumn{15}{l}{\textbf{AUROC $\uparrow$}} \\
\midrule
Avg Entropy & $1\times$ & 0.56 & 0.59 & 0.62 & 0.47 & 0.55 & 0.52 & 0.51 & 0.53 & 0.55 & 0.54 & 0.52 & 0.48 & 0.54 \\
Max Entropy & $1\times$ & 0.59 & 0.66 & 0.61 & 0.57 & 0.59 & 0.55 & 0.61 & 0.57 & 0.58 & 0.54 & 0.51 & 0.49 & 0.57 \\
Avg NLL & $1\times$ & 0.56 & 0.56 & 0.59 & 0.48 & 0.57 & 0.55 & 0.50 & 0.55 & 0.54 & 0.53 & 0.51 & 0.47 & 0.54 \\
Max NLL & $1\times$ & 0.59 & 0.65 & 0.62 & 0.54 & 0.59 & 0.55 & 0.59 & 0.56 & 0.59 & 0.50 & 0.51 & 0.48 & 0.56 \\
Avg Entropy (T=1) & $2\times$ & 0.56 & 0.58 & 0.53 & 0.44 & 0.52 & 0.47 & 0.58 & 0.57 & 0.51 & 0.53 & 0.47 & 0.48 & 0.52 \\
Max Entropy (T=1) & $2\times$ & 0.68 & 0.65 & 0.62 & 0.52 & 0.58 & 0.52 & 0.59 & 0.57 & 0.57 & 0.55 & 0.48 & 0.50 & 0.57 \\
Avg NLL (T=1) & $2\times$ & 0.55 & 0.59 & 0.56 & 0.45 & 0.55 & 0.48 & 0.55 & 0.56 & 0.53 & 0.53 & 0.49 & 0.47 & 0.53 \\
Max NLL (T=1) & $2\times$ & 0.62 & 0.62 & 0.58 & 0.55 & 0.55 & 0.50 & 0.61 & 0.58 & 0.56 & 0.49 & 0.47 & 0.48 & 0.55 \\
\midrule
Verbal Abstain & $1\times$ & 0.51 & 0.51 & 0.52 & 0.51 & 0.52 & 0.55 & 0.62 & 0.56 & 0.53 & 0.51 & 0.51 & 0.50 & 0.53 \\
P(true) & $2\times$ & 0.66 & \underline{0.75} & \underline{0.75} & \textbf{0.75} & 0.67 & \textbf{0.72} & 0.68 & 0.73 & \textbf{0.78} & 0.73 & \textbf{0.74} & \textbf{0.83} & \underline{0.73} \\
Verbal Confidence & $1\times$ & 0.62 & 0.67 & 0.62 & 0.68 & 0.62 & \underline{0.71} & 0.58 & 0.63 & 0.63 & 0.60 & 0.63 & 0.67 & 0.64 \\
\midrule
Semantic Entropy (N=10) & $11\times$ & \underline{0.80} & 0.71 & 0.74 & 0.70 & 0.73 & 0.67 & 0.76 & 0.73 & 0.73 & \underline{0.74} & 0.70 & 0.67 & 0.72 \\
Semantic Entropy (N=5) & $6\times$ & 0.72 & 0.71 & 0.72 & 0.69 & 0.68 & 0.65 & 0.71 & 0.69 & 0.69 & 0.68 & 0.68 & 0.67 & 0.69 \\
VASE (N=5) & $6\times$ & 0.80 & 0.75 & 0.73 & 0.71 & \underline{0.74} & 0.66 & \underline{0.78} & 0.72 & 0.72 & 0.72 & 0.71 & 0.68 & 0.73 \\
VL-Uncertainty (N=10) & $11\times$ & 0.72 & 0.73 & 0.74 & 0.70 & 0.72 & 0.71 & 0.73 & \underline{0.74} & 0.73 & 0.70 & 0.71 & \underline{0.70} & 0.72 \\
KLE (N=10) & $11\times$ & \textbf{0.84} & \textbf{0.75} & \textbf{0.77} & \underline{0.71} & \textbf{0.74} & 0.71 & \textbf{0.80} & \textbf{0.77} & \underline{0.75} & \textbf{0.75} & \underline{0.71} & 0.70 & \textbf{0.75} \\
\midrule
\multicolumn{15}{l}{\textbf{Cov@20 $\uparrow$}} \\
\midrule
Avg Entropy & $1\times$ & 0.00 & 0.00 & 0.00 & 0.03 & 0.17 & 0.06 & 0.00 & 0.00 & 0.11 & 0.00 & 0.00 & 0.00 & 0.03 \\
Max Entropy & $1\times$ & 0.00 & 0.00 & 0.17 & 0.06 & 0.19 & 0.00 & 0.00 & 0.00 & 0.18 & 0.00 & 0.00 & 0.00 & 0.05 \\
Avg NLL & $1\times$ & 0.00 & 0.00 & 0.05 & 0.04 & 0.17 & 0.06 & 0.00 & 0.00 & 0.14 & 0.00 & 0.00 & 0.00 & 0.04 \\
Max NLL & $1\times$ & 0.00 & 0.00 & 0.00 & 0.05 & 0.18 & 0.00 & 0.00 & 0.00 & 0.19 & 0.00 & 0.00 & 0.00 & 0.03 \\
Avg Entropy (T=1) & $2\times$ & 0.03 & 0.05 & 0.01 & 0.01 & 0.08 & 0.04 & 0.02 & 0.01 & 0.01 & 0.00 & 0.00 & 0.00 & 0.02 \\
Max Entropy (T=1) & $2\times$ & 0.07 & 0.13 & 0.09 & 0.02 & 0.09 & 0.05 & 0.05 & 0.07 & 0.09 & 0.00 & 0.01 & 0.00 & 0.06 \\
Avg NLL (T=1) & $2\times$ & 0.02 & 0.04 & 0.02 & 0.00 & 0.12 & 0.03 & 0.01 & 0.02 & 0.00 & 0.00 & 0.01 & 0.00 & 0.02 \\
Max NLL (T=1) & $2\times$ & 0.04 & 0.11 & 0.04 & 0.03 & 0.15 & 0.07 & 0.02 & 0.07 & 0.08 & 0.00 & 0.00 & 0.00 & 0.05 \\
\midrule
Verbal Abstain & $1\times$ & 0.00 & 0.00 & 0.00 & 0.00 & 0.00 & 0.00 & 0.00 & 0.00 & 0.00 & 0.00 & 0.00 & 0.00 & 0.00 \\
P(true) & $2\times$ & 0.09 & \textbf{0.25} & \underline{0.39} & \textbf{0.40} & 0.48 & \underline{0.63} & 0.02 & 0.15 & \textbf{0.61} & \textbf{0.31} & \textbf{0.21} & \textbf{0.52} & \textbf{0.34} \\
Verbal Confidence & $1\times$ & 0.00 & 0.00 & 0.00 & 0.00 & 0.00 & \textbf{0.80} & 0.00 & 0.00 & 0.00 & 0.00 & 0.06 & 0.26 & 0.09 \\
\midrule
Semantic Entropy (N=10) & $11\times$ & \underline{0.19} & 0.03 & 0.24 & 0.10 & \underline{0.61} & 0.44 & 0.16 & 0.22 & 0.48 & 0.01 & 0.14 & 0.11 & 0.23 \\
Semantic Entropy (N=5) & $6\times$ & 0.09 & 0.09 & 0.24 & 0.05 & 0.41 & 0.37 & 0.15 & 0.28 & 0.40 & 0.00 & 0.07 & 0.00 & 0.18 \\
VASE (N=5) & $6\times$ & 0.13 & 0.08 & 0.25 & \underline{0.14} & 0.55 & 0.49 & \underline{0.18} & 0.30 & 0.40 & \underline{0.10} & 0.04 & \underline{0.28} & 0.25 \\
VL-Uncertainty (N=10) & $11\times$ & 0.08 & \underline{0.16} & 0.26 & 0.12 & 0.55 & 0.58 & 0.05 & \underline{0.33} & 0.49 & 0.00 & 0.08 & 0.21 & 0.24 \\
KLE (N=10) & $11\times$ & \textbf{0.22} & 0.08 & \textbf{0.40} & 0.11 & \textbf{0.62} & 0.62 & \textbf{0.27} & \textbf{0.34} & \underline{0.54} & 0.03 & \underline{0.15} & 0.17 & \underline{0.30} \\
\midrule
\multicolumn{15}{l}{\textbf{CHDA $\uparrow$}} \\
\midrule
Avg Entropy & $1\times$ & 62.4 & 58.4 & 54.7 & 46.8 & 62.3 & 67.8 & 55.4 & 55.4 & 63.8 & 49.8 & 54.7 & 56.6 & 57.3 \\
Max Entropy & $1\times$ & 63.0 & 59.9 & 54.9 & 47.2 & 66.1 & 69.0 & 56.7 & 58.5 & 62.8 & 54.5 & 53.9 & 57.5 & 58.7 \\
Avg NLL & $1\times$ & 62.5 & 59.9 & 54.3 & 45.1 & 61.3 & 70.5 & 54.9 & 57.7 & 62.5 & 48.9 & 55.6 & 53.8 & 57.3 \\
Max NLL & $1\times$ & 62.4 & 59.9 & 57.8 & 46.3 & 64.8 & 67.6 & 56.7 & 56.5 & 63.3 & 54.0 & 53.5 & 56.7 & 58.3 \\
Avg Entropy (T=1) & $2\times$ & 61.9 & 54.2 & 56.7 & 54.0 & 65.9 & 67.7 & 56.6 & 61.3 & 63.5 & 55.1 & 53.9 & 58.6 & 59.1 \\
Max Entropy (T=1) & $2\times$ & 63.0 & 58.7 & 60.3 & 50.3 & 68.0 & 67.8 & 58.5 & 47.2 & 64.3 & 56.0 & 55.5 & 58.4 & 59.0 \\
Avg NLL (T=1) & $2\times$ & 60.9 & 57.6 & 58.1 & 52.1 & 66.0 & 70.3 & 55.5 & 59.4 & 64.2 & 54.9 & 55.8 & 57.6 & 59.4 \\
Max NLL (T=1) & $2\times$ & 62.1 & 57.8 & 52.3 & 44.2 & 66.0 & 67.9 & 56.4 & 57.9 & 62.2 & 52.6 & 55.6 & 57.3 & 57.7 \\
\midrule
Verbal Abstain & $1\times$ & 63.3 & 50.9 & 59.9 & 55.7 & 67.0 & \underline{72.8} & 50.5 & 58.3 & 63.8 & 47.2 & 53.4 & 59.7 & 58.5 \\
P(true) & $2\times$ & 63.6 & 65.7 & 68.0 & \textbf{69.6} & 67.6 & \underline{72.8} & \underline{65.0} & \underline{69.6} & 69.7 & 66.4 & \textbf{67.3} & \textbf{74.0} & \underline{68.3} \\
Verbal Confidence & $1\times$ & 60.1 & 65.0 & 62.5 & 67.9 & 71.3 & \textbf{76.1} & 53.5 & 66.7 & 69.0 & 54.1 & 65.7 & 64.8 & 64.7 \\
\midrule
Semantic Entropy (N=10) & $11\times$ & 72.3 & 66.7 & 68.0 & 67.7 & 71.1 & 70.3 & 59.8 & 63.3 & 67.0 & \underline{68.3} & 65.5 & 63.8 & 67.0 \\
Semantic Entropy (N=5) & $6\times$ & 69.4 & 65.9 & 67.4 & 65.7 & \underline{71.4} & 70.2 & 63.5 & 64.3 & 63.5 & 65.4 & 64.3 & 62.1 & 66.1 \\
VASE (N=5) & $6\times$ & \underline{73.7} & \underline{67.9} & 67.4 & \underline{69.4} & \textbf{72.3} & 70.2 & 60.3 & 67.1 & 68.3 & 65.8 & 66.9 & 62.2 & 67.6 \\
VL-Uncertainty (N=10) & $11\times$ & 69.0 & 67.3 & \underline{69.3} & 66.1 & 69.9 & 72.0 & 60.6 & 63.6 & \textbf{71.9} & 63.3 & \underline{67.2} & \underline{65.9} & 67.2 \\
KLE (N=10) & $11\times$ & \textbf{75.4} & \textbf{69.4} & \textbf{71.2} & 66.2 & 70.5 & 71.6 & \textbf{69.3} & \textbf{70.0} & \underline{70.3} & \textbf{68.3} & 66.1 & 64.5 & \textbf{69.4} \\
\bottomrule
\end{tabular}}

\caption{Per-model results on long open-form benchmark \textbf{MM-Vet Long} (12 MLLMs, mean over three seeds). KLE and P(true) lead on the same questions as MM-Vet answered at length, while the token-level $T{=}1$ methods that were competitive on MM-Vet collapse.}
\label{tab:mmvetlong}
\end{table*}

\begin{table*}
\centering
\tiny
{\footnotesize\setlength{\tabcolsep}{2.6pt}
\begin{tabular}{lc|ccc|ccc|ccc|ccc|c}
\toprule
Method & Cost & \multicolumn{3}{c|}{InternVL2} & \multicolumn{3}{c|}{InternVL3.5} & \multicolumn{3}{c|}{Qwen2-VL} & \multicolumn{3}{c|}{Qwen3-VL} & Avg \\
 & & 2B & 8B & 26B & 2B & 8B & 38B & 2B & 7B & 72B & 2B & 8B & 32B & \\
\midrule
\texttt{Model Acc} & -- & 35.0 & 46.1 & 63.9 & 50.6 & 55.0 & 55.6 & 54.4 & 70.0 & 69.4 & 55.0 & 60.3 & 63.9 & 56.6 \\
\midrule
\multicolumn{15}{l}{\textbf{AUROC $\uparrow$}} \\
\midrule
Avg Entropy & $1\times$ & 0.60 & 0.62 & 0.49 & 0.53 & 0.61 & 0.50 & 0.47 & 0.64 & 0.54 & 0.52 & 0.53 & 0.60 & 0.55 \\
Max Entropy & $1\times$ & 0.52 & 0.54 & 0.54 & 0.50 & 0.69 & 0.48 & 0.57 & 0.59 & 0.59 & 0.45 & 0.61 & 0.52 & 0.55 \\
Avg NLL & $1\times$ & 0.52 & 0.54 & 0.54 & 0.51 & 0.62 & 0.46 & 0.52 & 0.66 & 0.57 & 0.54 & 0.45 & 0.62 & 0.55 \\
Max NLL & $1\times$ & 0.55 & 0.60 & 0.60 & 0.55 & 0.61 & 0.52 & 0.58 & 0.63 & 0.55 & 0.49 & 0.47 & 0.59 & 0.56 \\
Avg Entropy (T=1) & $2\times$ & 0.59 & 0.59 & 0.53 & 0.59 & 0.64 & 0.51 & 0.54 & 0.57 & 0.61 & 0.49 & 0.56 & 0.62 & 0.57 \\
Max Entropy (T=1) & $2\times$ & 0.56 & 0.54 & 0.53 & 0.60 & 0.67 & 0.49 & 0.59 & 0.60 & 0.62 & 0.50 & 0.52 & 0.52 & 0.56 \\
Avg NLL (T=1) & $2\times$ & 0.58 & 0.55 & 0.59 & 0.54 & 0.60 & 0.46 & 0.56 & 0.59 & 0.60 & 0.57 & 0.63 & 0.64 & 0.58 \\
Max NLL (T=1) & $2\times$ & 0.62 & 0.55 & 0.51 & 0.50 & 0.63 & 0.53 & 0.55 & 0.55 & 0.63 & 0.53 & 0.54 & 0.54 & 0.56 \\
\midrule
Verbal Abstain & $1\times$ & 0.49 & 0.50 & 0.51 & 0.52 & 0.52 & 0.52 & 0.61 & 0.65 & 0.55 & 0.69 & 0.59 & 0.53 & 0.56 \\
P(true) & $2\times$ & 0.65 & 0.58 & \textbf{0.78} & 0.64 & 0.76 & 0.73 & \underline{0.73} & 0.75 & 0.71 & 0.71 & \textbf{0.73} & \textbf{0.77} & 0.71 \\
Verbal Confidence & $1\times$ & 0.48 & 0.61 & 0.66 & 0.64 & 0.61 & 0.63 & 0.52 & 0.46 & 0.58 & 0.63 & 0.58 & 0.59 & 0.58 \\
\midrule
Semantic Entropy (N=10) & $11\times$ & 0.74 & \underline{0.77} & 0.66 & 0.73 & 0.77 & \underline{0.77} & 0.63 & \underline{0.77} & 0.72 & \textbf{0.72} & 0.70 & 0.65 & 0.72 \\
Semantic Entropy (N=5) & $6\times$ & 0.70 & 0.71 & 0.63 & 0.64 & 0.75 & \textbf{0.77} & 0.62 & 0.72 & 0.76 & 0.65 & 0.69 & \underline{0.70} & 0.69 \\
VASE (N=5) & $6\times$ & \textbf{0.79} & 0.74 & 0.65 & 0.72 & \textbf{0.79} & 0.75 & 0.65 & 0.73 & \textbf{0.82} & \underline{0.72} & \underline{0.71} & 0.60 & \underline{0.72} \\
VL-Uncertainty (N=10) & $11\times$ & 0.78 & \textbf{0.81} & \underline{0.70} & \underline{0.73} & 0.78 & 0.76 & \textbf{0.75} & \textbf{0.79} & \underline{0.80} & 0.68 & 0.70 & 0.66 & \textbf{0.75} \\
KLE (N=10) & $11\times$ & \underline{0.79} & 0.76 & 0.66 & \textbf{0.73} & \underline{0.78} & 0.75 & 0.67 & 0.74 & 0.73 & 0.71 & 0.66 & 0.67 & 0.72 \\
\midrule
\multicolumn{15}{l}{\textbf{Cov@20 $\uparrow$}} \\
\midrule
Avg Entropy & $1\times$ & 0.00 & 0.11 & 0.10 & 0.08 & 0.14 & 0.00 & 0.16 & 0.52 & 0.18 & 0.00 & 0.00 & 0.06 & 0.11 \\
Max Entropy & $1\times$ & 0.00 & 0.11 & 0.11 & 0.08 & 0.18 & 0.06 & 0.14 & 0.24 & 0.29 & 0.00 & 0.03 & 0.06 & 0.11 \\
Avg NLL & $1\times$ & 0.00 & 0.13 & 0.12 & 0.08 & 0.16 & 0.00 & 0.13 & 0.62 & 0.43 & 0.08 & 0.02 & 0.34 & 0.18 \\
Max NLL & $1\times$ & 0.00 & 0.11 & 0.12 & 0.08 & 0.12 & 0.03 & 0.22 & 0.42 & 0.44 & 0.01 & 0.02 & 0.12 & 0.14 \\
Avg Entropy (T=1) & $2\times$ & 0.08 & 0.04 & 0.07 & 0.04 & 0.18 & 0.07 & 0.10 & 0.24 & 0.53 & 0.00 & 0.06 & 0.31 & 0.14 \\
Max Entropy (T=1) & $2\times$ & 0.08 & 0.09 & 0.09 & 0.11 & 0.16 & 0.09 & 0.11 & 0.38 & 0.34 & 0.00 & 0.01 & 0.12 & 0.13 \\
Avg NLL (T=1) & $2\times$ & 0.04 & 0.04 & 0.16 & 0.02 & 0.17 & 0.04 & 0.14 & 0.28 & 0.36 & 0.14 & 0.12 & 0.28 & 0.15 \\
Max NLL (T=1) & $2\times$ & 0.07 & 0.06 & 0.18 & 0.07 & 0.14 & 0.09 & 0.17 & 0.25 & 0.58 & 0.03 & 0.04 & 0.17 & 0.15 \\
\midrule
Verbal Abstain & $1\times$ & 0.00 & 0.00 & 0.00 & 0.00 & 0.00 & 0.00 & 0.00 & 0.00 & 0.00 & 0.00 & 0.00 & 0.00 & 0.00 \\
P(true) & $2\times$ & 0.07 & 0.01 & \underline{0.47} & 0.00 & 0.36 & 0.23 & \textbf{0.37} & \textbf{0.80} & 0.75 & \underline{0.34} & \textbf{0.46} & \textbf{0.63} & \underline{0.37} \\
Verbal Confidence & $1\times$ & 0.01 & 0.13 & 0.42 & 0.06 & 0.08 & 0.00 & 0.00 & 0.00 & 0.36 & 0.28 & 0.27 & \underline{0.62} & 0.19 \\
\midrule
Semantic Entropy (N=10) & $11\times$ & 0.16 & \underline{0.19} & 0.37 & \underline{0.14} & 0.31 & 0.41 & 0.25 & 0.77 & 0.79 & \textbf{0.35} & 0.31 & 0.46 & \textbf{0.38} \\
Semantic Entropy (N=5) & $6\times$ & 0.08 & 0.04 & 0.23 & 0.08 & \textbf{0.39} & \textbf{0.46} & 0.22 & 0.62 & 0.78 & 0.10 & 0.27 & 0.32 & 0.30 \\
VASE (N=5) & $6\times$ & \underline{0.21} & 0.16 & 0.37 & 0.14 & 0.36 & \textbf{0.46} & 0.17 & 0.75 & 0.79 & 0.19 & \underline{0.41} & 0.18 & 0.35 \\
VL-Uncertainty (N=10) & $11\times$ & 0.14 & \textbf{0.28} & 0.42 & \textbf{0.23} & \underline{0.38} & \underline{0.42} & \underline{0.33} & \underline{0.79} & \textbf{0.86} & 0.08 & 0.17 & 0.34 & 0.37 \\
KLE (N=10) & $11\times$ & \textbf{0.24} & 0.15 & \textbf{0.52} & 0.10 & 0.31 & 0.39 & 0.29 & 0.73 & \underline{0.80} & 0.26 & 0.21 & 0.39 & 0.37 \\
\midrule
\multicolumn{15}{l}{\textbf{CHDA $\uparrow$}} \\
\midrule
Avg Entropy & $1\times$ & 70.0 & 62.8 & 62.2 & 51.4 & 63.3 & 54.4 & 46.1 & 74.4 & 73.3 & 53.9 & 57.2 & 68.3 & 61.5 \\
Max Entropy & $1\times$ & 70.0 & 60.6 & 60.0 & 54.4 & 68.3 & 56.1 & 55.0 & 70.0 & 70.6 & 45.0 & 67.1 & 65.0 & 61.8 \\
Avg NLL & $1\times$ & 70.0 & 60.6 & 53.3 & 53.3 & 57.8 & 59.4 & 50.0 & 70.6 & 71.7 & 59.4 & 60.9 & 67.8 & 61.2 \\
Max NLL & $1\times$ & 68.3 & 62.8 & 57.8 & 54.4 & 56.1 & 55.6 & 52.2 & 67.8 & 72.8 & 49.4 & 61.1 & 67.2 & 60.5 \\
Avg Entropy (T=1) & $2\times$ & 68.9 & 59.2 & 63.3 & 62.6 & 61.5 & 56.7 & 46.1 & 60.6 & 74.4 & 52.2 & 60.6 & 68.3 & 61.2 \\
Max Entropy (T=1) & $2\times$ & 67.2 & 58.3 & 61.1 & 57.8 & 66.7 & 58.9 & 50.6 & 37.8 & 68.3 & 48.9 & 60.7 & 63.9 & 58.3 \\
Avg NLL (T=1) & $2\times$ & 68.9 & 57.2 & 62.2 & 55.6 & 56.1 & 56.1 & 45.6 & 61.4 & 73.9 & 58.9 & 65.8 & 68.3 & 60.8 \\
Max NLL (T=1) & $2\times$ & 65.6 & 58.9 & 49.4 & 54.4 & 62.2 & 57.2 & 44.4 & 56.7 & 73.3 & 56.1 & 60.0 & 63.3 & 58.5 \\
\midrule
Verbal Abstain & $1\times$ & 56.1 & 53.9 & 68.9 & 50.6 & 60.6 & 61.7 & \underline{61.1} & \underline{75.0} & 74.4 & \textbf{73.3} & \textbf{77.8} & \underline{70.4} & 65.3 \\
P(true) & $2\times$ & 71.1 & 63.9 & \textbf{78.3} & 60.6 & 63.3 & 61.7 & \textbf{63.3} & 73.3 & 72.8 & 65.0 & 68.9 & 69.4 & 67.6 \\
Verbal Confidence & $1\times$ & 56.1 & 62.8 & 69.4 & 62.8 & 66.1 & 68.3 & 53.9 & 56.1 & 76.1 & \underline{70.6} & 67.8 & 65.0 & 64.6 \\
\midrule
Semantic Entropy (N=10) & $11\times$ & 69.4 & 70.6 & 68.3 & 66.7 & 72.2 & 64.4 & 45.8 & 71.7 & 69.4 & 66.7 & 63.9 & 67.2 & 66.4 \\
Semantic Entropy (N=5) & $6\times$ & 71.7 & 66.1 & 68.3 & 63.9 & 70.0 & \textbf{72.2} & 55.0 & 67.2 & 71.7 & 63.9 & 69.4 & 68.9 & 67.4 \\
VASE (N=5) & $6\times$ & \textbf{76.1} & 65.0 & 71.1 & 66.1 & \textbf{76.1} & 68.9 & 51.1 & \textbf{75.5} & \underline{77.2} & 68.9 & \underline{70.6} & 67.2 & \underline{69.5} \\
VL-Uncertainty (N=10) & $11\times$ & 74.4 & \textbf{77.2} & \underline{73.3} & \textbf{67.8} & \underline{75.6} & \underline{69.4} & 58.3 & 68.9 & \textbf{78.9} & 66.7 & 67.8 & 68.9 & \textbf{70.6} \\
KLE (N=10) & $11\times$ & \underline{75.0} & \underline{70.6} & 65.6 & \underline{67.2} & 73.3 & \underline{69.4} & 56.4 & 67.8 & 69.4 & 69.4 & 62.8 & \textbf{72.2} & 68.3 \\
\bottomrule
\end{tabular}}

\caption{Per-model results on long-form open-ended benchmark \textbf{LLaVA-Bench} (12 MLLMs, mean over three seeds). Sampling-based semantic methods dominate AUROC and CHDA with the largest gap to token-level methods in our suite; P(true) is the cheap alternative, and verbalized methods remain near chance on AUROC but recover on CHDA for specific backbones.}
\label{tab:llava_bench}
\end{table*}

\subsection{Calibration}
\label{app:ece}
The error-probability map of Sec.~\ref{sec:metrics} is fit on $\mathcal{D}_{\text{cal}}$: the scores are split into 8 quantile bins, each bin is assigned its empirical error rate, and monotonicity is enforced with isotonic regression. One map is fit per model and method and applied unchanged to the full evaluation set; Verbal Abstain is binary and is omitted. Table~\ref{tab:appendix_ece} reports the ECE (10 equal-width bins) and Brier score of the mapped probabilities per answer-length regime, discussed in Sec.~\ref{sec:safety}, next to a constant prediction of the pooled error rate. Since ECE is minimized by such a constant, the Brier score, which also penalizes uninformative predictions, is the more informative of the two. Without the map, P(true) and Verbal Confidence reach an ECE of 0.25 and 0.29, and the remaining methods have no probability scale to evaluate.

\begin{table*}
\small
\centering
\begin{tabular}{lcccccccc}
\toprule
 & \multicolumn{2}{c}{Short} & \multicolumn{2}{c}{Sentence} & \multicolumn{2}{c}{Long} & \multicolumn{2}{c}{All} \\
Method & ECE & Brier & ECE & Brier & ECE & Brier & ECE & Brier \\
\midrule
Verbal Confidence & 0.136 & 0.216 & \textbf{0.097} & 0.197 & 0.089 & 0.222 & 0.114 & 0.215 \\
P(true) & 0.160 & 0.221 & 0.106 & \textbf{0.190} & 0.100 & 0.212 & 0.131 & 0.213 \\
Avg Entropy & 0.143 & 0.209 & 0.134 & 0.227 & \textbf{0.067} & 0.238 & 0.116 & 0.222 \\
Max Entropy & 0.151 & 0.212 & 0.118 & 0.223 & \textbf{0.067} & 0.238 & 0.117 & 0.222 \\
Avg Entropy ($T{=}1$) & 0.113 & 0.189 & 0.154 & 0.225 & 0.082 & 0.241 & \textbf{0.109} & 0.212 \\
Max Entropy ($T{=}1$) & 0.117 & 0.188 & 0.146 & 0.226 & 0.080 & 0.239 & \textbf{0.109} & 0.211 \\
Semantic Entropy ($N{=}10$) & 0.125 & 0.192 & 0.155 & 0.226 & 0.090 & 0.213 & 0.118 & 0.204 \\
Semantic Entropy ($N{=}5$) & 0.129 & 0.196 & 0.157 & 0.230 & 0.081 & 0.217 & 0.118 & 0.209 \\
VASE ($N{=}5$) & 0.128 & 0.193 & 0.150 & 0.226 & 0.089 & 0.211 & 0.119 & 0.205 \\
VL-Uncertainty ($N{=}10$) & 0.124 & 0.195 & 0.144 & 0.229 & 0.092 & 0.208 & 0.117 & 0.205 \\
KLE ($N{=}10$) & \textbf{0.111} & \textbf{0.176} & 0.154 & 0.218 & 0.081 & \textbf{0.202} & \textbf{0.109} & \textbf{0.192} \\
\midrule
\emph{Constant predictor} & 0.147 & 0.230 & 0.101 & 0.223 & 0.060 & 0.243 & 0.110 & 0.233 \\
\bottomrule
\end{tabular}

\caption{Expected calibration error (ECE) and Brier score (lower is better) of the mapped error probability, averaged over 12 MLLMs, three seeds and the benchmarks of each regime. The map is fit on the calibration set $\mathcal{D}_{\text{cal}}$ of \S\ref{sec:metrics}. Verbal Abstain is binary and omitted. Bold: best per column. The last row is a constant prediction of the pooled calibration error rate: it attains the lowest long-form ECE, which shows that ECE after mapping is dominated by the base-rate shift and must be read together with the Brier score, which ranks the methods like AUROC.}
\label{tab:appendix_ece}
\end{table*}

\subsection{Long-form results by model size}
\label{app:size_longform}
Table~\ref{tab:size_longform} splits the long-form results of Tab.~\ref{tab:answer_length} by model size. The advantage of sampling-based semantic methods is confined to small and medium models: KLE at 2B and VL-Uncertainty at 7--8B have the best AUROC  (0.75 each), while P(true) already matches their coverage. At 26--72B the order reverses: P(true) reaches 0.76 AUROC and 0.53 Cov@20 against 0.70--0.73 and 0.40--0.48 for the semantic methods, so for large models the cheapest estimator is also the strongest one on long-form answers. Over all twelve models KLE keeps the higher AUROC (0.74 vs.\ 0.72) and P(true) the higher coverage (0.36 vs.\ 0.33; Tab.~\ref{tab:answer_length}).

\begin{table*}[t]
\small
\centering
\setlength{\tabcolsep}{3.5pt}
\begin{tabular}{l|ccc|ccc|ccc}
\toprule
Method & \multicolumn{3}{c|}{2B ($n{=}4$)} & \multicolumn{3}{c|}{7--8B ($n{=}4$)} & \multicolumn{3}{c}{26--72B ($n{=}4$)} \\
 & AUROC & Cov@20 & CHDA & AUROC & Cov@20 & CHDA & AUROC & Cov@20 & CHDA \\
\midrule
Verbal Abstain & 0.56 & 0.00 & 57.2 & 0.54 & 0.00 & 62.1 & 0.52 & 0.00 & 66.5 \\
P(true) & 0.69 & \textbf{0.20} & 65.6 & 0.71 & \textbf{0.34} & 67.5 & \textbf{0.76} & \textbf{0.53} & \underline{70.8} \\
\midrule
Max Entropy ($T{=}1$) & 0.58 & 0.05 & 56.5 & 0.58 & 0.12 & 56.6 & 0.55 & 0.11 & 62.9 \\
\midrule
Semantic Entropy ($N{=}10$) & 0.73 & 0.17 & 64.6 & \underline{0.74} & 0.32 & 68.1 & 0.70 & 0.41 & 67.3 \\
VASE ($N{=}5$) & \underline{0.74} & 0.16 & \underline{66.4} & 0.73 & \underline{0.33} & \textbf{70.2} & 0.70 & 0.40 & 69.1 \\
VL-Uncertainty ($N{=}10$) & 0.72 & 0.13 & 65.8 & \textbf{0.75} & \textbf{0.34} & \underline{69.7} & \underline{0.73} & 0.45 & \textbf{71.2} \\
KLE ($N{=}10$) & \textbf{0.75} & \underline{0.19} & \textbf{68.4} & \underline{0.74} & \underline{0.33} & 68.8 & 0.72 & \underline{0.48} & 69.3 \\
\bottomrule
\end{tabular}

\caption{Long-form answers (MM-Vet Long and LLaVA-Bench) by model size, averaged over the models of each size bucket and three seeds. \textbf{Bold} / \underline{underlined}: best / second best per column. Sampling-based semantic methods have the best AUROC up to 8B, while P(true) already matches them on coverage; at 26--72B P(true) overtakes them and has the best AUROC and coverage.}
\label{tab:size_longform}
\end{table*}

\subsection{Router variants}
\label{app:router}
Table~\ref{tab:router_variants} reports every routing configuration we evaluated, with the same per-bin calibration as in Sec.~\ref{sec:router}. Three observations. First, the choice of the sentence branch matters most: replacing Verbal Abstain by P(true) on the 6--60-word bin raises Cov@20 from 0.38 to 0.46--0.48, since abstention is binary and cannot rank the answers it does not flag. Second, the long branch is nearly free to choose: P(true), SE and KLE give 0.74 AUROC within 0.01 of each other, so the cheapest one wins on cost (2$\times$ against 5$\times$) and its coverage is slightly higher. Third, the abstain override adds one pass and improves every variant by 0.03--0.05 AUROC and 1.7--2.7 CHDA points, because it catches the sentence-length errors on which the model itself abstains and which no score-based branch ranks highly. The recipe of Sec.~\ref{sec:router} is therefore not a tuned configuration but the cheapest member of a flat optimum.
\begin{table}[t]
\tiny
\centering
\begin{tabular}{lc|ccc}
\toprule
Short / Sentence / Long branch & Cost & AUROC & Cov@20 & CHDA \\
\midrule
KLE ($N{=}10$) & $11\times$ & 0.74 & 0.46 & 70.2 \\
P(true) & $2\times$ & 0.71 & 0.45 & 66.6 \\
Max Entropy ($T{=}1$) & $2\times$ & 0.67 & 0.36 & 64.9 \\
Verbal Abstain & $1\times$ & 0.60 & 0.17 & 66.3 \\
\midrule
Entropy / Abstain / SE & $5.1\times$ & 0.72 & 0.38 & 70.8 \\
\quad + abstain override & $5.9\times$ & 0.75 & 0.38 & 72.5 \\
Entropy / P(true) / SE & $5.1\times$ & 0.74 & 0.46 & 70.5 \\
\quad + abstain override & $6.1\times$ & 0.78 & 0.46 & 73.2 \\
Entropy / P(true) / KLE & $5.1\times$ & 0.74 & 0.47 & 71.5 \\
\quad + abstain override & $6.1\times$ & 0.78 & 0.47 & 73.9 \\
Entropy / P(true) / P(true) & $2.0\times$ & 0.74 & \textbf{0.48} & 72.1 \\
\quad + abstain override & $3.0\times$ & \textbf{0.79} & \textbf{0.48} & \textbf{74.6} \\
\bottomrule
\end{tabular}
\caption{Router variants, averaged over all benchmarks, 12 MLLMs and three seeds. Rows name the estimator of the $\leq$5-word, 6--60-word and $>$60-word bins; the first block repeats the single methods for reference. Cost is the mean number of forward passes per sample. Semantic long branches raise the cost to $5$--$6\times$ without improving over the $3\times$ recipe of Sec.~\ref{sec:router}; the abstain override helps every variant.}
\label{tab:router_variants}
\end{table}

\subsection{Ablation tables}
\label{app:extra_tables}
Table~\ref{tab:avg_vlms_benchmarks} gives, per model, the mean over the three representative methods of Tab.~\ref{tab:vlm_method_avgs}; 
Table~\ref{tab:verbal-conf-stats} the parse-failure and over-confidence rates of Verbal Confidence; Tables~\ref{tab:SE_perturbation_ablation} and \ref{tab:SE_N_ablation} the perturbation and sample-size ablations of Semantic Entropy.

\begin{table}[t]
\centering
\scriptsize
\begin{tabular}{llcccc}
\toprule
Family & Size & Acc & AUROC & CHDA & Cov@20 \\
\midrule
IVL2 & 2B & 49.5 & 0.68 & 65.7 & 0.32 \\
 & 8B & 57.3 & 0.72 & 66.7 & 0.36 \\
 & 26B & 62.9 & 0.73 & 68.5 & 0.46 \\
\midrule
IVL3.5 & 2B & 57.8 & 0.70 & 65.3 & 0.36 \\
 & 8B & 67.4 & 0.72 & 70.8 & 0.52 \\
 & 38B & 70.7 & 0.71 & 72.2 & 0.61 \\
\midrule
Qw2-VL & 2B & 48.0 & 0.67 & 62.8 & 0.20 \\
 & 7B & 61.2 & 0.71 & 63.4 & 0.44 \\
 & 72B & 67.7 & 0.75 & 70.1 & 0.58 \\
\midrule
Qw3-VL & 2B & 54.4 & 0.68 & 62.0 & 0.28 \\
 & 8B & 65.8 & 0.69 & 68.4 & 0.44 \\
 & 32B & 69.1 & 0.69 & 71.2 & 0.48 \\
\bottomrule
\end{tabular}
\caption{Effect of MLLM family, size and generation on uncertainty metrics. Each cell is the mean over benchmarks, three seeds and the overall-best method of each family (P(true), Max Entropy $T{=}1$, KLE; chosen on the all-benchmark average of Tab.~\ref{tab:answer_length}). AUROC varies little across models, CHDA and selective coverage improve with size and generation and track base accuracy.}
\label{tab:avg_vlms_benchmarks}
\end{table}

\begin{table}[t]
\centering
\scriptsize
\resizebox{\linewidth}{!}{
\begin{tabular}{lccc}
\toprule
Method & AUROC $\uparrow$ & Cov@20 $\uparrow$ & CHDA $\uparrow$ \\
\midrule
SE + Textual Perturb. & \underline{0.66} & 0.15 & 62.0 \\
SE + Visual Perturb.  & \textbf{0.67} & \underline{0.18} & \underline{63.3} \\
SE                    & \textbf{0.67} & \textbf{0.19} & \textbf{64.0} \\
VL-Uncertainty        & \textbf{0.67} & 0.14 & \underline{63.3} \\
\bottomrule
\end{tabular}}

\caption{Effect of textual (LLM rephrasing) and visual (blurring) perturbations on Semantic Entropy, averaged across MLLMs on a dataset subset. Neither textual perturbations, visual perturbations, nor their combination with increased temperature (VL-Uncertainty) consistently improve over base SE.}
\label{tab:SE_perturbation_ablation}
\end{table}

\begin{table}[t]
\centering
\scriptsize
\resizebox{\linewidth}{!}{
\begin{tabular}{lcccc}
\toprule
N & Cost & AUROC $\uparrow$ & Cov@20 $\uparrow$ & CHDA $\uparrow$ \\
\midrule
$N=5$  & $6\times$  & 0.66 & 0.18 & 63.3 \\
$N=10$ & $11\times$ & \underline{0.67} & 0.21 & 63.7 \\
$N=15$ & $16\times$ & 0.66 & 0.22 & \textbf{64.5} \\
$N=20$ & $21\times$ & \textbf{0.68} & \textbf{0.27} & \underline{64.0} \\
\bottomrule
\end{tabular}}
\caption{Ablation on the number of samples $N$ for Semantic Entropy, averaged across Qwen-family MLLMs on a subset of benchmarks. Larger $N$ brings diminishing returns relative to its cost.}
\vspace{-0.6cm}
\label{tab:SE_N_ablation}
\end{table}

\begin{table}[t]
\centering
\setlength{\tabcolsep}{3pt}
\small
\begin{tabular}{l S[table-format=2.1] S[table-format=2.1] l S[table-format=2.1] S[table-format=2.1]}
\toprule
MLLM & {PF (\%)} & {OC (\%)} & MLLM & {PF (\%)} & {OC (\%)} \\
\midrule
IVL2-2B & 25.9 & 72.8 & Qw2-2B & 6.3 & 56.2 \\
IVL2-8B & 6.9 & 86.1 & Qw2-7B & 6.2 & 80.8 \\
IVL2-26B & 6.1 & 89.0 & Qw2-72B & 2.7 & 83.6 \\
IVL3.5-2B & 20.9 & 72.4 & Qw3-2B & 3.5 & 90.5 \\
IVL3.5-8B & 5.1 & 89.6 & Qw3-8B & 9.0 & 85.3 \\
IVL3.5-38B & 6.4 & 88.6 & Qw3-32B & 16.3 & 73.9 \\
\midrule
\multicolumn{6}{c}{\textit{Mean:} PF 9.6\% \quad OC 80.7\%} \\
\bottomrule
\end{tabular}
\caption{Failure modes of \emph{Verbal Confidence} prompting across MLLMs (IVL = InternVL, Qw = Qwen-VL), three seeds. \textbf{PF} (Parse Fail) is the percentage of responses where a numeric confidence score could not be extracted; \textbf{OC} (Overconf.) is the percentage of parseable responses assigned the maximum confidence value. Even when parsing succeeds, models default to extreme overconfidence in roughly four out of five cases, and scale does not resolve it.}
\vspace{-0.2cm}
\label{tab:verbal-conf-stats}
\end{table}
\subsection{Length versus format}
\label{app:length_format}
\paragraph{Length axis.} MM-Vet and MM-Vet Long contain the same 218 questions and images; only the instruction changes, from a concise final answer to a complete response. Accuracy is unchanged (55.4\% vs.\ 54.6\%), so only the length of the scored answer varies. The method ranking inverts exactly as the regime averages predict (Tabs.~\ref{tab:mmvet} and \ref{tab:mmvetlong}). Max Entropy ($T{=}1$) falls from 0.74 to 0.57 AUROC and from 0.38 to 0.06 Cov@20, as the per-token signal is diluted over many tokens; Verbal Abstain drops to chance (0.66 to 0.53) because the long-answer instruction suppresses abstention (16.8\% to 3.3\%, Tab.~\ref{tab:abstain_cluster_merged}). P(true) moves the other way, from 0.63 to 0.73 AUROC and from 0.25 to 0.34 Cov@20, since a complete answer gives the model something to judge, and the semantic methods are the most stable (KLE 0.81 to 0.75, Semantic Entropy 0.71 to 0.72).

\paragraph{Format axis.} We re-run ScienceQA and MMMU in an open-ended variant that removes the answer options and asks for the answer itself, scored by the same LLM judge; questions, images and answer length are unchanged (Tab.~\ref{tab:format}). Removing the options makes the questions harder (accuracy 65\% to 36\%), which collapses Cov@20 for every method, but the leading group is the same in both formats: Max Entropy ($T{=}1$), Semantic Entropy and KLE stay within 0.03 AUROC of each other (0.76 / 0.74 / 0.73 multiple choice, 0.79 / 0.80 / 0.81 open-ended), with the semantic methods moving from slightly behind to slightly ahead, whereas lengthening the answer moved Max Entropy by $-0.17$ and P(true) by $+0.10$. Two second-order effects follow the mechanisms of Sec.~\ref{length_claims}: the semantic methods gain the most AUROC (Semantic Entropy 0.74 to 0.80, KLE 0.73 to 0.81) once free-form answers give the entailment judge lexical variation to cluster, and Verbal Abstain recovers from 0.59 to 0.65 once there is no option to commit to. Length, not format, is the axis along which the best estimator changes.

\begin{table}[t]\centering\setlength{\tabcolsep}{3pt}
\tiny
\begin{tabular}{lcccccc}
\toprule
 & \multicolumn{3}{c}{Multiple choice} & \multicolumn{3}{c}{Open-ended} \\
\cmidrule(lr){2-4}\cmidrule(lr){5-7}
Method & AUROC & Cov@20 & HDA$^*$ & AUROC & Cov@20 & HDA$^*$ \\
\midrule
Model accuracy (\%) & \multicolumn{3}{c}{65.3} & \multicolumn{3}{c}{36.4} \\
Verbal Abstain & 0.59 & 0.33 & 73.6 & 0.65 & 0.00 & 72.4 \\
P(true) & 0.72 & \textbf{0.55} & 76.7 & 0.70 & 0.10 & 74.5 \\
Max Entropy & 0.60 & 0.45 & 75.2 & 0.61 & 0.00 & 70.8 \\
Max Entropy ($T{=}1$) & \textbf{0.76} & \textbf{0.55} & \textbf{78.9} & 0.79 & \textbf{0.23} & 79.6 \\
Semantic Entropy ($N{=}10$) & 0.74 & 0.52 & 77.9 & 0.80 & 0.20 & 79.5 \\
KLE ($N{=}10$) & 0.73 & 0.51 & 77.5 & \textbf{0.81} & 0.20 & \textbf{79.7} \\
\bottomrule
\end{tabular}
\caption{Format ablation: ScienceQA and MMMU as multiple choice and as open-ended questions (same questions, images and answer length), averaged over the two benchmarks and 12 MLLMs (base models three seeds, large models seed 0). HDA$^*$ uses the threshold optimal on each set (an upper bound on CHDA), since the open-ended variants have no calibration split. Removing the options lowers accuracy and coverage for every method; the same three methods lead both formats within 0.03 AUROC, with the semantic methods gaining from free-form answers, and no method changes tier.}
\label{tab:format}
\end{table}

\subsection{Qualitative Results}
\label{sec:qualitative}

\label{sec:qualitative_short}
\begin{figure*}
    \centering
    \includegraphics[width=0.9\textwidth]{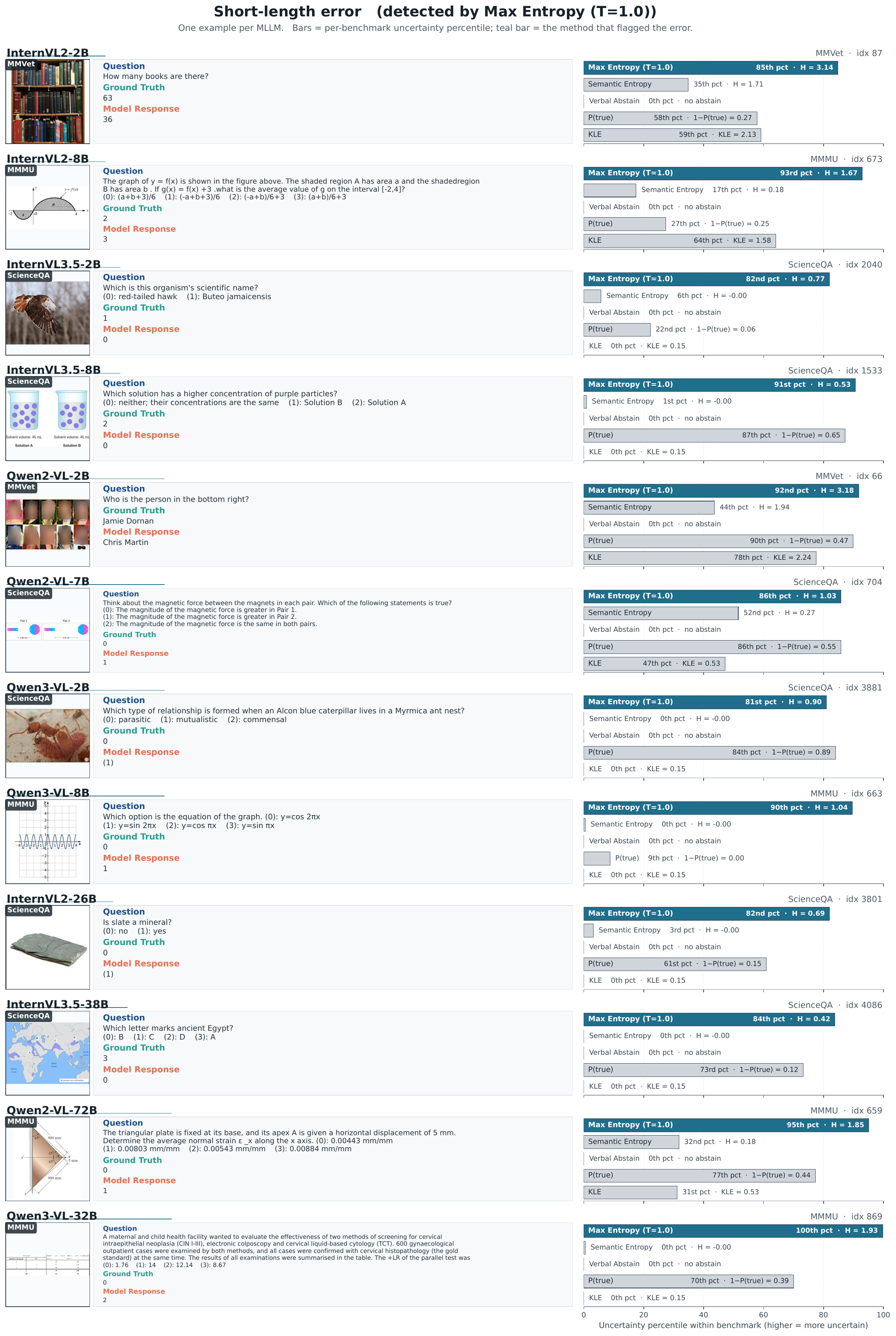}
    \caption{Short-length errors detected by Max Entropy ($T{=}1.0$). One example per MLLM (12 rows) from ScienceQA, MMMU and MM-Vet: the image, the question with all answer options, the ground truth and the (incorrect) deployed answer. Bars give the uncertainty percentile of each method within the benchmark (higher = more uncertain); the teal bar marks the method that flags the error. Token-level entropy at $T{=}1.0$ separates these errors even when all ten sampled answers agree (Semantic Entropy and KLE at the 0th percentile), the model never abstains, and P(true) is inconsistent.}
    \label{fig:qualitative_short}
\end{figure*}

\begin{figure*}
    \centering
    \includegraphics[width=0.9\textwidth]{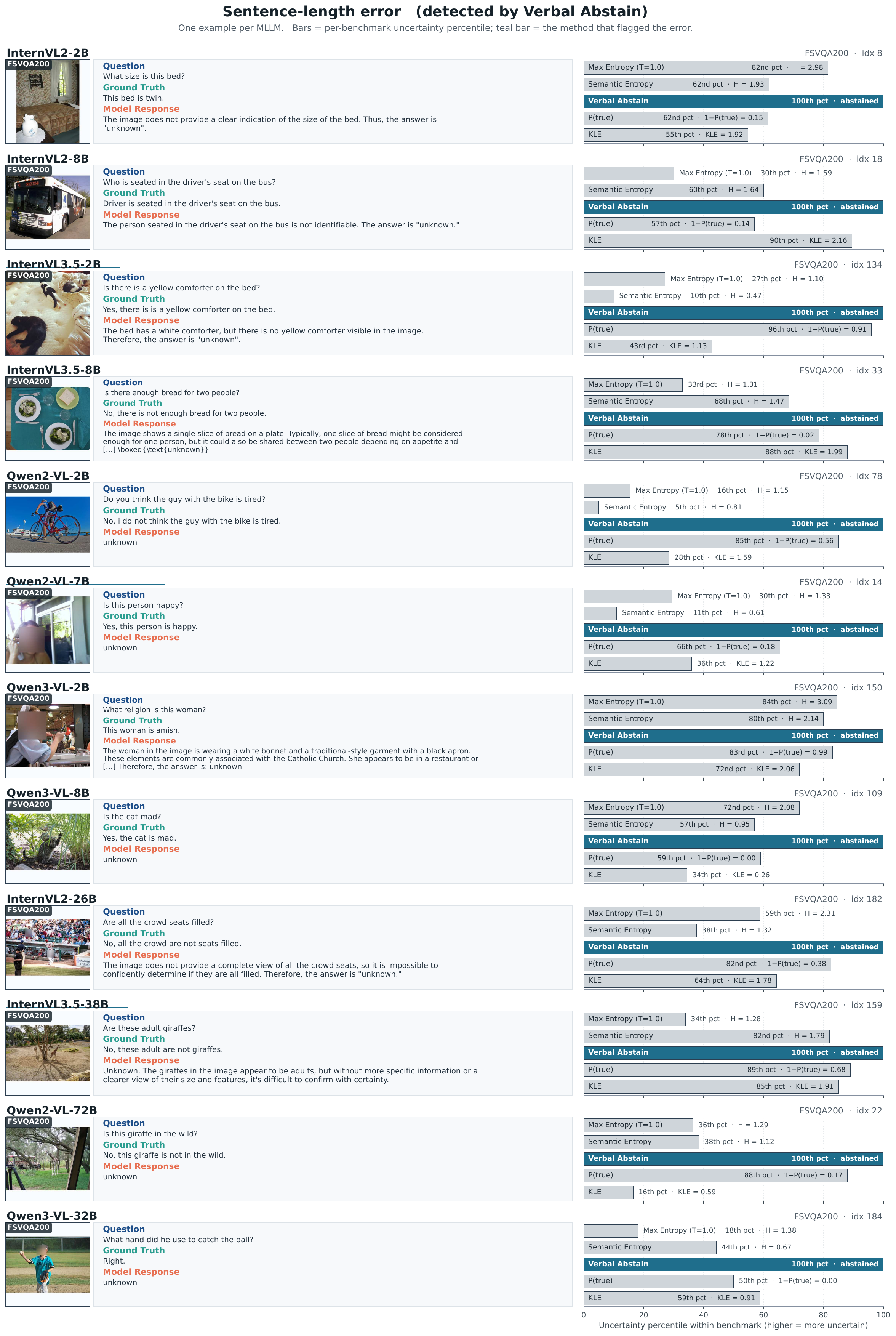}
    \caption{Sentence-length errors detected by Verbal Abstain. One example per MLLM (12 rows) from FSVQA200; the displayed response is the one produced under the abstention prompt. The model abstains (``unknown'') on every row, while the token-level and sampling-based scores of the same items fall anywhere between the 5th and the 90th percentile, so no single threshold on them would have caught the whole set.}

    \label{fig:qualitative_sentence}
\end{figure*}

\begin{figure*}
    \centering
    \includegraphics[width=0.9\textwidth]{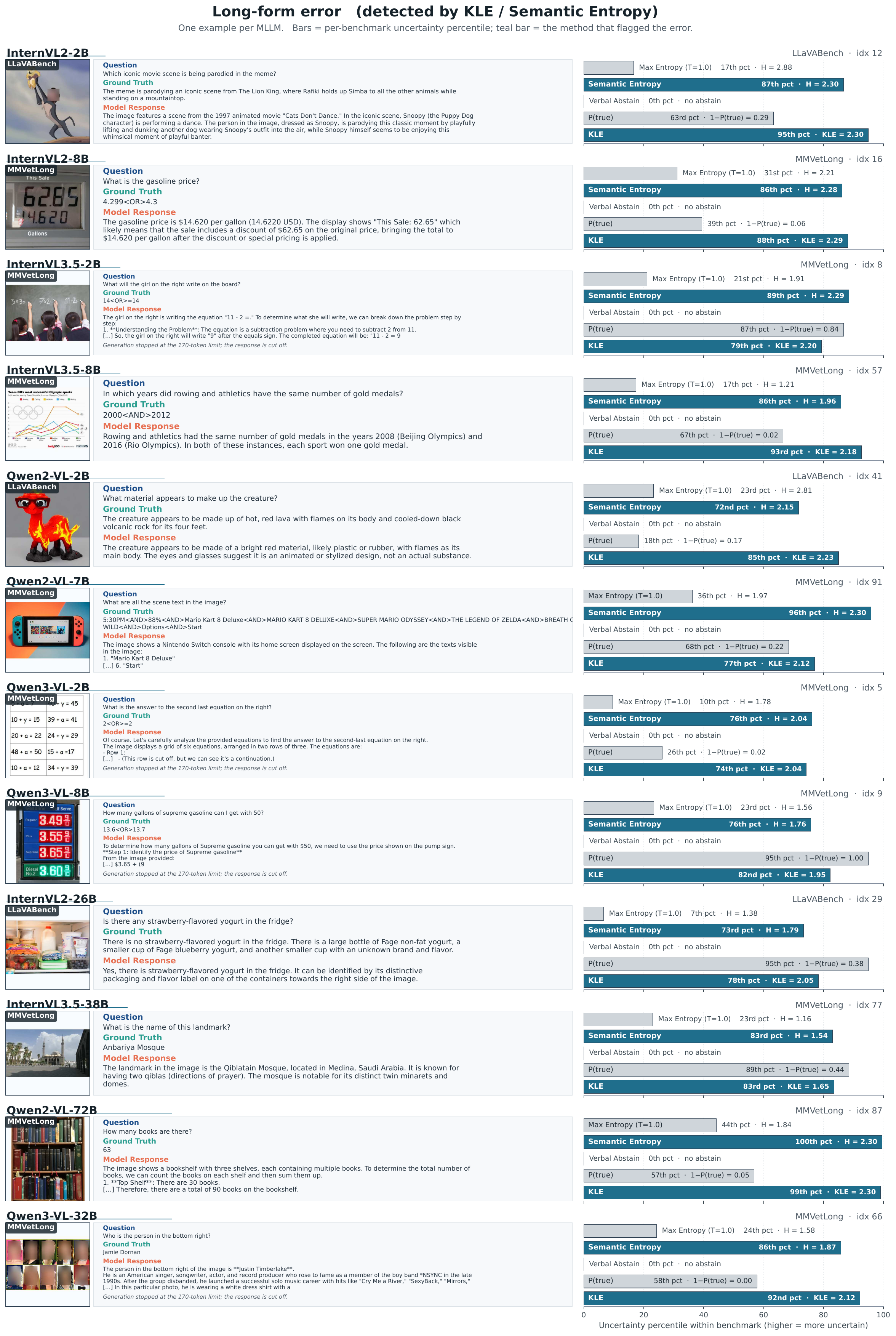}
    \caption{Long-form errors detected by KLE and Semantic Entropy. One example per MLLM (12 rows) from MM-Vet Long and LLaVA-Bench. Responses are abbreviated to their first lines and their last line (\emph{[\ldots]} marks the omitted middle); generation was capped at 170 tokens, which the Qwen3-VL models reach on almost all long-form prompts. The ten sampled answers scatter over many semantic clusters, so both semantic scores exceed the 70th percentile, whereas Max Entropy stays below the 55th percentile and the model never abstains; P(true) is high on some rows and low on others.}

    \label{fig:qualitative_long}
\end{figure*}

Figures~\ref{fig:qualitative_short}--\ref{fig:qualitative_long} illustrate the regime-dependent behavior summarized in Section~\ref{sec:results} on individual incorrect responses, one per MLLM. For each sample, we report the per-benchmark uncertainty percentile assigned by each method (higher = more uncertain), with the method that successfully flags the incorrect response highlighted. The examples make the failure modes concrete: on short-form questions (Fig.~\ref{fig:qualitative_short}), Max Entropy ($T{=}1.0$) places the wrong answer in the high-uncertainty tail while Semantic Entropy collapses to zero --- repeated samples produce the same short string, yielding a single cluster --- and Verbal Abstain rarely fires. On sentence-length FSVQA200 (Fig.~\ref{fig:qualitative_sentence}), the inverse pattern holds: models cleanly emit ``unknown'' on questions they get wrong, while token-level and semantic scores remain mid-range. On long-form generation (Fig.~\ref{fig:qualitative_long}), per-token entropy is diluted across many tokens and the abstention option is suppressed by the instruction to produce a complete answer, leaving meaning-level disagreement across resamples as the only signal that consistently separates correct from incorrect responses.

\begin{figure*}
    \centering
    \includegraphics[width=0.9\textwidth]{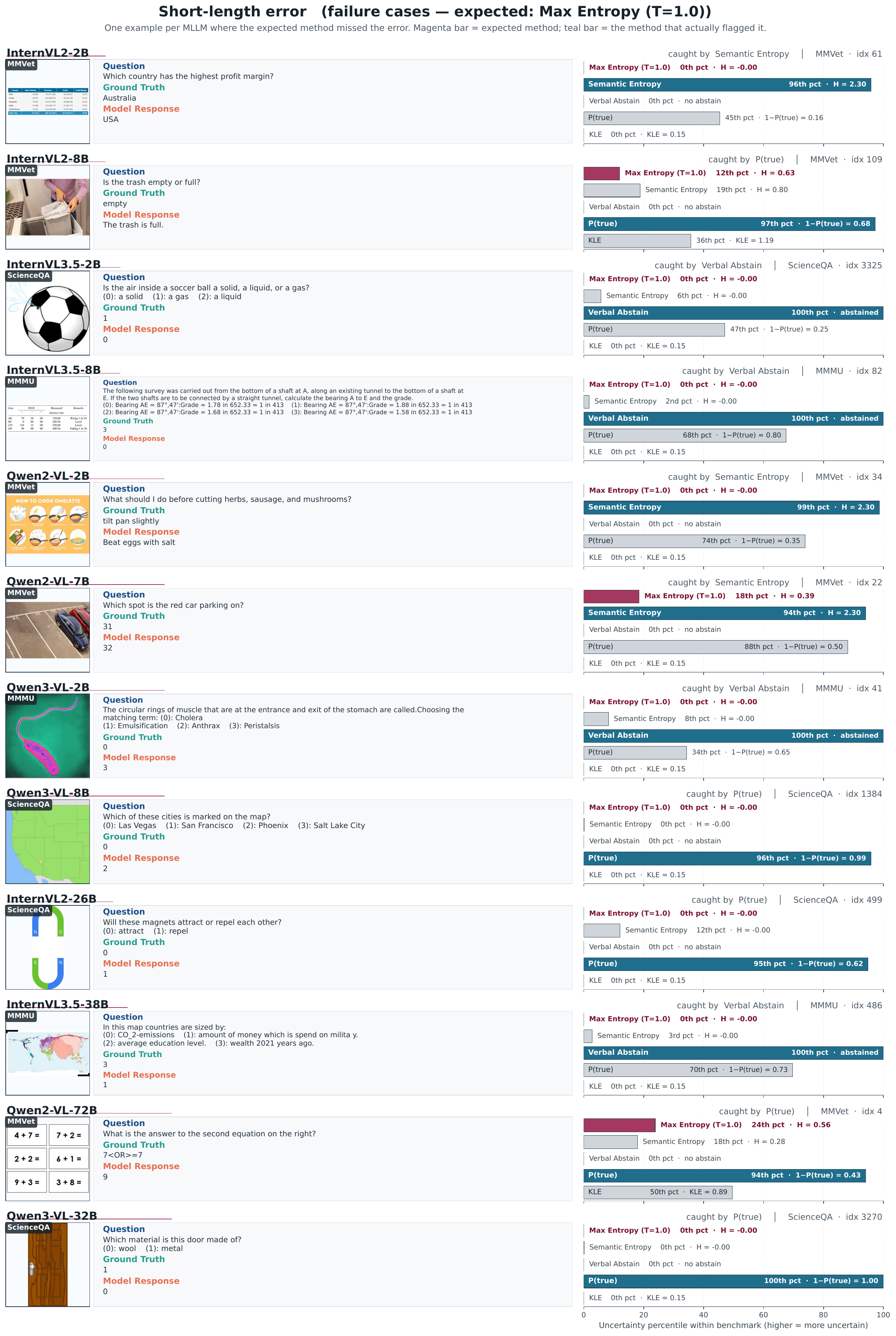}
    \caption{Short-length failure cases: errors that Max Entropy ($T{=}1.0$) misses. One example per MLLM (12 rows) from ScienceQA, MMMU and MM-Vet. The magenta bar is the expected method, which sits at or near the 0th percentile; the teal bar is the method that flags the error instead: P(true) on five rows, Verbal Abstain on four, and Semantic Entropy on three MM-Vet questions whose sampled answers disagree although the greedy answer is produced with near-zero entropy.}

    \label{fig:qualitative_short_fail}
\end{figure*}

\begin{figure*}
    \centering
    \includegraphics[width=0.9\textwidth]{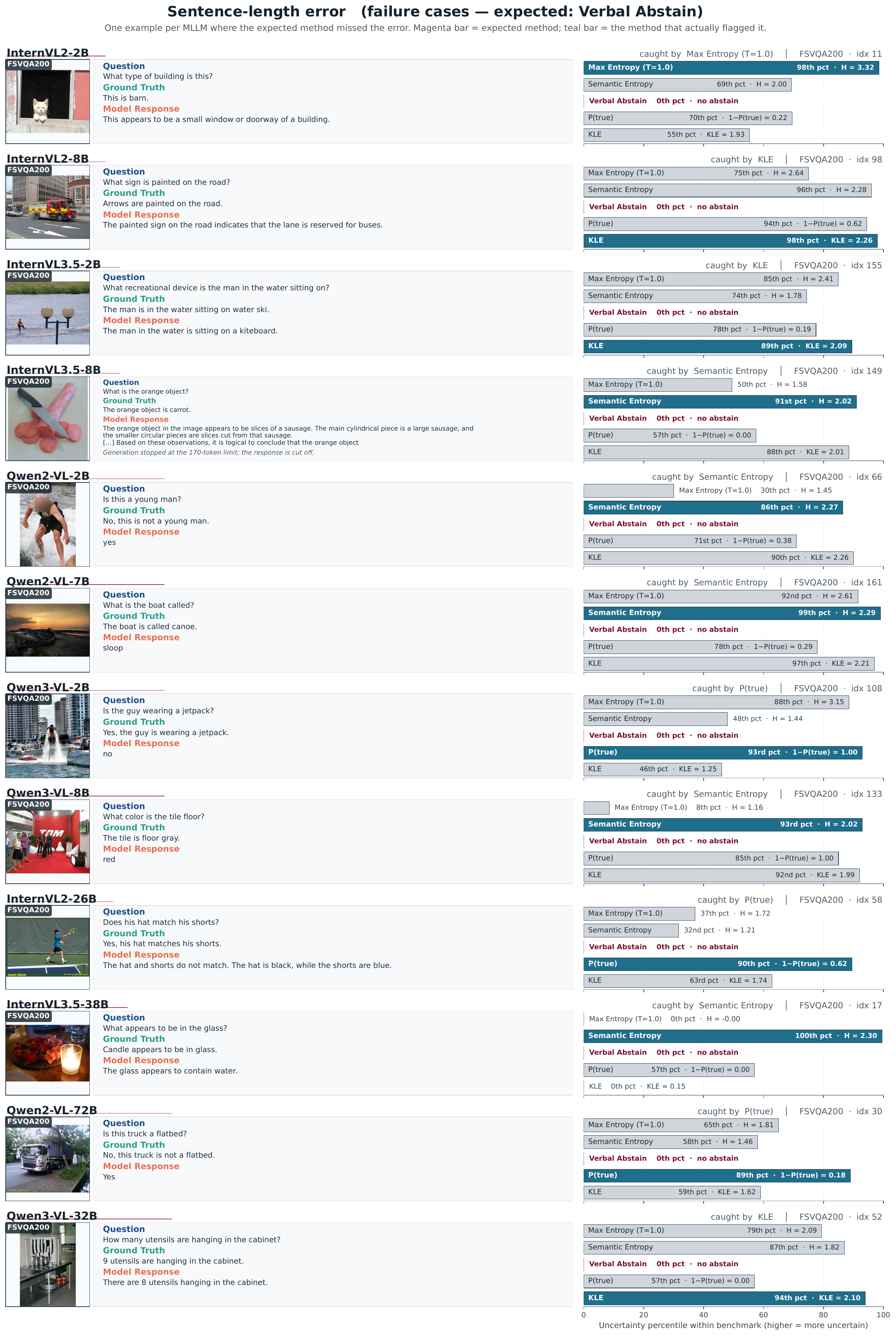}
    \caption{Sentence-length failure cases: errors on FSVQA200 that Verbal Abstain misses. One example per MLLM (12 rows); the model answers confidently and does not abstain on any row. The errors are hallucinated specifics (a misread sign, a kiteboard instead of water skis, eight instead of nine utensils), and they are flagged by KLE or Semantic Entropy when the samples disagree, by Max Entropy when the deployed answer itself is low-probability, and by P(true) on the yes/no questions.}

    \label{fig:qualitative_sentence_fail}
\end{figure*}

\begin{figure*}
    \centering
    \includegraphics[width=0.9\textwidth]{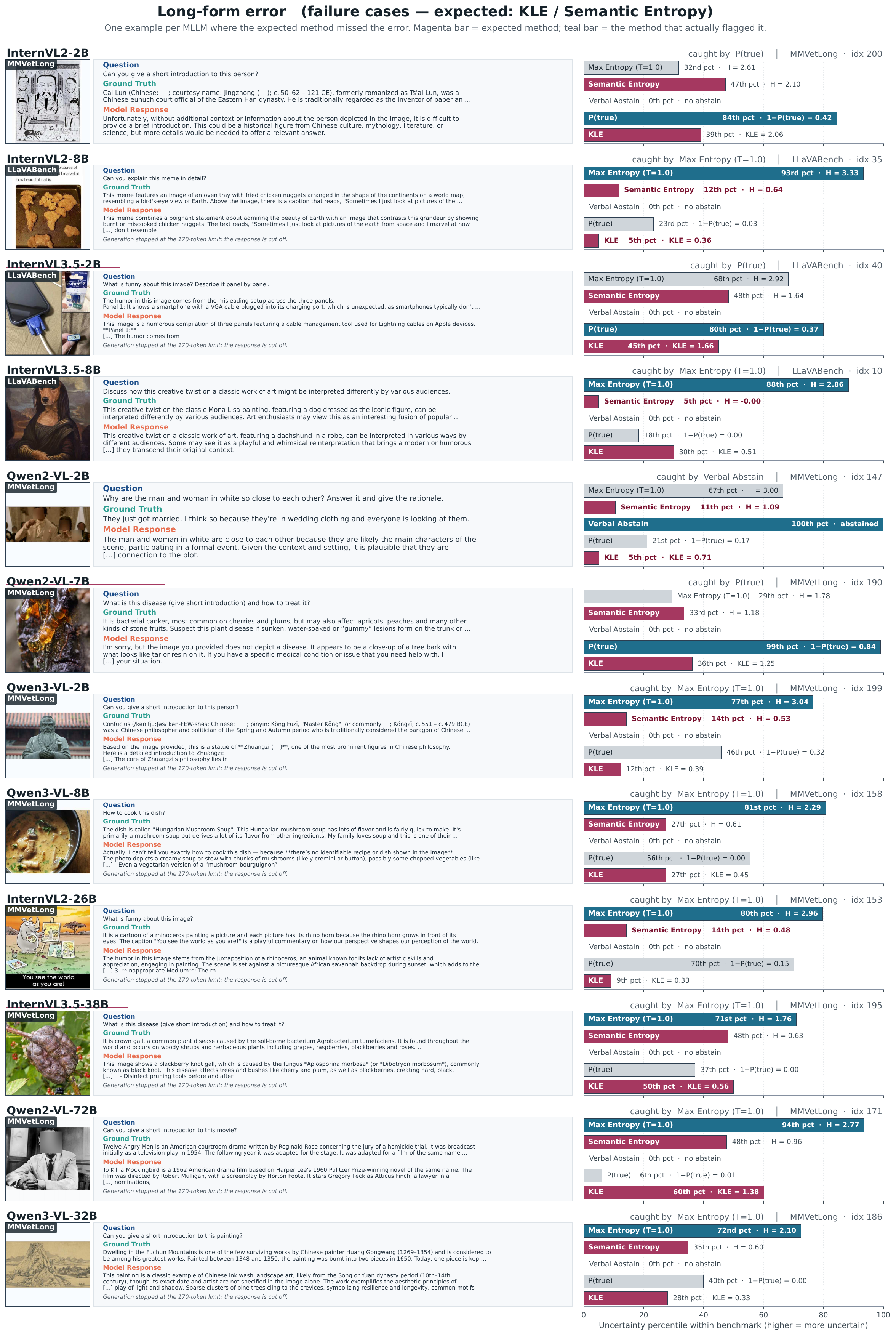}
    \caption{Long-form failure cases: errors that KLE and Semantic Entropy miss. One example per MLLM (12 rows) from MM-Vet Long and LLaVA-Bench, each with a long-form ground truth. Responses are abbreviated to their first and last lines ([\dots] marks the omitted middle); eight of them reach the 170-token generation cap. The sampled answers largely agree in meaning, so Semantic Entropy stays below the 50th percentile and KLE at or below the 60th, although the deployed answer is wrong, for example a misidentified person, film or plant disease, a misread meme or cartoon, or a refusal to answer. Magenta bars mark the expected methods; the teal bar marks the method that flags the error instead: Max Entropy ($T{=}1.0$) on eight rows, P(true) on three and Verbal Abstain on one.}

    \label{fig:qualitative_long_fail}
\end{figure*}

For transparency, we additionally include one failure case per regime (Figures~\ref{fig:qualitative_short_fail}--\ref{fig:qualitative_long_fail}), where the regime's best-performing method scores an incorrect response as low-uncertainty. These cases reflect the imperfect AUROC of even the strongest methods (0.72-0.76) and motivate the gap analysis in Section~\ref{sec:safety}.

\subsection{Licenses and Intended Use}
\label{sec:license}
All models and benchmarks used in this work are publicly released for research purposes. InternVL2 and InternVL3.5 are publicly released under the MIT License for their codebase, while their model weights are subject to the underlying Qwen and Apache 2.0 licenses depending on the checkpoint. Qwen2-VL is released under Apache 2.0 (for the 2B and 7B variants) and the Qwen license (for the 72B variant), while Qwen3-VL is released under the Apache 2.0 license. ScienceQA (CC BY-NC-SA 4.0), MMMU (Apache 2.0), MM-Vet (code under Apache 2.0; dataset under CC BY-NC 4.0), FSVQA (released for academic research use; derived from VQA and MS COCO and subject to their respective terms), and LLaVA-Bench (Apache 2.0) are used in accordance with their respective licenses and intended use for academic research on multimodal reasoning and evaluation. Any artifacts released as part of this work (including the FSVQA200 sample IDs and evaluation code) will be distributed under a permissive license such as MIT or CC BY 4.0, excluding any third-party benchmark content (notably the non-commercial ScienceQA and MM-Vet data), which remains subject to its original license.

\end{document}